\documentclass[onecolumn]{article}
\usepackage{amsmath}
\usepackage{multirow}
\usepackage[most]{tcolorbox}
\usepackage{longtable}
\usepackage{siunitx}

\usepackage[utf8]{inputenc}
\usepackage{enumitem}

\usepackage{listings}

\usepackage{xcolor}
\usepackage[a4paper,margin=2.5cm]{geometry} 

\usepackage{xcolor}
\usepackage{graphicx}

\usepackage{amsmath}
\usepackage{multirow}
\usepackage{pdfpages}
\usepackage{natbib}
\definecolor{colorLink}{rgb}{0.9,0,0} 
\definecolor{colorCite}{rgb}{0,0.7,0} 
\definecolor{colorURL} {rgb}{0,0,0.8} 
\usepackage[colorlinks=true,linktocpage=true,linkcolor=colorLink,citecolor=colorCite,urlcolor=colorURL]{hyperref}

\title{The \texttt{Denario} project:\\Deep knowledge AI agents for scientific discovery}
\author{
    Francisco Villaescusa-Navarro$^{1,2,*}$,
    Boris Bolliet$^{3,4,*}$,
    Pablo Villanueva-Domingo$^{5,*}$,\\
    Adrian E. Bayer$^{1,2}$, 
    Aidan Acquah$^{6}$,
    Chetana Amancharla$^{7}$,
    Almog Barzilay-Siegal$^{8}$,\\
    Pablo Bermejo$^{9,10,11}$,
    Camille Bilodeau$^{12}$,
    Pablo C\'ardenas Ram\'irez$^{13,14,15}$, 
    Miles Cranmer$^{16}$,\\
    Urbano L. Fran\c{c}a$^{17,18}$,
    ChangHoon Hahn$^{19, 20}$,
    Yan-Fei Jiang$^{1}$,
    Raul Jimenez$^{21,22}$,\\
    Jun-Young Lee$^{1}$,
    Antonio Lerario$^{23}$,
    Osman Mamun$^{13}$,
    Thomas Meier$^{24}$,\\
    Anupam A. Ojha$^{25,26}$,
    Pavlos Protopapas$^{27}$,
    Shimanto Roy$^{12}$,
    David N. Spergel$^{1}$,\\
    Pedro Taranc\'on-\'Alvarez$^{21,28}$,
    Ujjwal Tiwari$^{7}$,
    Matteo Viel$^{23, 29, 30, 31, 32}$, 
    Digvijay Wadekar$^{33, 40}$,\\
    Chi Wang$^{34}$,
    Bonny Y. Wang$^{35}$,
    Licong Xu$^{36,4}$,
    Yossi Yovel$^{8,37}$,
    Shuwen Yue$^{13}$,\\
    Wen-Han Zhou$^{38}$,
    Qiyao Zhu$^{25}$,
    Jiajun Zou$^{39}$,
    \'{I}\~{n}igo Zubeldia$^{16,36,4}$
}
\date{}

\begin{document}
\maketitle

\begin{center}

$^1$ Center for Computational Astrophysics, Flatiron Institute, New York, NY 10010, USA\\
$^2$ Department of Astrophysical Sciences, Princeton University, Princeton, NJ 08544, USA\\
$^3$ Cavendish Astrophysics, University of Cambridge, Madingley Road, Cambridge CB3 0HA, UK\\
$^4$ Kavli Institute for Cosmology, University of Cambridge, Madingley Road, Cambridge CB3 0HA, UK\\
$^5$ Computer Vision Center, Universitat Aut\`onoma de Barcelona, 08193 Bellaterra, Barcelona, Spain\\
$^{6}$ Big Data Institute, University of Oxford, Old Road Campus, Oxford OX3 7LF, UK\\
$^7$ Infosys Ltd\\
$^{8}$ School of Zoology, Faculty of Life Sciences, Tel-Aviv University, 6997801, Tel-Aviv Israel\\
$^9$ Donostia International Physics Center, Paseo Manuel de Lardizabal 4, E-20018 San Sebasti\'an, Spain\\
$^{10}$ Department of Applied Physics, University of the Basque
Country, 20018 San Sebastián, Spain\\
$^{11}$ Center for Computational Quantum Physics, Flatiron Institute, New York, NY 10010, USA\\
$^{12}$ Chemical Engineering Department, University of Virginia, Wilsdorf Hall, Charlottesville, VA 22903\\
$^{13}$ Robert F. Smith School of Chemical and Biomolecular Engineering, Cornell University, Ithaca, NY 14853, USA\\
$^{14}$ Ragon Institute of Mass General, MIT, and Harvard, 600 Main St, Cambridge, MA 02139, USA\\
$^{15}$ Department of Microbiology, Harvard Medical School, 77 Ave Louis Pasteur, Boston, MA 02115, USA\\
$^{16}$ Department of Applied Mathematics and Theoretical Physics, University of Cambridge, Cambridge, UK\\
$^{17}$ Boston Children's Hospital, Department of Anesthesiology, Critical Care and Pain Medicine, 300 Longwood Ave, Bader 6, Boston, MA, 02115, USA\\
$^{18}$ Harvard Medical School, 25 Shattuck Street Boston, MA, 02115, USA \\
$^{19}$ Department of Astronomy, The University of Texas at Austin, Austin, TX 78712, USA\\
$^{20}$ Steward Observatory, University of Arizona, 933 N. Cherry Avenue, Tucson, AZ 85721, USA \\
$^{21}$ Institut de Ci\'encies del Cosmos, Universitat de Barcelona, Mart\'i i Franqu\'es 1, Barcelona, Spain.\\
$^{22}$ ICREA, Pg. Lluis Companys 23, Barcelona, 08010, Spain\\
$^{23}$ SISSA - International School for Advanced Studies, Via Bonomea 265, I-34136 Trieste, Italy\\
$^{24}$ MCML - Munich Center for Machine Learning, LMU Munich, 80539 Munich, Germany\\
$^{25}$ Center for Computational Biology, Flatiron Institute, 162 5th Ave., New York, NY 10010, USA \\
$^{26}$ Center for Computational Mathematics, Flatiron Institute, 162 5th Ave., New York, NY 10010, USA \\
$^{27}$ Institute for Applied Computational Science, Harvard University, 150 Western Ave., MA 02134, USA\\
$^{28}$ Departament de F\'isica Qu\'antica i Astrof\'isica, Universitat de Barcelona, Barcelona, Spain\\
$^{29}$ INFN -- National Institute for Nuclear Physics, Via Valerio 2, 34127 Trieste, Italy\\
$^{30}$ ICSC - Centro Nazionale di Ricerca in High Performance Computing, Big Data e Quantum Computing, Via Magnanelli 2, Bologna, Italy\\
$^{31}$ INAF -- Osservatorio Astronomico di Trieste, Via G. B. Tiepolo 11, I-34143 Trieste, Italy\\
$^{32}$ IFPU -- Institute for Fundamental Physics of the Universe, Via Beirut 2, I-34151 Trieste, Italy\\
$^{33}$ Department of Physics and Astronomy, Johns Hopkins University, Baltimore, Maryland, 21218, USA\\
$^{34}$ Google DeepMind\\
$^{35}$ Department of Astronomy and Astrophysics, The University of Chicago, Chicago, IL 60637, USA\\
$^{36}$ Institute of Astronomy, University of Cambridge, Madingley Road, Cambridge CB3 0HA, UK\\
$^{37}$ Sagol School of Neuroscience, Tel-Aviv University, 6997801, Tel-Aviv Israel\\
$^{38}$ Department of Earth and Planetary Science, The University of Tokyo, Tokyo 113-8654, Japan \\
$^{39}$ SENSEx Lab, International School for Advanced Studies (SISSA), 34136 Trieste, Italy\\
$^{40}$ Weinberg Institute, University of Texas at Austin, Austin, TX 78712, USA\\

\end{center}

\def\thefootnote{*}\footnotetext{Equal contribution. Listing order of FVN, BB, PVD is random.}\def\thefootnote{\arabic{footnote}}

\begin{abstract}

We present \href{https://github.com/AstroPilot-AI/Denario}{\texttt{Denario}}, an AI multi-agent system designed to serve as a scientific research assistant. \href{https://github.com/AstroPilot-AI/Denario}{\texttt{Denario}} can perform many different tasks, such as generating ideas, checking the literature, developing research plans, writing and executing code, making plots, and drafting and reviewing a scientific paper. The system has a modular architecture, allowing it to handle specific tasks, such as generating an idea, or carrying out end-to-end scientific analysis using  \href{https://github.com/CMBAgents/cmbagent}{\texttt{Cmbagent}} as a deep-research backend. In this work, we describe in detail \href{https://github.com/AstroPilot-AI/Denario}{\texttt{Denario}} and its modules, and illustrate its capabilities by presenting multiple AI-generated papers generated by it in many different scientific disciplines such as astrophysics, biology, biophysics, biomedical informatics, chemistry,  material science, mathematical physics, medicine, neuroscience and planetary science. \href{https://github.com/AstroPilot-AI/Denario}{\texttt{Denario}} also excels at combining ideas from different disciplines, and we illustrate this by showing a paper that applies methods from quantum physics and machine learning to astrophysical data. We report the evaluations performed on these papers by domain experts, who provided both numerical scores and review-like feedback. We then highlight the strengths, weaknesses, and limitations of the current system. Finally, we discuss the ethical implications of AI-driven research and reflect on how such technology relates to the philosophy of science. We publicly release the code at \href{https://github.com/AstroPilot-AI/Denario}{this URL}. A \href{https://github.com/AstroPilot-AI/Denario}{\texttt{Denario}} demo can also be run directly on the web at \href{https://huggingface.co/spaces/astropilot-ai/Denario}{this URL}, and the full app will be deployed on the cloud. 
\end{abstract}

\setcounter{tocdepth}{2}
\tableofcontents

\section{Introduction} 
\label{sec:intro}

Modern science has been built on the rules of the scientific method \citep{Chalmers}. This empirical and iterative method begins by formulating an idea or hypothesis, which represents the foundation for the entire research endeavor. This initial idea might arise from a variety of sources: a theoretical question, an unexplained phenomenon, an unexpected result in prior work, or a curiosity-driven exploration of a dataset or simulation. The hypothesis can aim to explain an observed behavior in nature, test the validity of a theoretical model, or uncover new patterns or relationships within existing data.

Once the idea is identified, it is often refined through collaboration discussions, preliminary literature reviews, and feasibility assessments. Researchers seek to understand the broader context of the problem, what has already been studied, and where the current knowledge gaps lie. This step is crucial for ensuring that the proposed work is novel, feasible, and impactful.

Next, a detailed plan is developed to execute the project and investigate the proposed idea. This stage involves identifying the necessary data, selecting or designing appropriate methodologies, and outlining the sequence of steps required for the analysis. Depending on the nature of the research, this may include data collection or simulation, data preprocessing and cleaning, implementing algorithms or analytical models, and performing statistical or computational analyses. Along this process, scientists may use lower-dimensional representations of the data, e.g., via plots or diagrams, to help interpret intermediate results and to communicate findings effectively. The ultimate goal of this phase is to systematically explore the hypothesis, extract meaningful insights, and arrive at scientifically sound conclusions.

Finally, a scientific paper or report is written describing in detail the problem or hypothesis, the methods used, the results obtained, and the conclusions reached, describing how the research has advanced knowledge and its connections with previous works.

Each of the above steps is a highly sophisticated task that requires extensive time for highly skilled domain experts. Because of this, the generation of a scientific paper can take from months to years. Unfortunately, a significant fraction of that time is consumed on tasks such as literature search, learning new methods, debugging code, and writing text. These are tasks where recent advances in Artificial Intelligence (AI), such as large language models and AI agents, can help. Thus, developing AI tools that can assist scientists with some or all of the tasks involved in the generation of new knowledge can accelerate scientific discovery by enabling scientists to formulate new hypotheses and test those faster than in the traditional method.

In this work, we introduce and describe \texttt{Denario}, an AI multi-agent system designed to serve as a scientific research assistant. \texttt{Denario} can generate ideas, check the literature to evaluate the novelty of those, develop research plans, write and execute code, make plots, and draft and review scientific papers. It can also perform end-to-end scientific research by carrying out, sequentially, all the above tasks. We evaluate the capabilities and limitations of \texttt{Denario} by generating paper drafts from a very diverse set of disciplines, from biology to astrophysics, and reporting the positive and negative aspects of these papers, as judged by domain experts.  Hoping to be a tool that can accelerate scientific discovery, we also make \texttt{Denario} publicly available.

We want to stress that \texttt{Denario} is not the first attempt at building tools to help researchers perform end-to-end research. The idea of having machines perform research is much older. For instance, the robot scientist (also known as Adam) \citep{Robot_scientist} was able to formulate hypotheses, plan and run experiments (physically via robotics), interpret the results, and repeat the cycle. Adam was designed to study functional genomics and was the first machine to generate scientific knowledge independently of humans. The same group later developed Eve \cite{Eve_scientist}, another robot scientist tailored towards drug discovery. Another early important work is the Automated Statistician \citep{Automated_statistician}, which was able to perform a detailed statistical analysis of some data and provide a report about its findings. More recently, lots of recent work on AI assistants for research has been done by many different teams, such as Sakana AI-scientist \citep{AI_scientist, AI-scientist2}, Google co-scientist \citep{co-scientist}, AI-cosmologist \citep{AI_cosmologist}, Curie \citep{Curie}, and Agent laboratory \citep{schmidgall2025agentlaboratoryusingllm}.

We note that lots of works are also developing tools to help scientists with increasingly sophisticated tasks. For instance, AutoWEKA was developed to identify the best machine learning model to tackle a particular problem \citep{AutoWEKA}. ResearchAgent \citep{ResearchAgent} was able to formulate new hypotheses and design experiments by checking their novelty with the literature. \citep{Rapp_2023} created a self-driving lab to re-engineer enzymes without human intervention, while \citep{Swanson_2024} created a virtual lab with several AI-scientist experts in different areas capable of creating nanobodies. Another example is AstroAgents \citep{AstroAgents}, a multi-agent system designed to test and consolidate many different hypotheses about the origins of life on Earth. Recent works have also shown that AI systems can write expert-level scientific software and discover new algorithms \cite{AlphaEvolve, Google_software, ShinkaEvolve}. Other works are exploring the capabilities of agentic systems via graphs \citep{AgenticGraph1, AgenticGraph2}. We also note that AgentRxiv \citep{AgentRxiv} is a framework for Agents to learn from previous works.

While some of these technologies have the potential to accelerate scientific discovery --for instance, \citep{AI_chemist} designed an autonomous system combining AI with robotics to discover new materials a lot of serious debates about human standards for validation sparked \citep{Material_rebuttal}. These are very important considerations and we discuss them in detail, for the case of \texttt{Denario}. Furthermore, given the ethical considerations raised by such a technology, we devote a paper section to discuss ethical aspects, along with how this technology fits within the philosophy of science. 

We note that in scientific research, there is usually a tradeoff between depth and breadth, given a fixed amount of time to carry out a task or study. We believe that AI systems like \texttt{Denario} can help researchers explore a wide range of possibilities, at a more superficial level, while researchers can go deeper into the most interesting or promising ones. We emphasize that the goal of \texttt{Denario} is not to automate science, but to develop a research assistant that can accelerate scientific discovery. By making \texttt{Denario} publicly available and user-friendly, we hope it will become a standard tool that benefits the scientific community in their enterprises. 

This paper is organized as follows. In Section \ref{sec:methods}, we introduce the different pieces that made our system, such as LLMs, AI agents, and agentic orchestration. In Sec. \ref{sec:architecture} we describe in detail the different modules in \texttt{Denario} and show the different ways to run the system in Sec. \ref{sec:deployment}. In Sec. \ref{sec:examples}, we show different AI-generated papers from many different disciplines and discuss the positive and negative aspects of those. We then present in Sec. \ref{sec:discussion} some of the observed failure modes of \texttt{Denario} together with advantages, limitations, and future directions. We discuss the ethical considerations that systems like \texttt{Denario} raise and how the technology powering is seen from the philosophy of science in Sec. \ref{sec:ethics}. We conclude in Sec. \ref{sec:conclusions}.

\section{AI multi-agent systems} 
\label{sec:methods}

\texttt{Denario} is an AI multi-agent framework designed to function as a scientific research assistant. It is built with \texttt{AG2}  and \texttt{LangGraph}, and designed to assist in different stages of the research process. It can also execute multiple tasks sequentially, enabling it to perform end-to-end scientific research. 

\begin{figure*}[ht!]
\centering
    \includegraphics[width=1\textwidth]{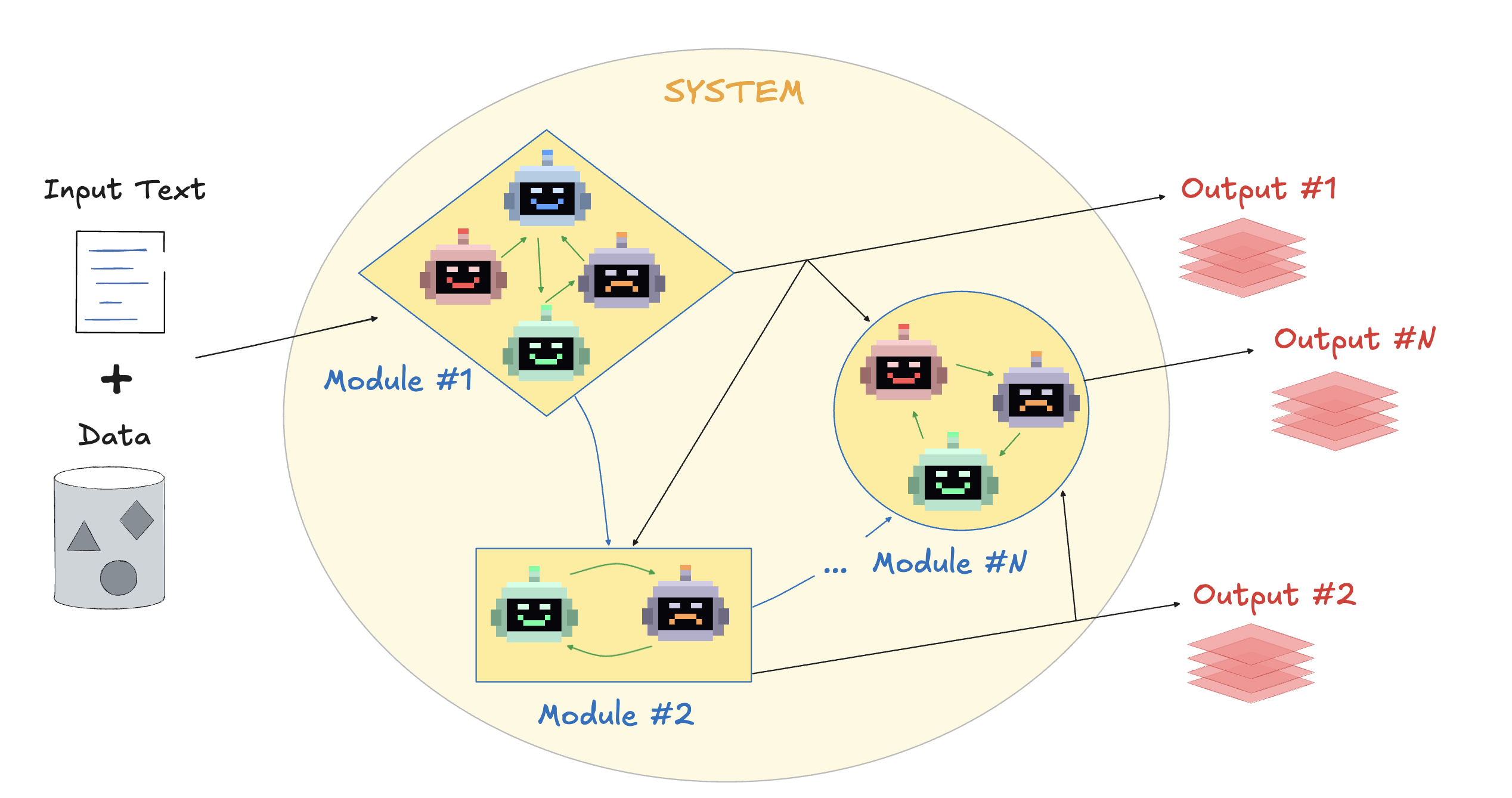}
    \caption{\label{fig:AI_system} This cartoon shows the different components of \texttt{Denario} and their interplay. The system is shown as an orange circle, and it takes some input text and data (left side), and can generate one or several outputs (red icons on the right). The system is composed of different modules (yellow shapes), which can exchange messages among them (blue arrows outside modules). The modules are composed of multiple agents (bots icons), that can communicate with each other (blue arrows inside modules).
    }
\end{figure*}

The system is composed of several modules, each focused on a specific task. Within each module, there are agents—autonomous or semi-autonomous computational entities that perceive their environment, process information, and take actions to achieve defined goals. Some of these are AI agents, which leverage Large Language Models (LLMs) to perform specialized tasks. Agents can communicate and collaborate to solve complex problems, while orchestration manages their interactions—determining task order, information flow, and overall coordination. The system takes some input text and returns one or several outputs depending on how \texttt{Denario} is run. We show a cartoon representation of the system in Fig. \ref{fig:AI_system}.

In this section, we describe the different components relevant to setting the agents, including the LLM models and the orchestration.

\subsection{Large Language Models}

The majority of the agents in \texttt{Denario} are AI-agents that typically call Large Language Models (LLMs) \citep{2024arXiv240206196M}, including multi-modal ones, with a very specialized prompt that may depend on the output of other agents. \texttt{Denario} allows the user to choose among many different models, including:
\begin{itemize}
    \item \texttt{gpt-5}, \texttt{gpt-5-mini}, \texttt{gpt-4o}, \texttt{gpt-4.1}, \texttt{gpt-4.1-mini}, \texttt{o3-mini} from OpenAI,
    \item \texttt{gemini-2.5-pro}, \texttt{gemini-2.5-flash}, \texttt{gemini-2.0-flash} from Google,
    \item \texttt{claude-3-7-sonnet}, \texttt{claude-4-Opus}, \texttt{claude-4.1-Opus}, \texttt{claude-4.5} from Anthropic.
\end{itemize}
We note that \texttt{Denario} is not limited to these LLMs, and new models, or models from other providers, can easily be incorporated into the system. Our choice of models has so far been made based on trial and error rather than systematic evaluation. For example, for coding, our primary models of choice are \texttt{gemini-2.5-pro}, \texttt{gpt-4.1}, and \texttt{o3-mini}. For critiquing and reviewing, our default choice is Claude models or reasoning models from OpenAI.

\subsection{Multi-Agent Orchestration and Strategy}
\label{sec:obs}

In \texttt{Denario}, the agent orchestration is carried out with two different frameworks: \texttt{AG2}\footnote{\url{https://ag2.ai/}} and \texttt{LangGraph}\footnote{\url{https://www.langchain.com/LangGraph}}. Some modules (e.g. idea module) are implemented in both frameworks, while others (e.g. analysis module) are only implemented in one framework. Some modules implemented in \texttt{AG2}  follow the \textit{Planning \& Control} strategy as implemented in \texttt{Cmbagent} that we describe below.

\subsubsection{Orchestration}

\begin{itemize}
\item \textbf{AG2}. \texttt{AG2} is an open-source AgentOS that supports advanced agentic AI applications and research. It offers an agentic abstraction that unifies AI foundation models like LLMs, non-LLM tools and human inputs, and intuitive conversation programming to orchestrate multiple agents. Stronger agents or agentic systems can be composed recursively using primitive agents and basic conversation patterns like sequential chats, group chats, nested chats. This modular architecture also enables rich agentic design patterns such as advanced tool use, structured context, conditional handoffs, dynamic updates, dynamic task decomposition and team building, guardrails, and many other extensions. All agents using \texttt{AG2} are part of \texttt{Cmbagent} \citep{Laverick:2024fyh,cmbagent2,CMBAGENT_2025}, an open source and publicly available\footnote{\url{https://github.com/CMBAgents/cmbagent}} multi-agent system designed to perform a variety of scientific research-related tasks. 

\item \textbf{LangGraph}. \texttt{LangGraph} is based on \texttt{LangChain}\footnote{\url{https://www.langchain.com/}} and is an open-source Python framework designed to orchestrate agents through directed graphs. In \texttt{LangGraph}, the graph nodes represent callable units, such as agents, functions, and tools. The edges between nodes represent the way the flow of the computation is carried out. Another important component of \texttt{LangGraph} is the graph state, an object (typically a Python class) passed through the different nodes and enabling agent communication, memory, coordination, etc. \texttt{LangGraph} supports both static and dynamic graphs. For instance, depending on the outcome of an agent, one may want to call a particular agent. Loops and interrupts are also allowed in \texttt{LangGraph}. 

\end{itemize}

\subsubsection{Planning and Control Strategy}

\begin{figure*}[t]
\centering
    \includegraphics[width=0.8\textwidth,trim=0 0 0 0.5cm,clip]{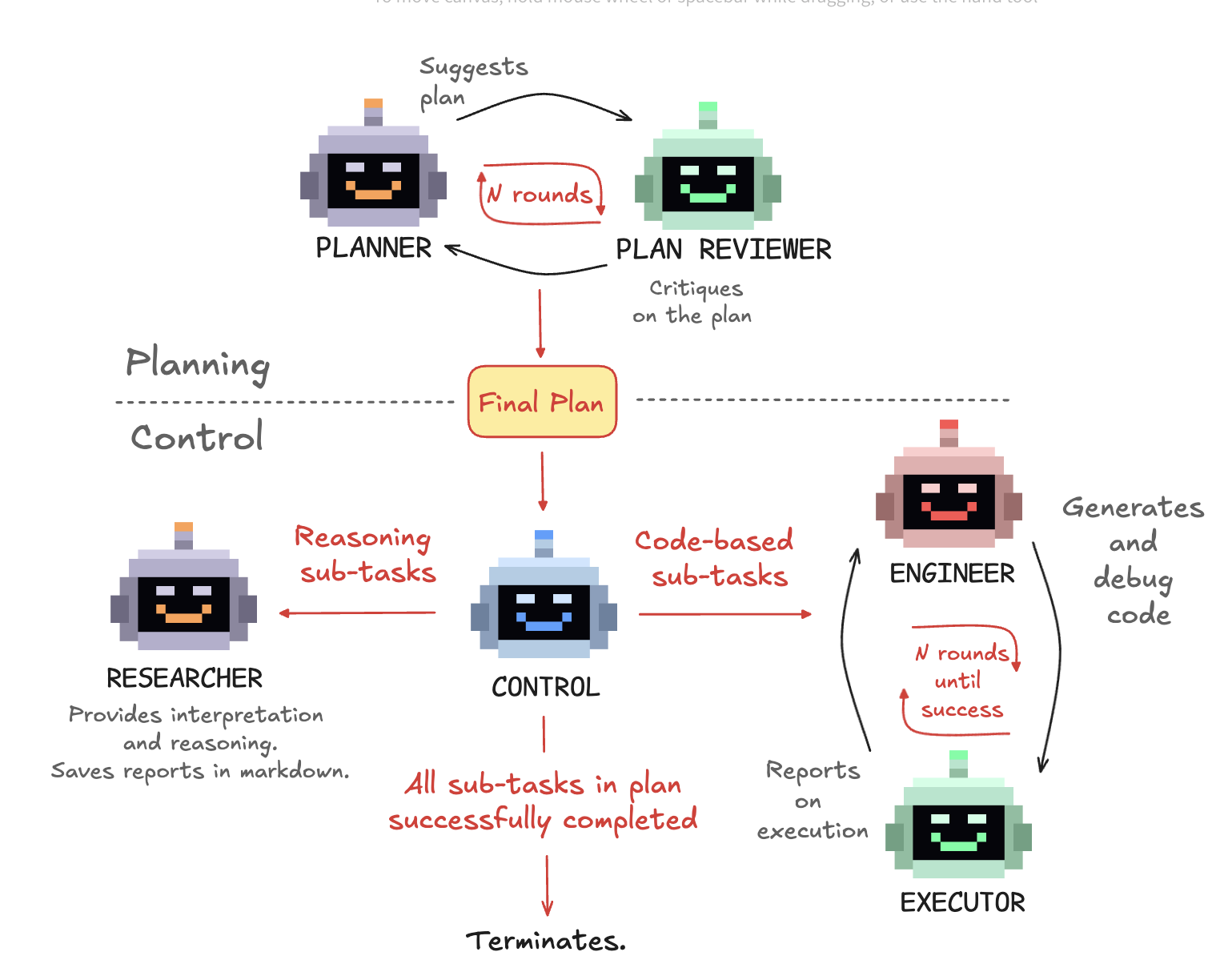}
    \caption{\label{fig:pc_cmbagent} \textit{Planning \& Control} implementation of ``Deep Research" from  \href{https://github.com/CMBAgents/cmbagent}{Cmbagent}. This is the essential orchestration strategy enabling the research analysis without human-in-the-loop. Additionally, it can be used for idea generation, method generation and data pre-processing.}
\end{figure*}

\texttt{Planning \& Control}, a fundamental concept in robotics, refers to strategies for orchestrating decision-making and action execution. The basic idea is that during the planning phase, a complex task is decomposed into smaller, manageable subtasks, and a strategy is defined for achieving them. During the control phase, these subtasks are executed, with monitoring and adjustments as needed to ensure the overall goal is achieved. Some modules in \texttt{Denario} follow this strategy, as implemented in \texttt{Cmbagent}. We now briefly describe the \texttt{Cmbagent} \texttt{Planning \& Control} implementation and refer the reader to \cite{cmbagent2} for further details. The strategy is illustrated in figure \ref{fig:pc_cmbagent}.

In \texttt{Cmbagent}, during the planning phase, the main task is broken down into sub-tasks, and an agent is assigned to each of them. During control, the agent responsible for a sub-task executes its instructions until completion. Once a sub-task is complete, a control agent records its status and proceeds to the next sub-task, continuing until the entire plan is complete.

The \textit{Planning} stage starts with a call to a \texttt{plan\_setter} agent, whose task is to select the set of agents that should be involved during the session. This function call stores into the context detailed instructions for the plan (namely, the agents and description of agents that should be involved) and ends on the next agent target, which is the \texttt{planner} agent. Then, the elaboration of a plan is done via a two-agent conversation where a \texttt{planner} agent suggests plans and a \texttt{plan\_reviewer} agent suggests recommendations on the plan. Recommendations and suggested plans are recorded in the context via function calls, including the number of steps in the plan. This information is injected into each agent's system message before reply. The number of review rounds, $n_\mathrm{reviews}$, is a 
hyperparameter of our system, and we generally set it to one\footnote{With more rounds, we often observe that the suggested plans become overly complex and ineffective.}. Once the number of rounds of recommendation $n_\mathrm{reviews}$ is reached, the final plan is recorded into the context, and the \textit{Planning} stage ends. The response of the \texttt{planner} is a list of  subtasks, where each subtask consists of three fields: 
\begin{itemize}
    \item \texttt{sub\_task}: the sub-task to be performed
    \item \texttt{sub\_task\_agent}: the name of the agent in charge of the sub-task.
    \item \texttt{bullet\_points}: A bullet-point list describing the specific actions the subtask is expected to perform.
\end{itemize}
The number of subtasks is constrained from the prompt of the \texttt{planner} and \texttt{plan\_reviewer} agents, which are instructed to produce a plan with, at most, $n_\mathrm{steps}$ steps, another tunable hyperparameter of our system. Typically, our plans contain between 3 and 8 steps\footnote{As the context passed to the agents tends to grow during a session, a plan with more steps will generally lead to a more costly session.}. 

The \textit{Control} stage is orchestrated by a \texttt{control} agent. The plan obtained from the \textit{Planning} stage is injected into the system message of the \texttt{control} agent and it replies by calling the \texttt{record\_status} function. This function updates the context variable dictionary to record the current step being executed, whether it is completed, failed, or in progress, whether some new plots or new code have been produced, whether some code execution has failed, and, depending on this information, which agent to transfer to next. The \texttt{control} agent records two types of termination signals: whether the number of times a code execution has failed exceeds the maximum number of failures allowed by the user, or whether the final step of the plan has been carried out successfully. In the former, the session aborts before the main task has been solved, and in the latter, the session ends positively with the full plan completely executed. Upon reception of a termination signal, \texttt{control} agent transfers to a \texttt{terminator} agent to end the session smoothly. 

Given the fact that the system may fall into an infinite loop (e.g., trying unsuccessfully to fix a bug), we set a hard limit to the maximum number of messages exchanged between the agents, $n_\mathrm{rounds}$. By default, we set this to $n_\mathrm{rounds}=500$. The system will fail if more than $n_\mathrm{rounds}$ are exchanged. We note the system can fail for many other reasons, including limits on API calls set on the provider platforms, or occasional exceptions/errors that are not fully dealt with.

\section{Architecture}
\label{sec:architecture}

\begin{figure*}[ht!]
\centering
    \includegraphics[width=1\textwidth]{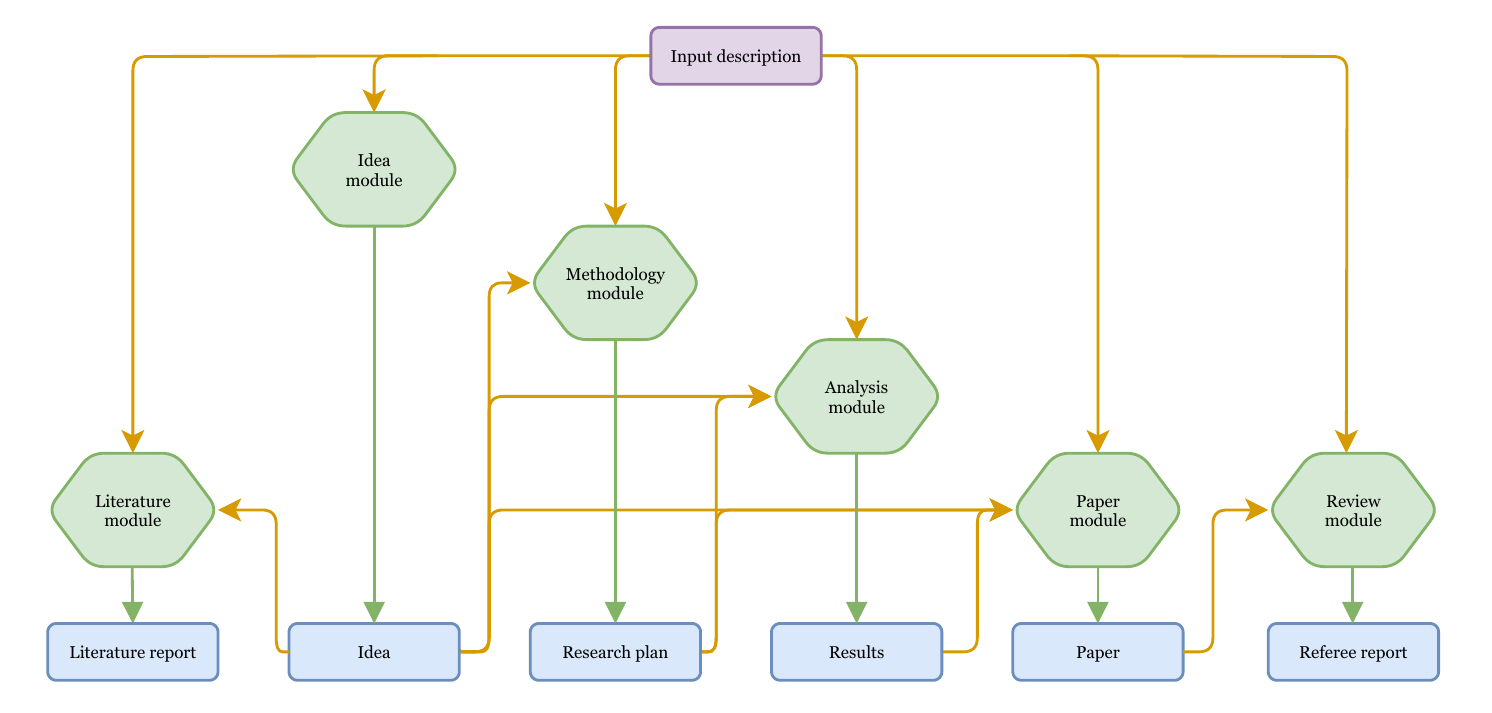}
    \caption{\label{fig:general_scheme} This scheme shows the architecture of \texttt{Denario}. The green hexagons represent the different modules in \texttt{Denario}. These modules are agentic systems that can 1) generate an idea for a research project, 2) check the literature to validate the novelty of the idea, 3) develop a plan to carry out the project, 4) implement the plan by writing and executing code, making plots, and describing the results, 5) write a scientific paper discussing the research and showing the plots, and 6) review the paper. Orange arrows indicate the input to the modules, while the green arrows show the output of them. The output of the modules are shown in the blue boxes. For details about the output, see Table \ref{tab:denario_io}. At the top, in magenta, there is the input description, where the user describes the data, postulates the problem... etc. 
    We note that \texttt{Denario} is fully modular, and human input can be added at any point. For instance, \texttt{Denario} can be run as a standalone program and generate a paper from the input text file (by running the idea, methods, analysis, and paper modules sequentially), or can also be used as a standalone tool to create plots and results from an input text, idea, and methods provided by the user.}
\end{figure*}

\texttt{Denario} is built as a modular system, allowing the user to execute a single or a combination of modules. Each module is an AI agentic system designed to carry out a specific scientific task. Figure~\ref{fig:general_scheme} illustrates the overall workflow and architecture of \texttt{Denario}.

For most of the modules, \texttt{Denario} requires an input text guiding the system's behavior. This text can contain the description of the data to be used together with some computational constraints. It can also simply contain a description of a problem to be solved. For some research applications, it can also be relevant to specify that the data can (or should) be generated by the agents. The six main modules in \texttt{Denario} can perform these tasks (see Table \ref{tab:denario_io} for a summary): 

\begin{table}[ht]
\centering
\begin{tabular}{|c|c|c|c|}
\hline
\textbf{Module} & \textbf{Task} & \textbf{Input} & \textbf{Output} \\
\hline
\hline
\textbf{Idea} & Generate project idea & input.md & idea.md\\
\hline
\multirow{2}{*}{\textbf{Literature}} & \multirow{2}{*}{Determine if idea is new} & input.md & \multirow{2}{*}{literature.md}\\
& & idea.md &\\
\hline
\multirow{2}{*}{\textbf{Methods}} & \multirow{2}{*}{Develop project plan} & input.md & \multirow{2}{*}{methods.md}\\
& & idea.md &\\
\hline
\multirow{3}{*}{\textbf{Analysis}} & Implement plan & input.md & \multirow{2}{*}{results.md}\\
& write and execute code & idea.md & \multirow{2}{*}{Plots}\\
& make plots & methods.md & \\
\hline
\multirow{4}{*}{\textbf{Paper}} & \multirow{4}{*}{Write paper} & input.md & \multirow{3}{*}{paper.pdf}\\
& & idea.md &\\
& & methods.md & \\
& & results.md & \\
\hline
\multirow{2}{*}{\textbf{Review}} & \multirow{2}{*}{Review paper} & (input.md) & \multirow{2}{*}{referee.md}\\
& & paper.pdf &\\
\hline
\end{tabular}
\caption{This table shows the different modules contained in \texttt{Denario} along with their tasks, inputs, and outputs. \texttt{input.md} contains a description of the data or problem of interest and is provided by the human researcher in markdown format. \texttt{idea.md} contains the project idea, \texttt{literature.md} a report on whether the idea is new, \texttt{methods.md} the research methodology, \texttt{results.md} the research analysis, Plots is a folder containing the plots and \texttt{referee.md} contains a referee report. \texttt{paper.pdf} is the scientific paper in pdf format. Files between parentheses are optional.
}
\label{tab:denario_io}
\end{table}

\begin{itemize}
\item \textbf{Idea module}. This module is in charge of generating an idea for a project. This module only requires the input text file to be run. After calling this module, \texttt{Denario} will print out the idea and save it in a markdown file called \texttt{idea.md}. The agents in charge of this module are described in Section \ref{subsec:idea}.

\item \textbf{Literature module}. This module is in charge of checking the literature to figure out if an idea is original or there are existing works that have carried it out. The input to this module is the input text together with the \texttt{idea.md} file. This module will then check the literature and and write a report on whether the idea is new or not. The report will be saved to a file called \texttt{literature.md}. The agents in charge of this module are described in Section \ref{subsec:literature}.

\item \textbf{Methodology module}. This module is in charge of generating a detailed plan to carry out the idea. To be run, it needs the input description text file and the \texttt{idea.md} file with the description of the idea. We note that \texttt{idea.md} can be generated by the previous module or provided by the user. After executing this module, \texttt{Denario} generates a detailed plan outlining the steps and methods needed for the project. \texttt{Denario} stores this generated text in a file called \texttt{methods.md}. The agents in this module are described in Section \ref{subsec:methods}

\item \textbf{Analysis module}. This module is in charge of implementing the research plan. It requires the input text, together with the \texttt{idea.md} and \texttt{methods.md} files (generated by previous modules or provided by the user). By calling this module, \texttt{Denario} will carry out the research plan, usually translating into writing and running code, making plots, etc. The output of this module is a file, \texttt{results.md}, with the outcome of the analysis, together with the plots generated. The agents behind this module are described in Section \ref{subsec:research}.

\item \textbf{Paper module}. This module is in charge of writing a scientific paper. This module requires the input text along with the \texttt{idea.md}, \texttt{methods.md}, \texttt{results.md}, and all plots. As above, these files can be the outcome of the modules or can be provided by the user. By calling this module, \texttt{Denario} writes a scientific paper and saves the .tex, .bib, and .pdf files. The agents behind this module are described in Section \ref{subsec:paper}. 

\item \textbf{Review module}. This module is in charge of reviewing a scientific paper. The input to this module is a pdf with the scientific paper along with an input description (optional). This module will review the paper and provide a report about the positive and negative aspects of the draft. The output of this module is a referee report that will be saved on a file called \texttt{referee.md}. The agents behind this module are described in Section \ref{subsec:review}. 

\end{itemize}

We emphasize that \texttt{Denario} is designed in a modular way. By calling sequentially all the above modules, \texttt{Denario} will perform end-to-end scientific research from an input text (e.g. from a description of a dataset). However, the user can also just run one (or several) of the modules to perform a specific task. For instance, given a well-defined idea and research plan (provided by the user), the user can run the analysis module to write code and create plots. We now discuss in detail the architecture and agents involved in each module of \texttt{Denario}.

\subsection{Input Text}

All modules in \texttt{Denario} accept an input text. This input text usually contains a description of the data to be analyzed or the problem to be solved. This text is later used by the modules to generate ideas, develop research plans, analyze the data, or check the literature. How well the system behaves depends critically on this text. We now outline a few guidelines on how to write the input text:

\begin{itemize}

\item If the project to carry out consists of analyzing some data, and the structure of the dataset is not well described, the analysis module will fail, as the code it will write will likely not be appropriate to read the data properly. We recommend that the user describe the data or the problem with as much detail as possible in the input text. For instance, if the project is about analyzing a tabular dataset, it is important to write what the data is about and some context. Besides, it is important to describe in detail what the different columns and rows are, and specify things like units, ranges (if available), etc.

\item If the goal is not to analyze some data but instead to solve some problem, it is very important to describe in detail the problem, connections with previous works, and some details about the steps to carry out if known.

\item It is also very important to write explicitly the location of the data (with absolute paths if possible); otherwise, the system will not be able to find the location of the data.

\item When used in end-to-end mode, \texttt{Denario} sometimes may generate papers with very few plots. One can fix this by writing in the input text to generate many plots for each step. Sometimes, the system may generate many similar plots, which will make the paper more like a research note than an actual paper. For this, the user can add some text like this: \textit{Join plots that are similar}. 

\item We have also seen cases where \texttt{Denario} fails to read the data, and to continue, it will create its own synthetic one. To avoid these situations, we usually write in the input text \textit{Do not create dummy data.}. 

\item One can also set computational constraints in the input text so that the system is aware of what resources it has when performing the analysis. Examples of this are: \textit{You have access to 1 GPU and 10 CPUs; use all the resources when training neural networks.}

\item We have observed cases where the system is not able to use a particular library (perhaps because it does not know about its documentation or because its API has recently changed). In these cases, the user can explicitly write \textit{Do not use X library to perform this analysis} to avoid that situation.

\item On the other hand, if the user has some piece of code, or a library, that may not be known to LLMs, the user can add instructions about its usage in the input text so that the system learns how to use it when analyzing the data.

\item In many occasions, it is necessary to iterate over the input text in order to guide the system in a particular direction or to improve the quality of the generated papers. In this case, the user should add details about previous experiments, or directions where the analysis is wanted, in order to achieve the desired outcome.

\item If arXiv URL's are provided in the input text description, \texttt{Denario} automatically downloads the pdf of the corresponding papers. The pdf's are OCR'ed with \texttt{mistral\_ocr\_lattest} and the resulting markdown versions are stored in the local repository. Then, the full markdown versions are sent to \texttt{o3-mini} for summarization and the summary of the papers are appended to the initial input text, providing the necessary contextual information for the subsequent LLM agents. (In the GUI this is done with the Enhance data description button, see Fig.~\ref{fig:gui}.)

\end{itemize}

We refer the reader to the example papers repository\footnote{\url{https://github.com/AstroPilot-AI/DenarioExamplePapers}} for a comprehensive list of examples, across disciplines, of input text used for different papers.

\subsection{Idea module}
\label{subsec:idea}

\begin{figure*}[ht!]
\centering
    \includegraphics[width=0.7\textwidth]{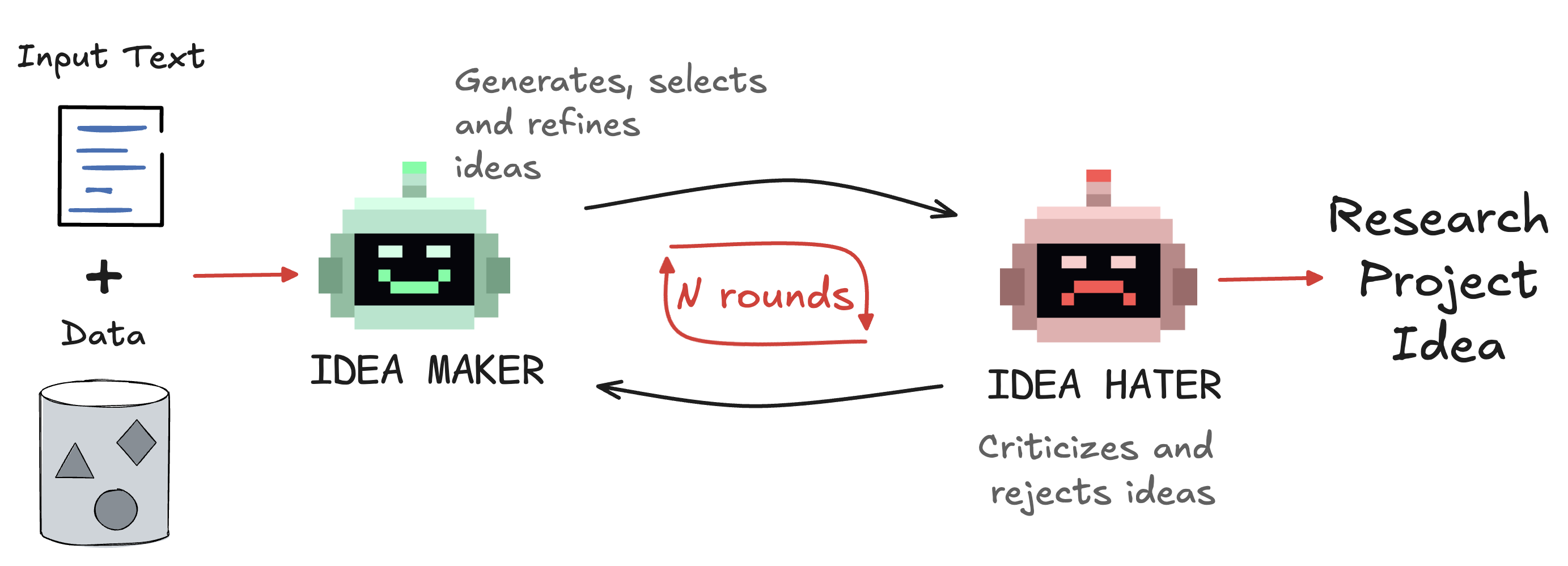}
    \caption{\label{fig:idea} The Idea module is a propose-critique block consisting  of two agents: \texttt{idea\_maker} and \texttt{idea\_hater}. The system first calls the \texttt{idea\_maker} agent, that will generate an idea given the input text. Next, the idea is sent to the \texttt{idea\_hater}, who will critique the generated idea. The \texttt{idea\_maker} will then take the input text, the generated idea, and the critique and will improve the idea. This process is then repeated several times until the final idea is generated.
    }
\end{figure*}

The goal of this module is to generate research ideas given some input text (e.g. describing some dataset). We note that recent work \cite{AI-ideas} has suggested that AI-generated ideas may be more novel (according to human evaluators), than human-generated ideas. The input to this module is the input text (see Fig. \ref{fig:general_scheme}) and the output of it is a file called \texttt{idea.md} that stores the generated idea. We now describe in detail how this module works. 

A critical part of scientific research is the generation and refinement of ideas. In any ambitious project, it's not enough to simply work on a fixed plan -- creativity and adaptability are needed to explore the best possible directions. However, creativity must be balanced by careful critique to ensure ideas are realistic, impactful, and tightly connected to the available data. To systematize this process, our pipeline includes specialized agents dedicated to idea generation (\texttt{idea\_maker}) and idea critique (\texttt{idea\_hater}).

In this pipeline, the \texttt{idea\_maker} agent generates creative and feasible research project ideas based on the input text and any feedback from the \texttt{idea\_hater} agent. Its goal is to propose diverse approaches for extracting new information, considering the current plan, task, and instructions. However, simply generating ideas is not enough -- to ensure that only the strongest, most relevant ideas move forward, each proposal must undergo careful evaluation. This is where the  \texttt{idea\_hater} agent steps in, critically reviewing each idea for feasibility, scientific value, and relevance to the dataset. By identifying weaknesses and suggesting improvements, the \texttt{idea\_hater} ensures that no idea is accepted blindly without scrutiny. Figure \ref{fig:idea} provides a schematic of the conversation between the \texttt{idea\_maker} and \texttt{idea\_hater} agents. This loop can be repeated as many times as requested.

This setup mirrors the natural human process of idea development, where individuals brainstorm concepts and refine them through feedback and discussion. Just like in human teams, iteration is key: the  \texttt{idea\_maker} improves its proposals based on the critiques received from the  \texttt{idea\_hater}, leading to a cycle of continuous refinement. This multi-agent dialogue helps distill raw ideas into well-formed, robust research directions, ultimately ensuring that the final projects are both innovative and grounded in realistic scientific practice. The concept of multi-agent systems to produce ideas has been motivated by previous works such as \cite{li2023two}, who used reinforcement learning between agents, and \cite{su2024two}, who showed that a large team of agents produced better ideas than a single agent. In its current version, \texttt{Denario} has two different implementations (see Section~\ref{sec:obs}) for this module:
\begin{itemize}
\item \textbf{Fast idea}. In this case, a single idea is generated by the \texttt{idea\_maker} agent and then improved (or completely changed) after receiving feedback from the \texttt{idea\_hater} agent. The final idea is obtained after three iterations of the two agents, followed by a further iteration from the \texttt{idea\_maker} agent. This implementation is fast (around 15 seconds with \texttt{gemini-2.0-flash}) and is written in \texttt{LangGraph}.

\item \textbf{Planning \& Control idea}. This implementation is carried out with \texttt{Cmbagent} using the planning and control strategy. The instructions passed to the \textit{Planning} agents are: 
\end{itemize}

\begin{tcolorbox}[colback=blue!5!white, colframe=blue!75!black, title=Idea Generation Planning Instructions, sharp corners=southwest, breakable]
Given these datasets and information, make a plan according to the following instructions: 
\begin{enumerate}[leftmargin=2em]
    \item Ask \texttt{idea\_maker} to generate 5 new research project ideas related to the datasets.
    \item Ask \texttt{idea\_hater}  to critique these ideas.
    \item Ask \texttt{idea\_maker}  to select and improve 2 out of the 5 research project ideas given the output of the \texttt{idea\_hater}.
    \item Ask \texttt{idea\_hater} to critique the 2 improved ideas.
    \item Ask \texttt{idea\_maker} to select the best idea from the two.
    \item Ask \texttt{idea\_maker} to report the best idea as a scientific paper title, accompanied by a 5-sentence description.
\end{enumerate}

\vspace{0.5em}
The goal of this task is to generate a research project idea based on the data of interest. 
Don't suggest performing any calculations or analyses here. The only goal of this task is to obtain the best possible project idea.
\end{tcolorbox}

During the \textit{Planning} stage, the \texttt{planner} and \texttt{plan\_reviewer} design the plan for the \texttt{idea\_maker} and \texttt{idea\_hater} agents to generate an idea, given the input text needed for the module. To illustrate this implementation, we show below the first two steps of an idea generation plan created by using as input text a description of the Perovskite database \citep{perovskite}: 

\begin{tcolorbox}[colback=blue!5!white, colframe=blue!75!black, title=Idea Generation Plan Example (First 2 Steps Only), sharp corners=southwest, breakable]

\begin{description}[leftmargin=2.5em, labelindent=0em]

\item[\textbf{Sub-task:}] Generate 5 new research project ideas related to the perovskite dataset. \\
\textbf{Agent:} \texttt{idea\_maker}\\
\textbf{Instructions:}

\begin{itemize}
    \item Review the dataset description and attributes.
    \item Propose 5 innovative and feasible research project ideas based on the dataset's scope and potential.
    \item Ensure the ideas are diverse and cover different aspects of the dataset.
    \item Consider focusing on different aspects such as material composition relationships, fabrication parameters, device architecture correlations, stability factors, and performance optimization based on the 95 attributes in the dataset.
\end{itemize}

\item[\textbf{Sub-task:}] Critique the 5 research project ideas. \\
\textbf{Agent:} \texttt{idea\_hater}\\
\textbf{Instructions:}
\begin{itemize}
    \item Evaluate the feasibility, originality, and potential impact of each idea.
    \item Identify weaknesses or limitations in the ideas.
    \item Suggest improvements or recommend discarding weak ideas.
    \item Evaluate each idea's scientific novelty and whether it leverages the unique comprehensive nature of this manually curated dataset with 42,400+ devices.
\end{itemize}
\end{description}
\end{tcolorbox}

As seen in this example, with this approach, the \texttt{idea\_maker} and \texttt{idea\_hater} agents receive explicit instructions specific to the data or problem of interest. This could potentially improve the quality of the generated ideas. However, this approach is time-consuming and incurs a significant cost. With our default LLM choices (we use \texttt{gpt-4o} and \texttt{claude-3-7-sonnet} for \texttt{idea\_maker} and \texttt{idea\_hater}, respectively), Planning takes $\sim$1 minute and costs $\sim\$0.20$, while Control takes $\sim$3 minutes and costs $\sim\$0.90$. This amounts to $\sim\$1$ and $\sim4$ minutes per generated idea, to be contrasted with the 15 seconds of the fast method that comes at no or significantly lower cost with \texttt{gemini-2.0-flash}.

The reason behind the two different implementations of this module is to provide the user more flexibility when using \texttt{Denario}. For instance, a new user may prefer to use the fast idea implementation to get results quicker, while in full production mode, the user may prefer a more robust approach by using the curated idea generation.

As stated above, this module will generate an idea and store it in a file called \texttt{idea.md}. This file is later used as input for other modules. We emphasize that the user can use the idea module to generate an idea, or can provide the idea directly by writing it to the \texttt{idea.md} file.

\subsection{Literature module}
\label{subsec:literature}

The goal of the literature module is to search the literature to determine whether an idea, in combination with the input text, is new or not. This module takes as input the input text together with the idea (usually in the form of the \texttt{idea.md} file). The module then searches the literature for papers similar to the proposed idea and evaluates whether the project idea is new or not. The input text is taken into consideration when evaluating the idea. We do this because in some research areas, standard techniques applied to new datasets may be seen as novel research. For instance, even if graph neural networks are very standard in deep learning, they may not have been applied to cosmological datasets, and this may be seen as novel in the cosmology research area. The module returns a file, \texttt{literature.md}, that contains a report on whether the idea is new or not. Denario provides two different implementations of this module:

\begin{figure*}[ht!]
\centering
    \includegraphics[width=0.5\textwidth]{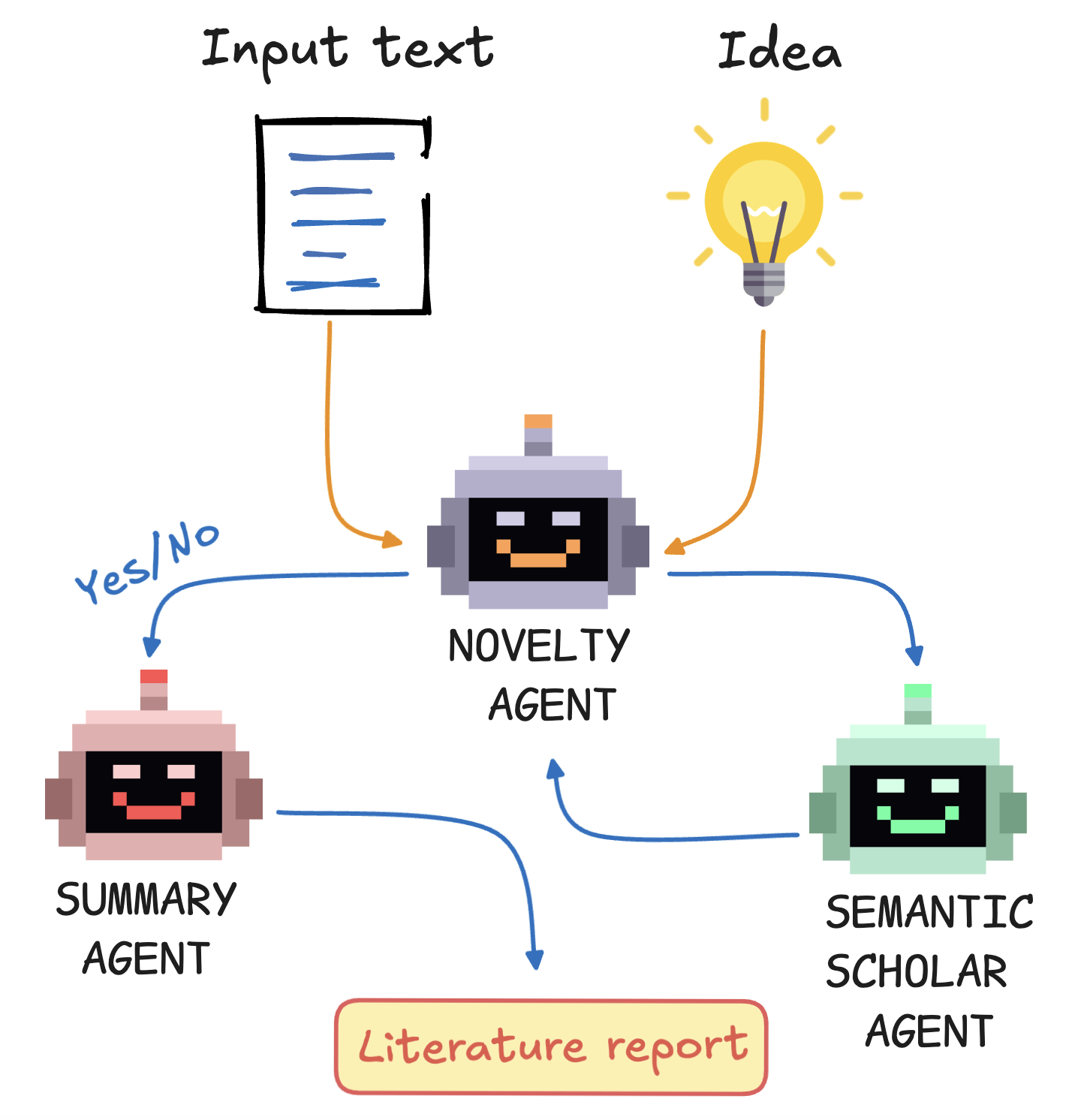}
    \caption{\label{fig:literature} This diagram shows the semantic scholar implementation of the literature module. The module takes as input the input text and the idea. These files are then passed to a \texttt{novelty agent} whose task is to determine if the idea is new or not. This agent can output 1) idea new, 2) idea not new, or 3) query. The agent will choose the later if it does not enough information to make a yes/no decision. In that case, it will generate a query that will be sent to the \texttt{semantic scholar agent} who will search the literature with it. This agent will then retrieve a series of papers (including titles and abstracts) and will send this information back to the \texttt{novelty agent}. This process is repeated several times until the relevant literature is well sampled. If no relevant papers are found after a number of iterations, the novelty agent will conclude that the idea is new. Once the \texttt{novelty agent} has determined that the idea is new/not new, all the papers found, the queries, and the responses will be sent to a \texttt{summary agent} that will write a report stating why the idea is new (or not new) and the most relevant papers.}
\end{figure*}

\begin{itemize}
\item \textbf{Semantic scholar}. In this implementation, carried out with \texttt{LangGraph}, we follow the workflow introduced in \citep{AI_scientist}. The module consist of two agents that collaborate to determine whether an idea, together with the associated data description, is new or has already been carried out. First, the \texttt{novelty agent} is presented with the idea, the data description, and with titles and abstracts of papers already found. The goal of this agent is then to determine whether, with the given information, the paper is new, is not new, or whether it does not have enough information to reach a conclusion. If it determines that the paper is new, or not new, all the information is sent to another agent to summarize the results and to write a report. This report will be placed on a file called  \texttt{literature.md}. If the \texttt{novelty agent} determines that it does not have enough information, then it can call the \texttt{semantic scholar agent} which can make a query using Semantic Scholar\footnote{\url{https://www.semanticscholar.org/}}. This other agent will then make an API call to Semantic Scholar, using the query sent by the \texttt{novelty agent}, and obtain papers (titles, abstracts, URLs, etc) that will be sent back to the \texttt{novelty agent} for evaluation. This procedure is repeated several times up to a maximum number, where the module will determine that the idea is new as it has not found relevant papers after multiple iterations. We show a scheme about the workflow of this implementation in Fig. \ref{fig:literature}.

\item \textbf{FuturHouse's Owl}. For this implementation, we just make use of the \texttt{Owl} agent developed by FutureHouse.\footnote{\url{https://www.futurehouse.org/}} This is a specialized agent designed to search the literature and provide an answer to the question: Has anyone implemented this idea before? We refer the reader to the FuturHouse documentation\footnote{\url{https://futurehouse.gitbook.io/futurehouse-cookbook/futurehouse-client}} for details about this agent.
\end{itemize}

Denario contains both implementations to allow the user to choose among the two, or perhaps better, to run both of them to get to "independent" opinions. This module will generate the literature report, located in the \texttt{literature.md} file. We note that in the current implementation, this file is not used by any other module. We do this intentionally so that a human evaluator decides whether the idea is truly new/original or not.

\subsection{Methods module}
\label{subsec:methods}

\begin{figure*}[ht!]
\centering
    \includegraphics[width=0.7\textwidth]{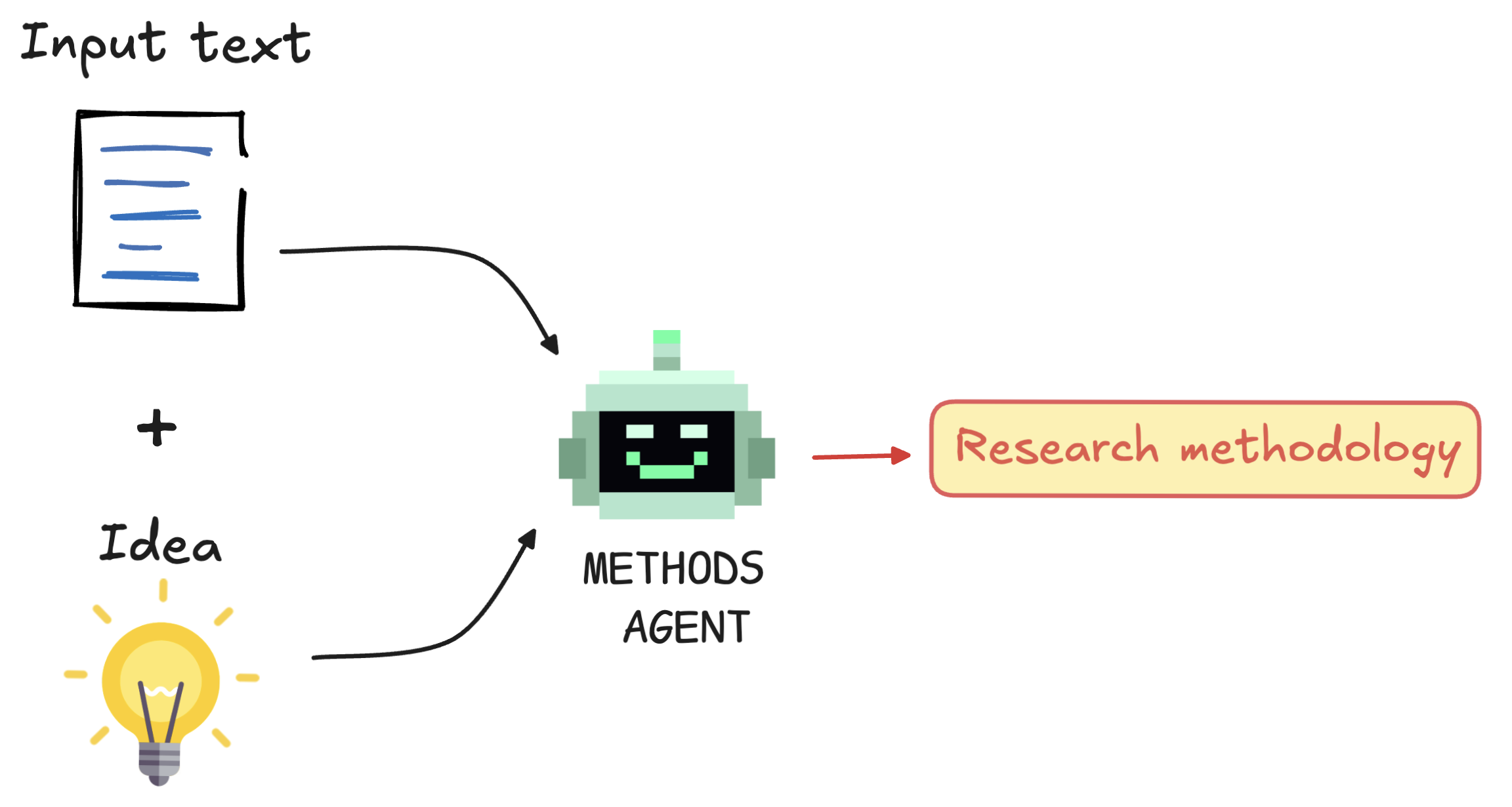}
    \caption{\label{fig:method} The methods module.
    }
\end{figure*}

The methods module is responsible for producing a detailed plan, or methodology, to carry out the project idea. This module takes the input text and the project idea (usually located in the \texttt{idea.md} file), as input. The module's outcome is a detailed research plan designed to carry out the idea. This output is saved in a file called \texttt{methods.md}. The current version of \texttt{Denario} contains two different implementations for this module:
\begin{itemize}

\item \textbf{Fast methods}. This is a simple implementation that takes the input text and the \texttt{idea.md} file and generates a methodology. It is implemented in \texttt{LangGraph} and designed to be run with fast models like \texttt{gemini-2.5-flash} (although it supports most LLMs). The prompt used in this case is this:

\begin{tcolorbox}[colback=blue!5!white, colframe=blue!75!black, title=Fast methods prompt, sharp corners=southwest, breakable]
You are provided with an input text and an idea for a scientific paper. Your task is to think about the methods to use in order to carry it out. Follow these instructions:
\begin{itemize}
\item Generate a detailed description of the methodology that will be used to perform the research project.
\item The description should clearly outline the steps, techniques, and rationale derived from the exploratory data analysis (EDA).
\item The focus should be strictly on the methods and workflow for this specific project to be performed. \textbf{Do not include} any discussion of future directions, future work, project extensions, or limitations.
\item The description should be written as if it were a senior researcher explaining to her research assistant how to perform the research necessary for this project.
\item Just provide the methods, do not add a sentence at the beginning saying showing your thinking process
\end{itemize}

Problem or data description:

\{Input text\}

\vspace{0.5cm}
Idea:

\{project idea\}

\vspace{0.5cm}
Respond with the methods you have generated.
\end{tcolorbox}

\item \textbf{Planning \& Control methods}. This mode
uses the \textit{Planning \& Control} strategy of \texttt{Cmbagent}, with the \texttt{researcher} agent in charge of the subtasks. The models chosen by default for the \texttt{researcher} agent in this mode are \texttt{gpt-4.1} or \texttt{gemini-2.5-pro}. The instructions for planning are these: 

\end{itemize}

\begin{tcolorbox}[colback=blue!5!white, colframe=blue!75!black, title=Methods Module; Planning Instructions, sharp corners=southwest, breakable]

Given these datasets, and information on the features and project idea, we want to design a methodology to implement this idea. The goal of the task is to write a plan that will be used to generate a detailed description of the methodology that will be used to perform the research project.

\begin{itemize}
\item Start by requesting the \texttt{researcher} to provide reasoning relevant to the given project idea.
\item Clarify the specific hypotheses, assumptions, or questions that should be investigated.
\item  This can be done in multiple steps. 
\item The focus should be strictly on the methods and workflow for this specific project to be performed. \textbf{Do not include} any discussion of future directions, future work, project extensions, or limitations.
\item The description should be written as if it were a senior researcher explaining to her research assistant how to perform the research necessary for this project.
\end{itemize}

The final step of the plan must be entirely dedicated to writing the full Methodology description. The only agent involved in this workflow is the \texttt{researcher}. In this task, we do not perform any calculations or analyses, only outline the methodology. 

\end{tcolorbox}

For this task, the \texttt{researcher} agent of \texttt{Cmbagent} is further instructed as follows:

\begin{tcolorbox}[colback=blue!5!white, colframe=blue!75!black, title=Methods Generation Appended Researcher Instructions, sharp corners=southwest, breakable]

Given this information, we want to design a methodology to implement this idea.
The goal of the task is to develop a detailed methodology that will be used to carry out the research project.
\begin{itemize}

\item  You should focus on the methods for this specific project to be performed. \textbf{Do not include} any discussion of future directions, future work, project extensions, or limitations.
\item  The methodology description should be written as if it were a senior researcher explaining to her research assistant how to perform the project. 
    
\end{itemize}

The designed methodology should focus on describing the research and analysis that will be performed. The full methodology description should be written in markdown format and include all the details of the designed methodology.
It should be roughly 500 words long.

\end{tcolorbox}

These instructions are appended to the default instructions of the \texttt{researcher} agent. We further constrain $n_\mathrm{steps}=4$ (maximum 4 steps in the plan). With the default models, at the time of writing this paper, \textit{Planning} typically costs around \$0.20 and takes around 1 minute, while \textit{Control} takes around 1.5 minutes per step at a cost of around \$0.10 per step. Thus, with \textit{Planning \& Control}, the total cost of designing the methodology is around \$0.50 and takes around 4 minutes. This is to be contrasted with the Fast mode that takes around 15 seconds at no cost to produce the methodology with \texttt{gemini-2.0-flash}. We note that using \texttt{gemini-2.5-pro} for the fast implementation has a small cost (around $\sim\$0.10$) and will take around 90 seconds, but typically will produce a much more detailed plan to help run the analysis pipeline.

As in the case of the idea module, we provide two different implementations to the user to enhance their range of options when executing this module. We note that the curated implementation may not necessarily yield a better research plan, while its cost and execution time will be significantly higher than the fast option when run with \texttt{gemini-2.5-pro}. However, we recommend the user try both options for the particular case study.

\subsection{Analysis module}
\label{subsec:research}

The goal of this module is to implement the analysis outlined in the methodology. This is done with \href{https://github.com/CMBAgents/cmbagent}{Cmbagent} \cite{Laverick:2024fyh,cmbagent2,CMBAGENT_2025}.  
The inputs to this module are the input text, the project idea, and the methodology (either created by earlier modules or supplied directly). The output of the module is 1) a report on the results obtained, located in a file called \texttt{results.md}, and 2) a folder called \texttt{Plots} that contains all plots generated. In the current version of \texttt{Denario}, the analysis module relies exclusively on \texttt{Cmbagent} and its \textit{Planning \& Control} strategy for ``Deep Research" orchestration, similar in nature to the workflows of agentic tools like Claude Code or Cursor agent mode. See Fig. \ref{fig:pc_cmbagent} for an illustration of this strategy.

For the \textit{Planning} phase, we append the instructions reported hereafter to the planning agents. 

\begin{tcolorbox}[colback=blue!5!white, colframe=blue!75!black, title=Analysis Generation Planning Instructions, sharp corners=southwest, breakable]
        \{research\_idea\}\\
\vspace{0.2em}\\
        \{methodology\}\\
\vspace{0.2em}\\
        Given these datasets, project idea and methodology, we want to perform the project analysis and generate the results, plots and insights.\\
        
        The goal is to perform the in-depth research and analysis. \\

        The plan must strictly involve only the following agents: \{involved\_agents\_str\}.\\
        
        The goal here is to do the in-depth research and analysis, not an exploratory data analysis.\\

        The final step of the plan, carried out by the researcher agent, must be entirely dedicated to writing the full Results section of the paper or report. If this research project involves code implementation, this final step should report on all the qualitative and quantitative results, interpretations of the plots and key statistics, and references to the plots generated in the previous steps.
        The final result report will be what will be passed on to the paper writer agents, so all relevant information must be included in the final report (everything else will be discarded).
        
\end{tcolorbox}

As we store the full idea and methodology in the context variable dictionary, the planning instructions contain these explicitly. The context variable \texttt{involved\_agents\_str} is a string with the names of the agents involved for the analysis generation, which by default are the \texttt{engineer} and \texttt{researcher} agents of \texttt{Cmbagent}. We append the following instructions to these two agents. 

\begin{tcolorbox}[colback=blue!5!white, colframe=blue!75!black, title=Analysis Generation Appended Engineer Instructions, sharp corners=southwest, breakable]
        \{research\_idea\}\\
\vspace{0.2em}\\
        \{methodology\}\\
\vspace{0.2em}\\
      Given these datasets, and information on the features and project idea and methodology, we want to perform the project analysis and generate the results, plots and key statistics.
        The goal is to perform in-depth research and analysis. This means that you must generate the results, plots, and key statistics.\\

        Warnings for computing and plotting:\newline
        - make sure dynamical ranges are well captured (carefully adjust the limits, binning, and log or linear axes scales, for each feature).\\

        For histograms (if needed):\newline
        -Use log-scale for features with values spanning several orders of magnitude.\\

        \textbf{GENERAL IMPORTANT INSTRUCTIONS}: You must print out in the console ALL the quantitative information that you think the researcher will need to interpret the results. (The researcher does not have access to saved data files, only to what you print out!)
        Remember that the \texttt{researcher} agent can not load information from files, so you must print ALL necessary info in the console (without truncation). For this, it may be necessary to change \texttt{Pandas} (if using it) display options.
\end{tcolorbox}

The \texttt{engineer} agent is in charge of writing Python code analysis pipelines. The \texttt{engineer} agent call triggers a \textit{nested chat} (similar to \textit{self-debug} \cite{chen2023teachinglargelanguagemodels}), of which only the output is kept. This allows for long sessions without hitting LLM provider API token limits. A failed code execution may trigger different transitions. If the number of failed executions is lower than $n_\mathrm{fails}$, the module would generally transit back to the \texttt{engineer} agent for another attempt. If the failure is caused by a missing Python package, the system will send a request to an \texttt{installer} agent, which runs a bash \texttt{pip\,\,install} command before attempting the code execution again. If the number of failed executions equals $n_\mathrm{fails}$, a transition to the \texttt{terminator} agent is triggered and the session ends.

Given the planning instructions reported above, the \texttt{engineer} agent is generally in charge of the bulk of the analysis, while the \texttt{researcher} agent is called in the final step of the workflow to report on the results. For \texttt{Denario}, we append the \texttt{researcher}' system message with the following instructions:

\begin{tcolorbox}[colback=blue!5!white, colframe=blue!75!black, title=Analysis Generation Appended Researcher Instructions, sharp corners=southwest, breakable]
        \{research\_idea\}\\
\vspace{0.2em}\\
        \{methodology\}\\
\vspace{0.2em}\\
        At the end of the session, your task is to generate a detailed/extensive \textbf{discussion} and \textbf{interpretation} of the results. 
        If quantitative results were derived you should provide interpretations of the plots and interpretations of the key statistics, including reporting meaningful quantitative results, tables and references to material previously generated in the session.
        The results should be reported in full (not a summary) and in academic style. The results report/section should be around 2000 words.

        The final result report will be what will be passed on to the paper writer agents, so all relevant information must be included in the final report (everything else will be discarded).
\end{tcolorbox}
 Each step during control, with \texttt{gemini-2.5-pro} or \texttt{gpt-4.1} as engineer and researcher, costs around \$0.30, amounting to around \$2 for a plan in 6 steps. The time needed to run this module depends critically on the amount of data, the technique used to analyze it, and the available computational resources. For simple analyses with less than 1 Gigabyte of data, the typical time is $\sim$30 minutes on a personal computer.
 The user can (and should) specify which computational resources are ready to be committed to the analysis.

\subsection{Paper module}
\label{subsec:paper}

\begin{figure*}[ht!]
    \centering
    \includegraphics[width=0.99\textwidth]{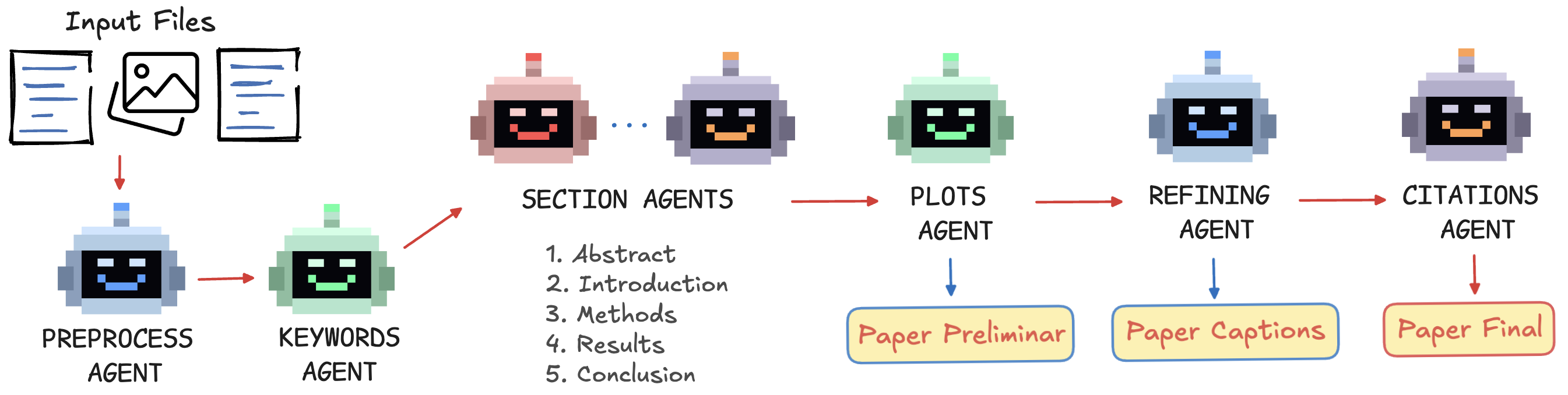}
    \caption{\label{fig: paper_writing} This scheme shows the workflow of the paper writing module. Each node represents an agent (or a group of agents) in charge of a task. The preprocess agent is in charge of reading the input files and performing some basic operations, such as identifying duplicated plots. Later, a series of agents will identify keywords and write different sections of the paper draft. Next, an agent will process the plots, generating captions for them and outputting a preliminary version of the paper draft. Next, an agent will refine the results section to improve references to the plots and to polish the writing on it, generating a second version of the paper draft. Next, an agent will place citations in different sections of the paper, producing a third and fourth final version of the paper draft.
    }
\end{figure*}

This module is in charge of writing a scientific paper or report. In its current form, it should be seen as a paper draft more than a publication-ready paper. This module is entirely written in \texttt{LangGraph} (but for the keyword search, which uses \texttt{Cmbagent}). The module takes as input the input text, the project idea, the project methodology, and the project results. As for the other modules, these files can be either provided by the user or generated by previous modules. The output of this module is a set of LaTeX files together with associated pdfs. A scheme showing the workflow of this subsystem is shown in Fig. \ref{fig: paper_writing}. We note that the user has multiple options available when running this module, such as the LLM to use, the journal of the paper, and whether the paper includes citations or not\footnote{For research notes or similar, the user may not find it useful to spend time and money adding citations. Given the cost, we tend to add citations only to the papers that have been reviewed and filtered by human experts.}.

The process starts with an \texttt{preprocess} agent reading the input files, which will be kept in memory until it has finished writing the entire paper. We note that it is easy to include more files in case the paper writing requires more information. For instance, the user may provide an additional file with details about the data, references, literature report, etc. The \texttt{preprocess} agent performs other tasks, such as counting the number of plots, identifying and removing repeating plots, and creating temporary files.

Next, given the texts read above, another agent will select a set of keywords that characterize the paper (see Section \ref{subsec:keywords}). After that, another agent will choose a title for the paper and write the abstract. Next, several agents will write the paper's different sections: 1) Introduction, 2) Methods, 3) Results, and 4) Conclusions. For each section, an agent is passed the input texts and asked to write the assigned section for the paper. We note that the sections are written sequentially, with each section passed to the next agent to improve the coherence throughout the paper. We perform self-reflections for some sections to improve the quality of the generated text.

Once these sections are written, another agent is asked to take all the plots available and generate captions for them. This is achieved by using the multimodal capabilities of the underlying LLM. We note that the captions are generated taking into account the input files to improve the context and coherence of the captions. Next, another agent is tasked with inserting the figures and their generated captions into the results section. This is done in batches of 7 figures\footnote{This is in principle a hyperparameter. We have noticed that with current LLMs, including more than 10 figures tends to result in errors, omissions, etc}, as the LLM tends to forget inserting some figures when there are many figures in the prompt. At this point, the system will generate a first LaTeX file, compile it, and generate the corresponding pdf.

After this, another agent is asked to rewrite and polish the results section so that the text describes and refers to the added figures. Next, version 2 of the paper, with the polished results section, will be generated. We do this because we have observed the module failing at this point and we use these versions as checkpoints in case we need to rerun the module.

This module can be run with the option to add citations to the paper. If run in that mode, the text inside the different sections is passed through a Perplexity agent in order to get citations (with arXiv IDs) to the relevant sentences. We use the Perplexity API with the \texttt{sonar-reasoning-pro} model for this. A script is then used to fetch the BibTeX information from arXiv, which is appended to the LaTeX. At this point, version 3 of the paper is generated and saved to file. Finally, another agent is asked to make a final pass through the different sections of the paper in order to improve the clarity of the text and to fix LaTeX errors. After this, version 4 of the paper is produced. 

We note that while multiple agents are always checking for LaTeX errors, and some agents are designed to take the LaTeX errors and fix the text, we find that mistakes can still occur, and that may cause problems with the document compilation. E.g. misplaced citations, tables not displayed properly, problems with symbols like \_ and \&, and \%. For that reason, the paper writing module does not produce a single version of the paper but four different ones, as progress is made. If a problem arises, and the agents themselves are not able to fix it, then, a human can try to fix it manually from of these checkpoints. 

\subsubsection{Keyword Search module}
\label{subsec:keywords}

\texttt{Cmbagent} provides a method to get the keywords which best characterize a given input text, such as the abstract of a paper. The \texttt{Denario} Paper module makes use of this functionality to find the correspondent keywords of the project by passing as input text the idea and methodology. There are three keyword lists that can be searched by this method: 
\begin{itemize}
    \item \textbf{UNESCO Nomenclature}. This taxonomy has three levels for classifying science and technology domains, subfields and specific areas. We do the keyword search in four steps. First, the keyword agent is provided with the list of domains (2-digit codes) and is asked to select at most 3 domains. Second, for each selected domain, the keyword agent is asked to select at most 3 subfields. Third, for each subfields, the keyword agent is asked to select  3 specific areas. Domains, subfields and specific areas are aggregated into one list and in the last step the agent is asked to select the $N$ most relevant keywords where $N$ is the number of keywords requested by the user. The original UNESCO document is publicly available.\footnote{\url{https://unesdoc.unesco.org/ark:/48223/pf0000082946}} In principle, this list of keywords is broad enough to be used in all scientific domains. Nevertheless, it is from 1988 and recent technical vocabulary is sometimes lacking. 

    \item \textbf{AAAI Keywords}. Here, the keyword agent is provided with the full list of AAAI (The Association for the Advancement of Artificial Intelligence) keywords and asked to select the $N$ most relevant keywords where $N$ is the number of keywords requested by the user. The AAAI list of keywords is publicly available.\footnote{\url{https://aaai.org/conference/aaai/aaai-26/keywords/}} This list is relevant for AI-focused research and is updated yearly.
    \item \textbf{AAS Keywords}. The keyword agent is provided with the full AAS (American Astronomical Society) list of keywords and asked to select the $N$ most relevant keywords where $N$ is the number of keywords requested by the user. This list should only be used for astrophysics related papers, and can be consulted in the AAS journal page.\footnote{\url{https://journals.aas.org/keywords-2013/}}
\end{itemize}

\subsection{Review module}
\label{subsec:review}

This module is in charge of reviewing the generated paper and providing a report about what is good, interesting, and innovative in the paper, but also about the bad things, flaws, etc. The input to it is a pdf document together with the input text (optional). The module will then break down the paper into a set of images, one image for each page, that will be passed to a multi-modal agent along with the input text (usually \texttt{gemini-2.5-pro}). The agent will then review the paper and write a report about it, highlighting the good but also the bad things. The report will be placed in a file called \texttt{referee.md}.

\begin{figure*}[ht!]
\centering
\includegraphics[width=0.99\textwidth]{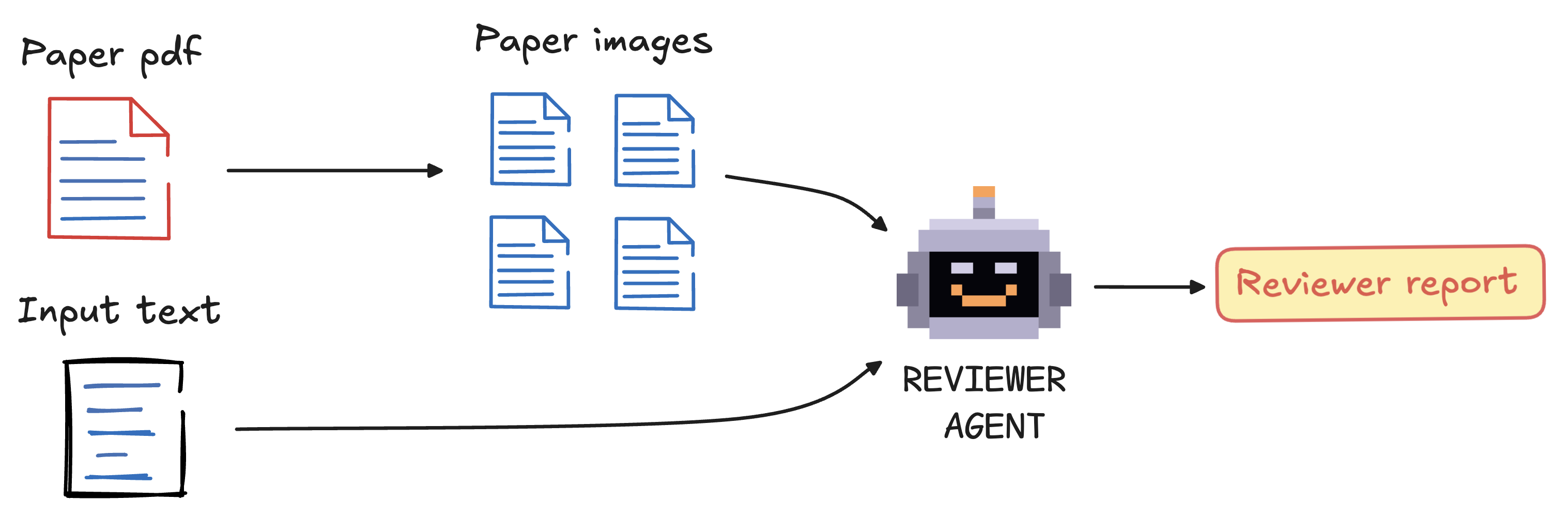}
    \caption{\label{fig:review} This diagram shows how the reviewer module works. The input to the module is the input text together with the PDF of the document to be reviewed. The first step is to transform the pdf document into png images; each image corresponds to a different page of the paper. Next, the images and the input text are passed to a reviewer agent in charge of understanding and evaluating the document. The outcome of the agent is a report with the positive and negative aspects of the document.}
\end{figure*}

The purpose of this module is to help humans review and filter papers. While some of the recommendations we have seen from executing this module are related to ways to describe and connect things, the module is also able to provide much more in-depth suggestions, such as logical flaws or underlying assumptions that invalidate reasoning or conclusions. The report generated by this module can be used to improve the quality of the generated papers via human intervention or automatic agent refinement. The prompt used for this module is found below.

\begin{tcolorbox}[colback=blue!5!white, colframe=blue!75!black, title=Reviewer prompt, sharp corners=southwest, breakable]
\{ "type":"text", "text": """You are a scientific referee. Below, you can find a scientific paper written in latex. Your task is to read and understand the paper. Next write a detailed report about the good/interesting aspects of the paper but also bad things, failures...etc. For the bad things, please provide comments on what would be needed to do in order to improve it. Note that you may be reviewing an AI-generated paper, so the author may not be human, and keywords may be missing. No need to mention those. \\

- Find all flaws in the paper \\
- Find things that may not be done correctly\\
- Identify places where further revisions would make the paper better\\
- Check carefully that there is enough evidence in the paper to support the conclusions \\
- If the results are not good, reason whether this is a surprising thing or just it used the wrong strategy and failed. If the latter, the paper should be considered bad.\\ 

Try to judge whether the paper will be worth a publication or not. Give a score from 0 (a very bad paper) to 9 (an amazing paper). For bad papers, give a low score.\\

**Respond in exactly this format**:\\
\begin{lstlisting}[language = Python]
\begin{REVIEW}
<REVIEW>
\end{REVIEW}
\end{lstlisting}
In $\langle$REVIEW$\rangle$, put your report.\},\\
\{ "type": "image\_url",\\
        "image\_url": \{"url": f"data:image/png;base64,{image\_b64}"\}\}\\
        for image\_b64 in state['referee']['images']
\end{tcolorbox}

\section{Usage}
\label{sec:deployment}

In its current form, \texttt{Denario} can be run in two different ways:

\begin{itemize}
\item Python API
\item Graphical User Interface (GUI)
\end{itemize}

The Python API can be run in Python scripts or Jupyter notebooks, while the GUI can be run either locally or through a web application. In the following subsections we detail the usage of each case.

\subsection{Python API}

The main entrypoint of Denario is its Python API, which allows the user to call the different modules. For that, the user firstly needs to install the software locally. \texttt{Denario} is publicly available on GitHub\footnote{\url{https://github.com/AstroPilot-AI/Denario}} and pypi,\footnote{\url{https://pypi.org/project/denario/}} and can be easily installed by running the command: \texttt{pip install denario}. We refer the reader to the documentation\footnote{\url{https://denario.readthedocs.io}} for further details on how to install the package. A typical Python script to perform end-to-end research is this:

\begin{lstlisting}[language=Python]
from denario import Denario, Journal

# Initiate Denario by setting the working directory
den = Denario(project_dir="project_dir")

# Set the input text with data description
input_text = """
Analyze the experimental data stored in /path/to/data.csv using sklearn and pandas.
This data includes time-series measurements from a particle detector.
"""
den.set_data_description(input_text)

# Generate a research idea from the input text
den.get_idea()

# check whether the idea is new or not
den.check_idea()

# Generate a research plan to carry out the idea
den.get_method()

# Follow the research plan, write and execute code, make plots, and summarize the results
den.get_results()

# Write a paper for the APS journal
den.get_paper(journal=Journal.APS)

# Review the paper
den.referee()
\end{lstlisting}

The Python API also allows to set research idea and methods provided by the user:

\begin{lstlisting}[language=Python]
# Set the idea given by a text file in the project folder
den.set_idea()

# or providing an input string
den.set_idea("Write a research idea here")

# Similarly, set the methods given by a text file in the project folder (or providing an input string)
den.set_method()
\end{lstlisting}

It is possible to select between the different backends, \texttt{Cmbagent} and the default \textit{fast} method based on \texttt{LangGraph} for idea and method generation:
\begin{lstlisting}[language=Python]
# Generate an idea using the fast method based on the LangGraph implementation (default option)
den.get_idea(mode="fast")

# Generate an idea using the Cmbagent backend
den.get_idea(mode="cmbagent")
\end{lstlisting}

The code above can be executed in several ways. For example, the lines can be saved in a Python script and run using \texttt{python script.py}. Alternatively, the code can be executed interactively in environments such as a Jupyter notebook or Google Colab. For more details on the implementation and syntax of the Python API, please refer to the \texttt{Denario} documentation\footnote{\url{https://denario.readthedocs.io}}.

\subsection{Graphical User Interface}

\subsubsection{Launch locally}

\begin{figure*}[ht!]
\centering
\includegraphics[width=0.99\textwidth]{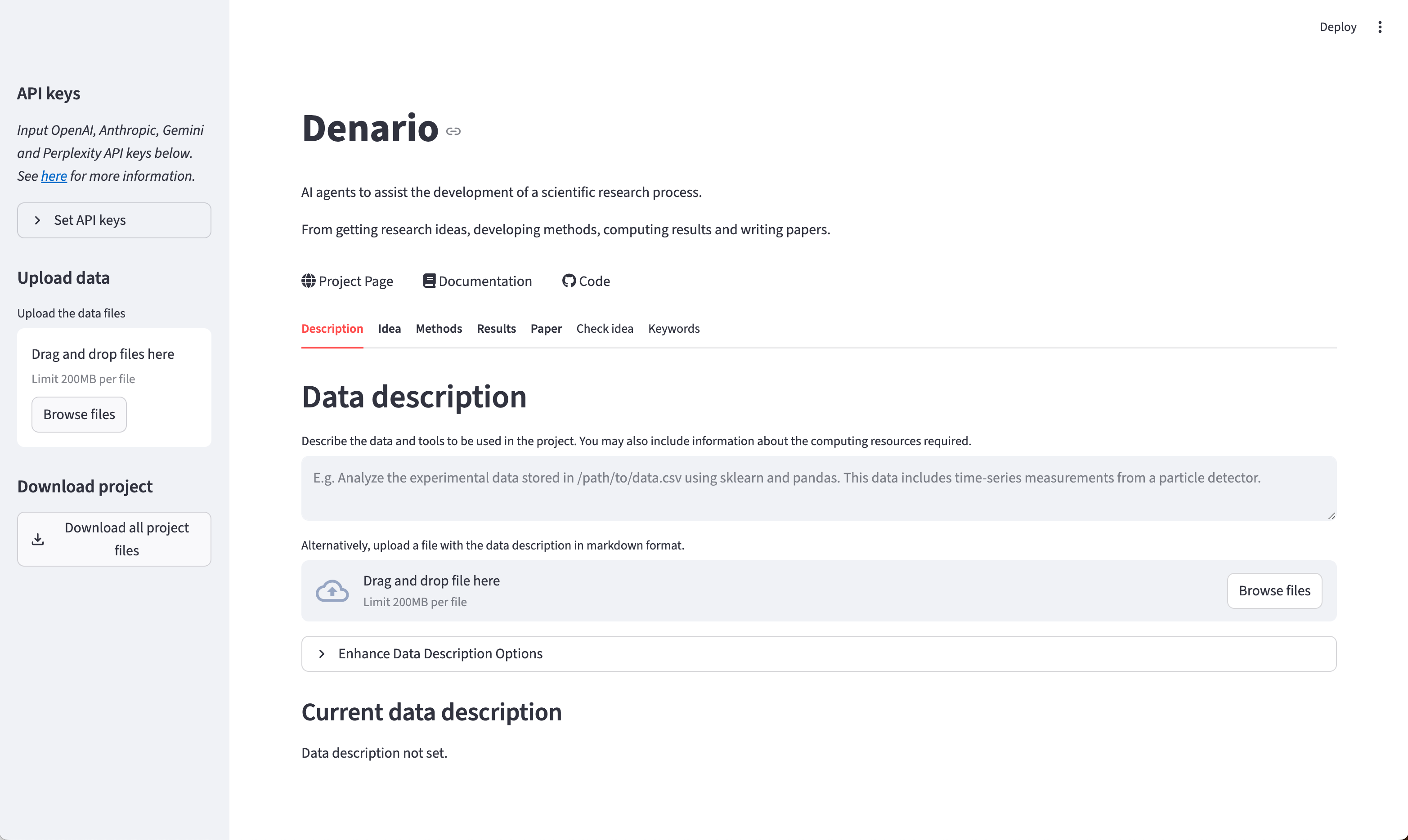}
    \caption{\label{fig:GUI} This figure shows the Graphical User Interface (GUI) of \texttt{{Denario}}. In the upper-left part, the user can set the API keys for the LLMs. Below that part, the user can upload data and download the files generated by \texttt{Denario}. In the central part, the user can choose which module to run and can tune the options available for it (e.g. LLM model). If the user wants to perform end-to-end research, she/he will have to run the different modules sequentially.}\label{fig:gui}
\end{figure*}

\texttt{Denario} also provides a Graphical User Interface (GUI) to facilitate its usage, which is shown in Fig. \ref{fig:GUI}. It can be installed from pypi with the extension \texttt{pip install denario[app]}. The GUI can be then launched running \texttt{denario run}, which will open a tab in the browser with the application. Its repository is also publicly available at GitHub\footnote{\url{https://github.com/AstroPilot-AI/DenarioApp}}.

In the GUI, the user can set the API keys for the LLM models to use, upload the data to be analyzed and other input files, and download the files generated by \texttt{Denario}. The input text can be passed either by typing it directly into the GUI, or by uploading a file with the text. 

The GUI includes several tabs for each of the Denario modules, where the user can choose which one to run. It is possible to provide input files too through these tabs. For instance, if the user wants to generate a research methodology, but wants to provide the idea to carry out, he/she should first select the Idea tab and upload the research idea there. Next, the user should go to the Methods tabs and run the module there. After running it, \texttt{Denario} will generate the research methodology that can be used for other modules, such as the analysis one. The required inputs and the outputs generated by each module can be found in Table \ref{tab:denario_io}.

\subsubsection{Web application}

The GUI discussed above is meant to be used locally, e.g., on a personal computer. The advantage is that the user can leverage local resources such as a GPU or multiple CPUs, while a potential barrier is that the user needs to install \texttt{Denario} and its dependencies locally. To circumvent this and to provide a quick way to experiment with \texttt{Denario}, we also provide users with the option to run it as a web application. In this case, no installation is needed and the user can start using \texttt{Denario} immediately. On the other hand, the free resources available to the user (e.g., data storage, CPUs, GPUs) may be limited. We currently provide two ways to access it online:

\begin{itemize}

\item \textbf{HuggingFace Spaces app}. A demo of \texttt{Denario} is hosted as a web app on HuggingFace Spaces\footnote{\url{https://huggingface.co/spaces/astropilot-ai/Denario}}. The interface is essentially the same as the local GUI (see Fig.~\ref{fig:GUI}).

\item \textbf{Cloud-hosted app}. \texttt{Denario} can also be deployed on general cloud platforms, allowing users to access and run it directly through a browser. Please refer to the \texttt{Denario} documentation for up-to-date details on the available deployment options.

\end{itemize}

\section{End-to-end research examples} 
\label{sec:examples}

In this section, we illustrate the capabilities of \texttt{Denario} by showing and discussing multiple papers fully generated by it. To demonstrate the versatility of the system, we show papers generated on a very diverse set of disciplines, including astrophysics, biology, biophysics, biomedical informatics, chemistry, material science, mathematical physics, medicine, neuroscience, and planetary science. Furthermore, \texttt{Denario} can also be used for interdisciplinary research. We illustrate this by showing a paper that applies techniques from deep learning and quantum physics to cosmological data in Sec. \ref{sec:qitt}.

Most of the generated papers have little to none human intervention in the generation process, besides providing and adjusting the input prompts and data. Some of them have been tested varying the input prompt with different levels of detail. For instance, in the paper discussed in Sec. \ref{sec:bh} we showcase a \textit{human-in-the-loop} example, where the prompt is iteratively improved using \texttt{Cmbagent} and human feedback.

In each of the subsections below, we provide details about the data used, the input text, the generated paper and briefly discuss the positive and negative aspects of the generated paper. The reader can find all generated files, plots, text, etc for all papers below and many more at the example papers repository\footnote{\url{https://github.com/AstroPilot-AI/DenarioExamplePapers}}.

\subsection{Mathematical Physics. High-level analysis and abstraction of complex structures: the physics of non-linear PDEs}

A central challenge of research in physics lies in interpreting complex results and reducing them to a framework from which general principles can be drawn. It is this capacity for abstraction that enables scientific advancement. The problem we posed \texttt{Denario} to solve is the following: we would like to understand the structure of the solutions to the Burgers equation in fluid dynamics for different values of the viscosity $\nu$ and initial conditions. In 1-dimension, the Burgers equation is:

\begin{equation}
    \frac{\partial u}{\partial t} + u \frac{\partial u}{\partial x} = \nu \frac{\partial^2 u}{\partial x^2}
\end{equation}

\subsubsection{Background: PINNs}

Our tool to solve the Burgers equation is a neural network algorithm called PINNs (physics-informed neural networks) in its implementation in the package {\tt neurodifeqq\footnote{\url{https://github.com/NeuroDiffGym/neurodiffeq}}}. The PINN will find a space of solutions for different values of the viscosity and initial conditions. This creates a so-called latent space.

Physics-Informed Neural Networks (PINNs), first introduced in the works of Dissanayake and Phan-Thien \cite{Dissanayake1994} and Lagaris et al. \cite{Lagaris2000}, have emerged as a powerful
tool for solving partial and ordinary differential equations (PDEs/ODEs). Raissi
et al. \cite{Raissi2019} proved this approach to be valid on a wide variety of
challenging physical problems, while other researchers have made significant advancements in applying neural networks to PDEs and ODEs, including notable examples
such as \cite{Cai2019,Mattheakis2020,Sirignano2018}. In this paradigm, one typically trains a distinct neural network for each
unknown function in the governing equations. The training process involves minimizing a loss function that encodes the squared residuals of these differential equations,
thereby embedding the physics directly into the learning objective.

In the PINNs setup, we use one NN for every unknown function appearing in the differential equations. We call these objects $\psi^{NN}(x^\mu)$ and they represent the approximate solution given by the NNs for different points along the independent variable and possibly different values of parameter or boundary conditions (all of them are represented by $x^\mu$). The training process involves minimizing the loss function, typically defined as the residuals of the differential equations squared, summed across all the sample points of the input domain:

\begin{equation}\label{eq: loss function }
    L \,=\, \sum_{\text{batch}}(D\psi^{NN}(x^\mu))^2 + \lambda L_{add}
\end{equation}
where $D$ represents a differential operator that will depend on the differential equations that we are considering. Here $L_{add}$ represents any additional contribution to the loss that can be added to give extra information to the model during training. The weight of this piece is controlled by $\lambda$.

While traditional numerical methods often outperform PINNs in terms of computational efficiency and accuracy, recent developments in PINNs are offering unique advantages. One key benefit is the ability of PINNs to learn the parametric dependence of the solutions to different initial or boundary conditions, or even different values of parameters appearing in the equations. In contrast with traditional numerical methods, this allows us to train the model on a wide range of different parameters at once, whereas a traditional solver must be rerun for different values of initial/boundary conditions or parameters. This eventually will make a PINNs setup much more efficient compared with a numerical method, since obtaining a solution for a particular value of a parameter just requires the evaluation of the model. This efficient evaluation permits us to treat otherwise untreatable problems. One of which is the so-called inverse problem \cite{Bea:2024xgv,Tarancon-Alvarez:2025wux}. This consists of recovering a free-form function or parameter appearing on the differential equations just by considering boundary data. This capability is extremely useful for this work, since we will use it to recover the viscosity of Burger's equation by considering the initial and final time configuration of a solution. Another key advantage of PINNs is their capability of learning a general high-dimensional space of functions called the latent space. The final solution would just be a combination of these functions in a certain way. This latent space will, in general, depend on the independent variables of the problem, but it can also be learned for more parameters, like different boundary conditions or different values of a parameter appearing in the equations. In this work, we learn a latent space of Burger's equation for different values of the viscosity $\nu$ in the so-called Multihead training. We will explain this framework in the following subsection.

Although we have presented the two main advantages that we are going to exploit throughout this work of using PINNs instead of traditional numerical methods, there exist many more. PINNs are mesh-free, allowing for obtaining a solution at any point lying inside the training set. Moreover, they can also be used to find new solutions on stiff regimes using their ability to transfer learning \cite{Tarancon-Alvarez:2025wux}.

\subsubsection{Multihead training}
 Multihead (MH) training consists of breaking the neural network into two pieces: the body, which will learn a higher-dimensional representation (latent space) of the solutions depending on the values of different parameters or boundary conditions; and the head, which will take care of learning how to project from this latent space to the actual solution. This setup has already been proven useful to transfer learn solutions of differential equations to stiffer regimes. However, we use MH training to explore the functions of the latent space when the head is forced to present a linear response. This will tell us valuable information about the functional ``basis'' that the model is learning to solve the problem.

 The NN gives us the final solution $\psi^{NN}(x^\mu)$  composed of two parts
\begin{itemize}
    \item \underline{Body}: this itself is going to be a NN which is going to learn a high-dimensional space called the latent space of solutions. We will denote it as $H_i(x^\mu)$, where $i$ runs from $1$ to the total dimensionality of this space. In general, this NN must be large enough such that it can learn a proper representation of the solutions in this higher-dimensional space. In general, this body will take as inputs the independent variable plus variations of parameters or different boundary conditions of the equations, denoted by $x^\mu$. After passing them through the model, it will return us the latent space $H_{i}(x^\mu)$, which will be composed of $d$ components.
    \item \underline{Head}: this component is also going to be a NN and will take care of processing the latent space component and combine them on a particular way such that the final result is going to be the desired approximate solution to the differential equation $\psi^{NN}(x^\mu)$. In general, this model is much smaller than the body, since only a few operations on the latent space will be required to produce the final solution. Generally, the head will take the latent space of dimension $d$, and combine its components such that the final output is dimension $1$. We will have one head for each of the values of the free parameter or boundary conditions that we are considering. Thus, each of them will learn a particular set of weights and biases that will project down the latent space to the solution for a particular value of the parameters and boundary conditions chosen.
\end{itemize}
At the end, the solution can be expressed as
\begin{equation*}
    \psi^{NN}(x^\mu) \,=\, \text{head}_W[H_i(x^\mu)]
\end{equation*}
where $\text{head}$ denotes the response of the head, and $W$ the set of its weights and biases.

The model is trained as follows. Once one approximated solution is obtained, we can compute its loss function just by evaluating the residues of the differential equations squared (\ref{eq: loss function }).  The total loss will be the sum of each of the individual losses across all the heads. We can then train the model by optimizing this loss function. At the end, the model will solve simultaneously all specified differential equations problems. It is important to note that, since the body is shared across all NNs, it will capture global properties of the solutions for different values of free parameter or boundary conditions.

\subsubsection{Latent space structure with \texttt{Denario}}

Our question to \texttt{Denario} then was to find the structure of the latent space generated by training on Burger's equation for different initial conditions and values of the viscosity. The reason behind it was that the structure of the latent space might reveal the actual physics determining the structure of the solutions to this equation. So we asked \texttt{Denario} to explore the structure of the latent space of solutions, having given it the following results: we provided an array that contained the latent space solutions as a function of $\nu$, space, and time. 
The input text passed to \texttt{Denario} was this:

\begin{tcolorbox}[colback=blue!5!white, colframe=blue!75!black, title=Input text, breakable]

We haved PINNs to solve a Burger equation in 2 dimensions for a fixed value of the viscosity with multihead set for different initial conditions. The latent space is located in this file 
\begin{lstlisting}[language = Python]
/Users/fvillaescusa/Documents/Software/Denario/Project\_turbulence/data\_for\_Paco\_turbulence\_bundle.npy
\end{lstlisting}

To read the data, you can do:
\begin{lstlisting}[language = Python]
import numpy  as np
fin = '/Users/fvillaescusa/Documents/Software/Denario/Project_turbulenceV1/data_for_Paco_turbulence_bundle.npy'
data = np.load(fin)
\end{lstlisting}

data is now a [101,103,25,13] matrix, where the first axis represents the x-coordinate, the second axis is time, and the third component represents the viscosity. The fourth axis is organized as follows:
- first three components are the mesh of space, time and the viscosity, respectively
- the rest (10 last components) the latent space

Perform a detailed study of the latent space structure. We are interested in knowing if there is some pattern or simple way to describe its properties.

\end{tcolorbox}

We have run Denario end-to-end from the above prompt to generate a final article. The generated paper, titled \textit{Geometric Structure of PINN Latent Space for Burger’s Equation: Low-Dimensional Manifolds and Initial Condition Encoding}, can be found in the Appendix \ref{paper:matphys}. In this paper, \texttt{Denario} performed a detailed analysis of the given latent space of Burger's equation. It applied common techniques of dimensionality reduction on the provided data set, mainly Principal Component Analysis (PCA) and subspace similarity measures. Although this analysis is quite standard, it includes some points that were surprising. The first one was a detailed analysis of the centroids of the manifolds, computing their PCA individually and discovering that there was only one dominant component when varying the viscosity. Then, it performed a similar analysis on the 3-dimensional principal subspaces for each value of the viscosity. This analysis included subspace similarity between these manifolds and the dot products of these vectors. The conclusion was that these manifolds were almost parallel to each other, and that they described up to $97.67 \%$ of the variance. At last, it analyzed the global structure of the latent space using the previously mentioned techniques. It concluded that the PINN is learning a 3-dimensional manifold to describe a solution for each value of the viscosity. Variations of this parameter are encoded as translations of this 3-dimensional space across a 1-dimensional sub-manifold. We would also like to remark that \texttt{Denario} produced unexpected results at one point. Throughout the whole paper, it is remarked that the data is obtained for different initial conditions. However, this is not true, and the index that was treated as different initial conditions corresponded to different values of the viscosity. This confusion was due to a misleading prompt on our side. In this respect, \texttt{Denario} showed robustness in that it followed the prompt instructions and did not simply try to repeat what is in the literature.

As a summary, we would like to remark on the analysis that \texttt{Denario} has performed on the data, not as a novel and groundbreaking job, but as a tool for researchers to unveil future directions that are worth investigating in detail.

\subsection{Planetary Science. Mapping Size and Compositional Gradients in the Asteroid Belt}

The asteroid belt preserves a fossil record of the solar system's formation and dynamical evolution. We compiled a dataset of over 35,000 main-belt asteroids, integrating orbital parameters, diameters, and spectral types. These data were obtained from a wide range of sources, including the Lowell dataset \citep{Bowell1994}, the Asteroid Lightcurve Database (LCDB) \citep{Warner2009}, DAMIT \citep{Durech2010}, mp3c \citep{mp3c}, and the Python tool {\tt classy} \citep{Mahlke2022}. We then asked \texttt{Denario} to analyze the data. The input text we used was this:

\begin{tcolorbox}[colback=blue!5!white, colframe=blue!75!black, title=Input text, breakable]
We have some data about asteroids. The data is located in /mnt/ceph/users/fvillaescusa/AstroPilot/Asteroids/data\\

Each asteroid is identified by a unique number. For example, 1 corresponds to the first discovered asteroid, Ceres.\\

All CSV files share a common structure:
\begin{itemize}
\item Column 1: Asteroid identification number
\item Column 2: Corresponding physical or orbital property
\end{itemize}
These files are recommended to be opened with the Python package "pandas".

File Descriptions:
\begin{itemize}
\item asteroid\_name.csv: number + name of the asteroid
\item asteroid\_diameter.csv: number + diameter of the asteroid in kilometers
\item asteroid\_semimajor\_axis.csv: number + semimajor axis of the orbit in astronomical units (AU)
\item asteroid\_eccentricity.csv: number + eccentricity of the orbit
\item asteroid\_inclination.csv: number + inclination of the orbit in degrees
\item asteroid\_arg\_peri.csv: number + argument of periapsis of the orbit in degrees
\item asteroid\_long\_asc\_node.csv: number + longitude of ascending node of the orbit in degrees
\item asteroid\_spin\_period.csv: number + spin period of the asteroid in hours
\item asteroid\_obliquity.csv: number + obliquity of the asteroid in degrees (Obliquity is defined as the angle between the spin vector and the orbital angular momentum vector.)
\item asteroid\_type.csv: number + spectral type of the asteroid
\item asteroid\_family.csv: number + family name of the asteroid (The family name corresponds to the largest member of the family)
\item asteroid\_age.csv: number + age of the asteroid in gigayears (Gyr)
\end{itemize}

Please come up with an interesting project for a PhD thesis. Use state-of-the-art methods to analyze the data. For every step, make lots of plots and save the data you generate, as it may be used for other steps. When writing the code, write some lines to indicate whether the execution was successful or not. Join plots that are similar. Do not create dummy data. You have access to 128 cpus; for computationally heavy tasks, try to use all of them.
\end{tcolorbox}

The generated paper, titled \textit{The Spatial Architecture of the Main Asteroid Belt: Size, Composition, and Dynamical Gradients}, can be found in the Appendix \ref{paper:planet}. By applying density estimation, clustering (DBSCAN and GMM), and predictive modeling, \texttt{Denario} mapped spatial variations in size and composition, identified asteroid families, and examined the influence of Jupiter's resonances on the belt’s structure. A particular strength of the paper lies in its ability to automate the quantification of compositional gradients across the belt. This paper reveals an important compositional trend: the dominant composition shifts from S-type in the inner solar system to C-type in the outer solar system. By mapping the changing fractions of spectral types as a function of semimajor axis, it offers an accessible, up-to-date reference that will be useful for researchers studying planetesimal formation and solar system evolution. The clustering analyses also effectively reproduce known asteroid families, confirming the AI’s capacity to extract meaningful dynamical structures from complex data.

However, while these results are scientifically sound, they are not fundamentally new. The compositional gradient and dynamical families are well-established in the literature. The observed trend of increasing average asteroid size with distance from the Sun is also correctly noted as likely driven by observational bias rather than a genuine physical gradient. One area for improvement is the depth of interpretation. While \texttt{Denario} correctly applied clustering and density estimation, it stops at describing known structures without critically assessing the potential for new dynamical groupings that might hint at unrecognized collisional families or early migration signatures. This illustrates a broader limitation: although AI can uncover patterns efficiently, it currently lacks the domain intuition to pose new hypotheses or recognize subtle, novel features without human input.

Similarly, the paper’s predictive modeling using only orbital data yielded limited success, appropriately acknowledging the role of stochastic collisional histories. But it could have gone further by discussing how including additional parameters—such as albedo, spin state, or detailed spectral data—might improve model performance. On the other hand, in its current form, \texttt{Denario} is not designed to do this kind of critical thinking, so perhaps it is not a surprise that it does not perform well here. 

It is also worth noting the experience with another AI-generated paper (not included here, but available in the example papers repository\footnote{\url{https://github.com/AstroPilot-AI/DenarioExamplePapers}}), which focused on spin-orbit coupling effects. In that case, \texttt{Denario} produced a figure that appeared to show features not previously discussed in the literature. However, \texttt{Denario} itself did not recognize or highlight these potentially novel patterns, and other figures in that paper were poorly prepared, limiting its usefulness. This example underscores both the promise and the current limitations of AI approaches: when handling large, complex datasets, AI can be more efficient than humans at surfacing candidate phenomena, but human expertise is essential to interpret, validate, and refine these discoveries. With careful collaboration, such human–AI iteration could focus attention on these features in subsequent analyses, potentially leading to truly novel scientific breakthroughs.

Overall, while this paper does not claim any fundamentally new discovery, that is not a significant flaw. It demonstrates that \texttt{Denario} can systematically confirm and clearly document established knowledge, producing scientifically valuable and well-visualized results in an efficient, reproducible manner. As the era of big data arrives with observatories like Very Large Telescope (VLT) \citep{Lenzen2003}, the Large Synoptic Survey Telescope (LSST) \citep{Ivezic2019}, and The Extremely Large Telescope (ELT) \citep{Gilmozzi2007}, tools like \texttt{Denario} will become indispensable in planetary and exoplanetary science—helping researchers sift massive datasets, probe early solar system conditions, and advance our understanding of the origins of life both here and beyond.

\subsection{Medicine and Health Services Research. Assessment of Fertility Clinic Outcomes in the United States}

Health services research is a multidisciplinary field that studies healthcare utilization, access, delivery, and financing, with the goal of improving the quality, efficiency, and equity of healthcare. Combining insights from medicine, public health, economics, and statistics, health services research aims to inform decisions by providers, payers, policymakers, and patients.

To evaluate \texttt{Denario}'s capability to design and perform a project in this field, we used the publicly available datasets from the National ART Surveillance System\footnote{\url{https://www.cdc.gov/art/php/nass}}. These annual datasets, published in the United States by the CDC (US Centers for Disease Control and Prevention), contain clinic-level aggregated data on the utilization and success rates of approximately 98$\%$ of all assisted reproductive technology (ART) services in the United States. Originally mandated by the US Congress to provide transparency and information to patients seeking fertility treatments, these datasets have been extensively used to evaluate the quality, efficiency, and distribution of fertility services. 

More specifically, we provided \texttt{Denario} with the three most recent years available for NASS (2020-2022) and a description of the dataset and its variables. We then instructed \texttt{Denario} to suggest and develop potentially interesting venues of scientific investigation.  The exact input text we provided it with is this:

\begin{tcolorbox}[colback=blue!5!white, colframe=blue!75!black, title=Input text, breakable]
The Final Assisted Reproductive Technology (ART) Success Rates dataset is part of the CDC’s National ART Surveillance System (NASS) and provides clinic-level success rate data for procedures conducted between 2020 and 2022. It includes information from U.S. fertility clinics on ART cycles, distinguishing whether patients used their own eggs or donor eggs, and whether ART was their first treatment or a follow-up. The dataset presents cumulative success rates (including embryo transfers over 12 months after retrieval) for own‑egg patients by retrieval and transfer, and non-cumulative rates for donor‑egg cycles. It also includes statistics like average transfers per retrieval and retrievals per live birth. These data—updated most recently in December 2024 and made public in early 2025—offer insights into clinic performance and national ART outcomes.

Among the dataset's relevant columns are:

\begin{itemize}
\item ClinicID – unique identifier linking this Success Rates dataset to other ART data (Services, Characteristics, Summary).

\item Reporting Year – the calendar year.

\item Age Group – patient age brackets (e.g., $<$35, 35–37, 38–40, etc.).

\item Live Birth per Intended Retrieval – percentage of intended egg retrieval cycles resulting in live births within a year.

\item Live Birth per Transfer – percentage of embryo transfer cycles resulting in live births within the same reporting window.

\item Live Birth per Actual Retrieval – reflecting actual retrieval procedures conducted.

\item Avg Transfers per Intended Retrieval – average number of embryo transfers per intended retrieval.

\item Avg Intended Retrievals per Live-Birth – average number of retrieval cycles needed to yield a live birth.

\item Donor Egg Live Birth Rate – percentage of donor‑egg embryo transfer cycles leading to live births.

\item Clinic Export Variables – additional metrics such as clinic location, number of cycles, and whether it was first-time ART cycles.
\end{itemize}

These columns capture both performance and efficiency metrics, enabling comparisons across clinics and patient groups—own‑egg vs. donor‑egg, first‑time vs. any‑time ART, age categories, and aggregated national data. The description of the columns, together with their data type is this:

\small  
\begin{longtable}{|p{2.8cm}|p{6.5cm}|p{2.8cm}|p{1.0cm}|}
\hline
\textbf{Column Name} & \textbf{Description} & \textbf{API Field Name} & \textbf{Data Type} \\
\hline
\endfirsthead

\hline
\textbf{Column Name} & \textbf{Description} & \textbf{API Field Name} & \textbf{Data Type} \\
\hline
\endhead

Year & Reporting year & year & number \\
\hline
LocationAbbr & Abbreviated state for ART clinic location & locationabbr & text \\
\hline
LocationDesc & State for ART clinic location & locationdesc & text \\
\hline
FacilityName & ART clinic name or national data indicator & facilityname & text \\
\hline
MedicalDirector & Name of medical director verifying data & medicaldirector & text \\
\hline
Address & Address for ART clinic location & address & text \\
\hline
City & City for ART clinic location & city & text \\
\hline
ZipCode & Zip code for ART clinic location & zipcode & text \\
\hline
Phone & Phone number for ART clinic & phone & text \\
\hline
Clinic Status & Operating status for ART clinic, e.g. open, closed, reorganized & clinic\_status & text \\
\hline
Type & Type description, e.g. patients using their own eggs, patients with no prior ART using their own eggs & type & text \\
\hline
Topic & Topic description, e.g. clinic services and profile, patient and cycle characteristics, success rates: patients using donor eggs, success rates: patients using own eggs, or summary & topic & text \\
\hline
Question & Question description, e.g. specific patient and cycle characteristic or specific success factor & question & text \\
\hline
Filter & Filter description, e.g. infertility diagnosis & filter & text \\
\hline
Breakout\_Category & Stratification grouping, e.g. egg/embryo type, age group of patient & breakout\_category & text \\
\hline
Breakout & Stratification value, e.g. fresh embryos fresh eggs, $<$35 years & breakout & text \\
\hline
Data\_Value & Data value, e.g. number, percentage, or fraction within the breakout, question or subtopic & data\_value & text \\
\hline
Data\_Value\_num & Data numeric values only, e.g. number, percentage within the breakout, question or subtopic & data\_value\_num & number \\
\hline
Data\_Value Footnote\_Symbol & Symbol used to flag footnotes, if applicable & data\_value footnote\_symbol & text \\
\hline
Data\_Value\_Footnote & Footnote description, if applicable & data\_value\_footnote & text \\
\hline
Cycle\_Count & Total number of ART cycles & cycle\_count & number \\
\hline
ClinicId & ART clinic identifier & clinicid & number \\
\hline
TypeId & Type identifier & typeid & text \\
\hline
TopicId & Topic identifier & topicid & text \\
\hline
QuestionId & Question identifier & questionid & text \\
\hline
FilterId & Filter identifier & filterid & text \\
\hline
BreakOutCategoryId & Stratification grouping identifier & breakoutcategoryid & text \\
\hline
BreakOutId & Stratification value identifier & breakoutid & text \\
\hline
GeoLocation & Location code for ART clinic location to be used for geocoding & geolocation & point \\
\hline
\end{longtable}

The data is located here in /mnt/ceph/users/fvillaescusa/AstroPilot/Medicine/Fertility1/data/art\_data\_2020\_2024.csv. Please come out with interesting ideas for a PhD student to analyze this data and learn some insight from it. Use state-of-the-art methods to analyze the data. For every step, make lots of plots and save the data you generate, as it may be used for other steps. When writing the code, write some lines to indicate whether the execution was successful or not. Join plots that are similar. Do not create dummy data. You have access to 128 cpus; for computationally heavy tasks, try to use all of them.

\end{tcolorbox}

The generated paper, titled \textit{Characterizing the Variability and Correlates of U.S. ART Clinic Performance During the COVID-19 Pandemic (2020-2022)}, can be found in the Appendix \ref{paper:med}. In this paper, \texttt{Denario} characterized the year-to-year variability in success and efficiency metrics of ART clinics, such as live birth rates per retrieval and average retrievals per live birth, stratified by patient age and egg source (own vs. donor). The dataset was evaluated using standard statistical analyses, such as the coefficient of variation, to analyze fluctuations in success rates, ordinary least squares regression models, and Spearman correlation to characterize associations between variability and clinic volume. While the analyses were relatively straightforward and would likely need to be redone using other statistical tools, the contextualization and characterization of the research question were remarkable. Despite the recognized importance of {\it inter}-clinic variability in outcomes, little attention has been given to the characterization of year-to-year {\it intra}-clinic fluctuations in ART clinic performance, particularly in the context of a major systemic shock like the COVID-19 pandemic. \texttt{Denario} contextualized the research by not only evaluating an overlooked topic in the literature, but also by realizing the impact of the global pandemic on healthcare systems through disruptions like clinic closures, supply chain issues, and shifts in patient behavior.

The final manuscript generated by \texttt{Denario} had some limitations on the presentation of the results and in the citations to the literature. The results and overall content of the manuscript were significantly more technical than the standard in the field, which generally focuses more on the discussion and implications of the results. Statistical analyses were limited to standard tools such as ordinary least squares and correlations, which (as recognized in the manuscript) would need to be re-evaluated given the distributional properties of the variability measures. Another limitation was the fact that \texttt{Denario} used only open-access manuscripts available at preprint repositories, such as the arXiv. While the use of preprint repositories is a common practice in most technical areas, including economics, their use is still limited in the field of health services research and most areas of medicine. For this reason, the references included in the final manuscript generated by \texttt{Denario} were not necessarily applicable and were, at best, marginally relevant to the sentence in which they are used. 

Nonetheless, despite the limitations described above, the research conducted by \texttt{Denario} was genuinely engaging and interesting, highlighting an understudied topic in the area and contextualizing it in a novel way. This manuscript not only identifies promising directions for future investigation but also includes methodology and results that would be useful resources for researchers in the field to further their investigations.

\subsection{Chemistry. Molecular dynamics simulations of self-assembling peptides}
Self-assembling peptides (SAPs) are promising candidates for biomedical applications such as tissue engineering, wound healing, and drug delivery. Their modular structure and biocompatibility make them especially attractive for engineering soft materials with targeted functions. Understanding the relationship between structure and properties in SAPs is key to unlocking this potential. To investigate this, we tasked \texttt{Denario} with analyzing a 1200 ns all-atom molecular dynamics (MD) trajectory containing 30 copies of a known self-assembling pentapeptide: KYFIL. We conducted the study over three rounds\footnote{In other words, we generated papers from slightly different prompts.}, each time increasing the level of contextual information provided in the input text.

In the first round, we instructed \texttt{Denario} on which Python tools to use for trajectory analysis but gave no domain-specific context. \texttt{Denario} correctly identified the system as a protein but misinterpreted it as a single large molecule rather than 30 individual peptides. As a result, it analyzed residues in isolation, failing to recognize that they were equivalent across different peptide chains. This led to an overemphasis on per-residue metrics like dihedral angles or motif detection, rather than ensemble behaviors across peptides. Consequently, key phenomena like peptide aggregation and clustering were missed.

In the second round, we added a system description and a research abstract to provide context. With this information, \texttt{Denario} identified more appropriate analyses related to self-assembly, such as contact mapping and cluster size distributions. While the outputs were more relevant, they remained conventional and did not go beyond standard MD analysis pipelines. However, the automation of routine essential tasks was already a valuable outcome.

In the third round, we explicitly posed hypotheses and research questions. Specifically, we suggested modeling peptide clusters as graphs and proposed several graph-theoretical analyses, along with practical motivations for each. The input text we used for this round is this:

\begin{tcolorbox}[colback=blue!5!white, colframe=blue!75!black, title=Input text, breakable]

We have data from a molecular dynamics numerical simulation. The data is located in /mnt/ceph/users/fvillaescusa/AstroPilot/Project\_chemist1/data. This trajectory contains an all atom molecular dynamics simulation of 30 copies of the self-assembling pentapeptide, K-Y-F-I-L. The simulation starts with the peptides fully separated and over the course of the simulation the peptides form noncovalent interactions with one another to form an aggregate.\\

\textbf{Motivation:} The development of biomaterials that mimic the native extracellular matrix (ECM) of brain tissue represents a promising frontier for neural repair and regeneration. Injectable hydrogels made of short, self-assembling peptides offer a promising platform for the delivery and directed differentiation of therapeutic stem cells. Despite the growing interest in this field, the rational design of self-assembling peptide sequences remains largely empirical. Our research aims to address this gap by computationally modeling a library of rapidly assembling pentapeptides for injectable delivery (RAPID)-a family of C-terminally amidated, five-residue peptides engineered to self-assemble into shear-thinning, self-healing hydrogels. These RAPID hydrogels have demonstrated the potential to improve cell survival upon minimally invasive delivery.  To better understand the mechanisms behind assembly behavior, we examined KYFIL, a pentapeptide known to self-assemble, using molecular dynamics (MD) simulations to determine the molecular scale behaviors that distinguish gelling from non-gelling sequences.\\

The folder contains two files:
\begin{enumerate}
\item stripped.nc: the trajectory file which contains the coordinates as a function of time
\item stripped.parm7: contains information about what each of the atom types are (which coordinates belong to which molecules).
\end{enumerate}
 
The data can be read as:

\begin{lstlisting}[language = Python]
import MDAnalysis as mda
 
# Load the AMBER topology and trajectory
topology_file = 'system.prmtop'       # Replace with your actual .prmtop file
trajectory_file = 'trajectory.nc'     # Replace with your actual trajectory file
 
# Create the Universe object
u = mda.Universe(topology_file, trajectory_file)
 
# Print some basic info
print("Number of atoms:", len(u.atoms))
print("Number of frames in trajectory:", len(u.trajectory))
 
# Accessing trajectory frames
for ts in u.trajectory[:5]:  # Iterate through the first 5 frames
    print(f"Frame: {ts.frame}, Time: {ts.time} ps")
\end{lstlisting}

\textbf{Considerations:}
\begin{itemize}
\item In order to understand how the simulated system compares to experiments, we need to understand how the system behaves at equilibrium (or as close to equilibrium as possible). Therefore, we are most interested in measuring quantities starting from about 100ns into the trajectory until the end of the trajectory (as opposed to the beginning of the trajectory).
\item We are interested in making graphs that are either a function of time or averaged over long time intervals. We are not interested in measurements that are averaged over small time intervals or representative of a single snapshot.
\end{itemize}

\textbf{Research Questions:}
\begin{itemize}
\item You can consider the peptides aggregates as two different types of graphs:
    \begin{itemize}
  \item First, a graph where each peptides is a node and an edge between nodes represents the binary property of whether or not those two nodes have a heavy atom contact
  \item Second, where the aggregates are a hierarchical graph, where the coarse grained graph is the same as above, but each peptide is additionally made up of a fine grained graph where every amino acid is a node
  \end{itemize}

\item With these two graph definitions we can measure graph theoretical properties, which tell us about the overall structure of the graph
    \begin{itemize}
  \item We can construct the graph laplacian and calculate eigenvectors and eigenvalues
  \item We can calculate the Fiedler vector
  \item Are there other interesting graph theoretical quantities to explore?
  \end{itemize}

\item We would want to know how do these properties:
\begin{itemize}
    \item Look on average?
    \item Fluctuate about that average?
    \item Change dynamically over time?
\end{itemize}

\item We would like these properties to tell us:
\begin{itemize}
    \item How densely packed are the clusters?
    \item Can this approach be used to predict whether a cluster will split or be long lived?
    \item What are different ways that we can measure peptide clusters or aggregates?
    \begin{itemize}
        \item Our current definition is that a cluster is a set of nodes where none of the nodes is fully disconnected from any of the other nodes. But one challenge with this is that the number of clusters varies widely as clusters form and uniform over the simulation. Is there a way to combine this measurement with some graph theoretical definition of graph density in order to construct a more stable order parameter?
        \end{itemize}
\end{itemize}
\end{itemize}

Use state-of-the-art methods to analyze the data. For every step, make lots of plots and save the data you generate, as it may be used for other steps. When writing the code, write some lines to indicate whether the execution was successful or not. Take into account that all the things print on screen will be passed to a LLM, thus, only write essential information and don't print unnecessary information. Join plots that are similar. Do not create dummy data. You have access to 128 cpus on a linux machine; for very computationally heavy tasks, try to use all of them. 
\end{tcolorbox}

With this guidance, \texttt{Denario} performed novel, domain-relevant analyses that had not been previously utilized in the field of peptide self-assembly—and executed them efficiently. The generated paper, titled \textit{Dynamic Multiscale Graph Analysis Reveals Structural Signatures of Peptide Aggregate Stability and Splitting} can be found in the Appendix \ref{paper:chem}. These results showed that, for \texttt{Denario} to exhibits its full capability, a very detailed and extensive description of the data is needed, and in some cases, guidance over the research direction.

Overall, \texttt{Denario} demonstrated a clear progression in capability based on the depth of contextual input. With minimal context, it could automate standard analyses; with targeted hypotheses, it extended into novel analytical territory. While \texttt{Denario} didn’t independently generate insights a human expert wouldn’t have, it significantly accelerated the exploration of both standard and creative research avenues.

\subsection{Astrophysics. Properties of recent high-mass binary black hole merger from gravitational wave data}

We provided \texttt{Denario} with publicly available gravitational wave posteriors \cite{GW231123_data} from multiple waveform models for a recently announced gravitational wave event GW231123 \cite{LIGOScientific:2025rsn}. GW231123 is the most massive confident black hole merger detected till now and has implications for the formation channels of astrophysical black holes. However, there is a significant disagreement between properties of the event as predicted by different waveform models (see their description in the prompt below). This was the motivation for us to perform an exploratory study with \texttt{Denario} regarding the high-dimensional posterior dataset from multiple models. 

The input text we used for this paper is this:

\begin{tcolorbox}[colback=blue!5!white, colframe=blue!75!black, title=Input text, breakable]
We have data from a recent gravitational wave event: GW231123. The data is located in five files:
\begin{lstlisting}[language = Python]
- /mnt/ceph/users/fvillaescusa/AstroPilot/GW/Iteration1/data/GW231123\_NRSur7dq4.csv
- /mnt/ceph/users/fvillaescusa/AstroPilot/GW/Iteration1/data/GW231123\_IMRPhenomXO4a.csv
- /mnt/ceph/users/fvillaescusa/AstroPilot/GW/Iteration1/data/GW231123\_SEOBNRv5PHM.csv
- /mnt/ceph/users/fvillaescusa/AstroPilot/GW/Iteration1/data/GW231123\_IMRPhenomXPHM.csv
- /mnt/ceph/users/fvillaescusa/AstroPilot/GW/Iteration1/data/GW231123\_IMRPhenomTPHM.csv
\end{lstlisting}

These files contain samples from the posterior distribution corresponding to five different gravitational-wave waveform models which are detailed below. The meaning of each column can be found in the file header, but is also provided below for reference.\\

We are interested in understanding which ways the different models agree or differ in their prediction for GW231123. There are complex degeneracies in the high-dimensional posterior space which make this problem challenging. Analyze the datasets in detail and tell us what you have learned from them. Mention also any interesting astrophysical insights that you learn from this analysis, and what can be robustly concluded about statistical properties of the high-mass black hole merger GW231123. Make sure there is no repetition in plots in the paper. For reference, the event was reported in https://arxiv.org/pdf/2507.08219\\

Below is a description of the models used for the five files mentioned above:

\begin{itemize}
\item NRSur7dq4 is a time-domain waveform model that has been directly calibrated using Numerical Relativity simulations of binary black hole mergers. This model is particularly well-suited for events with short gravitational wave signals, such as GW231123, as it is designed to accurately capture the dynamics of the late inspiral, merger, and ringdown phases.

\item IMRPhenomXO4a, on the other hand, is a phenomenological model that operates in the frequency domain. It combines post-Newtonian (PN) approximations during the inspiral phase with a numerical relativity calibration near the merger. This model provides a more global approximation of binary black hole waveforms and is specifically tailored for events that can be characterized over a wider range of frequencies.

\item SEOBNRv5PHM
Domain: Time-domain, effective-one-body (EOB)
Form: Co-precessing to inertial frame
Key: Efficient inspiral-merger-ringdown with higher modes
Calibrated to: NR + perturbation theory, high accuracy (unfaithfulness $<3\%$ in ~99.8\% of cases)

\item IMRPhenomXPHM
Domain: Frequency-domain, phenomenological
Form: "Twisting-up" non-precessing multipoles into precessing frame
Key: Multipolar, fast, calibrated to waveform catalogs 

\item IMRPhenomTPHM
Domain: Time-domain, phenomenological
Form: Time-based twisting-up using Euler angles + improved ringdown precession treatment
Key: Avoids stationary phase, accurate and computationally efficient
\end{itemize}

Here is a description of the columns in the csv file:
\begin{itemize}
\item `mass\_1\_source': Mass of the primary black hole
\item `mass\_2\_source': Mass of the secondary black hole
\item `a\_1': Spin magnitude of the primary black hole
\item `a\_2': Spin magnitude of the secondary black hole
\item `final\_mass\_source': Final mass the remnant black hole
\item `final\_spin': Final spin of the remnant black hole
\item `redshift': Redshift of the event
\item `cos\_tilt\_1': Cosine of the spin tilt-angle of the primary black hole
\item `cos\_tilt\_2': Cosine of the spin tilt-angle of the secondary black hole
\item `chi\_eff': Effective sum of spins components aligned with the orbital angular momentum
\item `chi\_p': Effective spin parameter related to spin-orbit precession of the binary
\item `cos\_theta\_jn': Cosine of the inclination angle of the binary to the observer
\item `phi\_jl': Azimuthal angle between the total and orbital angular momentum (called spin azimuth)
\item `log\_likelihood': Log-likelihood of the samples
\end{itemize}

Use state-of-the-art methods to analyze the data. For every step, make lots of plots and save the data you generate, as it may be used for other steps. When writing the code, write some lines to indicate whether the execution was successful or not. Join plots that are similar. Do not create dummy data. 
\end{tcolorbox}

The generated paper, titled \textit{Physics-Informed Discrepancy Decomposition and Robust Astrophysical Inference for GW231123}, can be found in the Appendix \ref{paper:gw}. In this paper, \texttt{Denario} performed the exploratory data analysis very well. Without an explicit mention of which tool(s) would be the best to use, \texttt{Denario} identified dimensionality reduction tools like UMAP (which have not been widely used in the previous gravitational wave literature) for this problem. The resulting 2D UMAP embeddings in Fig.~2 of the paper have a very interesting structure. The different waveform models are built in either the time or frequency domain. The UMAP embeddings produce a common cluster for the three time-domain models, suggesting that their predicted posteriors are consistent with each other. However, for the two frequency domain models, the embeddings form separate isolated groups, highlighting that their predictions are inconsistent with each other (Fig.~2). \texttt{Denario} also verified that the physical explanations are independent of the hyperparameters (see Fig.~3). The Jensen-Shannon divergence plots in Fig.~4 are useful for getting an intuition of which parameters the different models differ. 

 This paper can be comparable to a report generated from an undergraduate-level data exploration project. While the paper is not currently ready to be published in a scientific journal, it definitely provides interesting ideas for further exploration (it can be turned into a scientific paper upon further refinement of some of the ideas presented).
With each paper taking a very small time to generate, speed definitely is the biggest advantage of using \texttt{Denario} and it can be used to obtain quick project ideas from a dataset. The drawback is \texttt{Denario} not having knowledge of the broader context of the field (apart from the information specified in the prompt, and to some extent the summaries provided by the `enhanced data', and `check idea' options). Thus, it could not fully place the methods and results of the paper within the context of existing literature (which is crucial for a research paper). However, one can envision further developing our literature agents so they can perform these tasks using tools such as Semantic Scholar.

\subsection{Astrophysics. Impact of Convection and Radiation Pressure on the Dynamics of Roche Lobe Overflow in Binary Systems}

Massive stars play many important roles in astrophysics. However, the structures and evolutions of massive stars remain very uncertain due to complicated dynamical processes controlled by both gas pressure and radiation pressure. In particular, the majority of massive stars are found in binary systems. Interactions with the companion can significantly impact the evolution of massive stars. Binary interactions and the associated mass transfer via the Roche Lobe overflow are the physical origins of many astrophysical phenomena. In recent years, with the advanced computational power and improved numerical algorithms, it is now possible to simulate the process of binary mass transfer between massive stars using three-dimensional radiation hydrodynamic simulations. 

One simulation we have analyzed is a binary system with a 12 solar mass Red supergiant (RSG) star as a donor and a 9 solar mass main-sequence star as an accretor. The simulation resolved the three-dimensional structures at the surface of the donor and followed the mass transfer to the accretor due to the binary interaction. The surface of RSG is fully convective, which can make the Roche Lobe overflow significantly different from what has been imagined in classical theoretical models. We used \texttt{Denario} to analyze one snapshot of the simulation results to understand how the binary mass transfer happened in the simulation. The prompt we used as input to Denario is this:

\begin{tcolorbox}[colback=blue!5!white, colframe=blue!75!black, title=Input text, breakable]
We have a snapshot of a simulation designed to study the impact of realistic 3D envelope structures on the mass transfer in a binary system. The binary is made of a 12.9 solar mass red supergiant (RSG) and a 9 solar mass companion. The companion is located at a separation of 2000 solar radii, which is about three times the radius of the donor (the RSG). The simulation shows that the convective surface structures significantly modify the mass transfer rate in the binary system. The snapshot is located in /mnt/ceph/users/fvillaescusa/AstroPilot/Yan-Fei/data/star.out1.16543.athdf\\

To load the data use this example:

\begin{lstlisting}[language = Python]
import matplotlib
matplotlib.use('Agg')
%matplotlib inline
import sys
sys.settrace
import numpy as np
import matplotlib.pyplot as plt
from matplotlib.colors import LogNorm
from pylab import *
import struct
import array
import os
import glob
import h5py
import sys
sys.path.append("/mnt/ceph/users/fvillaescusa/AstroPilot/Yan-Fei/data/")
import athena_read

filename = '/mnt/ceph/users/fvillaescusa/AstroPilot/Yan-Fei/data/star.out1.16543.athdf'

with h5py.File(filename, 'r') as f:
  attributes = f.attrs.items()
  attrs = dict(attributes)
  level = f.attrs['MaxLevel']
  time = f.attrs['Time']
  subsample = False

  nx1 = attrs['RootGridSize'][0] * 2**level
  nx2 = attrs['RootGridSize'][1] * 2**level
  nx3 = attrs['RootGridSize'][2] * 2**level


data = athena_read.athdf(filename, level=level, subsample=subsample)

print(data.keys())

dict_keys(['Coordinates', 'DatasetNames', 'MaxLevel', 'MeshBlockSize', 'NumCycles', 'NumMeshBlocks', 'NumVariables', 'RootGridSize', 'RootGridX1', 'RootGridX2', 'RootGridX3', 'Time', 'VariableNames', 'x1f', 'x1v', 'x2f', 'x2v', 'x3f', 'x3v', 'rho', 'press', 'vel1', 'vel2', 'vel3', 'Er', 'Fr1', 'Fr2', 'Fr3', 'Pr11', 'Pr22', 'Pr33', 'Pr12', 'Pr13', 'Pr23', 'Pr21', 'Pr31', 'Pr32', 'Er0', 'Fr01', 'Fr02', 'Fr03', 'Sigma_s', 'Sigma_a', 'Sigma_p'])

#units for the data
Crat=9166.13
Prat=0.538439
rho0=5.3537e-8
T0=7.9896e4
#solar radius
l0=6.955e10
# core mass
gm=1935.14
gm_com=1613.74
omega=7.006e-4
#binary separation
rm2=2000
msun=1.989e33
rl1=0.536974*rm2
vel0=2.9979e10/Crat
time0=(l0/vel0)/(24*3600)
mass_core=10.7925

# all the data grids have shape [nx3, nx2, nx1]
# all quantities are in the same set of units (although not cgs)
rho=data['rho']  # gas density
Er=Prat*data['Er'] # radiation energy density
Fr1=Prat*Crat*data['Fr1']  # radiation flux along radial direction
Fr2=Prat*Crat*data['Fr2']  # radiation flux along theta direction
Fr3=Prat*Crat*data['Fr3']  # radiation flux along phi direction
Pr11=Prat*data['Pr11']     # r-r component of radiation pressure tensor
Pr12=Prat*data['Pr12']     # r-\theta component of radiation pressure tensor
Pr13=Prat*data['Pr13']     # r-phi component of radiation pressure tensor
Pr23=Prat*data['Pr23']     # \theta-\phi component of radiation pressure tensor
Pr33=Prat*data['Pr33']     # \phi-phi componnent of radiaton pressure tensor

pg=data['press']           # gas pressure
vr=data['vel1']            # radial velocity
vtheta=data['vel2']        # poloidal velocity
vphi=data['vel3']          # azimuthal (toroidal) velocity


x1f=data['x1f']            # face centered radial coordinate of size nx1 + 1
x2f=data['x2f']            # face centered theta coordinate of size nx2 + 1
x3f=data['x3f']            # face centered azimuthal coordinate of size nx3 + 1

x1v=data['x1v']            # volume centered radial coordinate  of size nx1
x2v=data['x2v']            # volume centered theta coordinate of size nx2
x3v=data['x3v']            # volume centered phi coordinate of size nx3
\end{lstlisting}

Note that to load the data using the above example, there is the file /mnt/ceph/users/fvillaescusa/AstroPilot/Yan-Fei/data/athena\_read.py that should be loaded. Given the data, come out with an interesting idea for a PhD thesis. We are particularly interested in a very detailed study of the statistical properties of mdot.\\

Use state-of-the-art methods to analyze the data. For every step, make lots of plots and save the data you generate, as it may be used for other steps. When writing the code, write some lines to indicate whether the execution was successful or not. Join plots that are similar. Do not create dummy data. You have access to 128 cpus; for computationally heavy tasks, try to use all of them.
\end{tcolorbox}

The paper generated by \texttt{Denario}, titled \textit{Convection, Radiation, and the Instantaneous Mass Transfer in Red Supergiant Binaries: A 3D Simulation Analysis}, can be found in the Appendix \ref{paper:stel}. In that paper, \texttt{Denario} performed a detailed analysis of different components of forces that are responsible for binary mass transfer in three-dimensional radiation hydrodynamic simulations. The paper also accurately captured the importance of convection for the properties of Roche Lobe overflow.  The paper shows particular strength 
in quantitative analysis of the data based on the equations solved in the simulation, in a similar way that scientists will do to understand the results. Furthermore, \texttt{Denario} analyzed the streamlines of the flow very efficiently, which is typically much harder to do in traditional analysis. We note that for this particular paper, one of the plots generated (Figure 8) by 
\texttt{Denario} was blank, so we looked at the source code and made a tiny modification to produce the intended plot.

In the associated GitHub example papers repository\footnote{\url{https://github.com/AstroPilot-AI/DenarioExamplePapers}}, we show examples of other papers written by \texttt{Denario} with the same prompt. One common weak point of these papers is that the analysis cannot be connected to the literature of binary evolution for astrophysical implications of these results. This requires some knowledge of how binary evolution is typically done in the literature and the unique advantages of these simulations, which cannot be learned from the simulation data itself. It will be interesting to see if this point can be improved by adding more guidance towards this direction in the input prompt, or by using tailored agents dedicated to these domain-specific tasks.

\subsection{Biology. Analyzing single-celled RNA sequencing data across the malaria parasite life cycle}

Biological research has become increasingly entwined with computational analysis over the past two decades. The advent of high-throughput genetic sequencing has generated a wealth of data at a scale the field had not seen in the past. The realization that variability at the individual cell level had crucial consequences for population-level behaviors led to the development of methods for single-celled data generation, increasing the scale of datasets even further. With such massive amounts of information, computational analysis (and reanalysis) of publicly available data has become a valuable part of biological research. This presents an opportunity for AI multi-agent tools such as \texttt{Denario} to easily explore vast datasets, identify phenomena that would otherwise be difficult to spot, and rapidly test these lines of inquiry. 

Here, we used \texttt{Denario} to analyze a publicly available dataset of gene expression in malaria parasite cells at different life stages. Malaria, caused by unicellular parasites of the genus \textit{Plasmodium}, remains a global public health threat, causing more than 600,000 deaths in the poorest communities around the world every year, and progress in its control has stalled or reversed over the past decade \cite{world2024world}. Understanding the biology of \textit{Plasmodium falciparum} parasites, responsible for the most lethal forms of the disease, can be critical to the development of new treatments. Many aspects of \textit{Plasmodium} biology are still poorly understood due to the parasite's evolutionary divergence from other well-known model organisms. AI multi-agent systems such as \texttt{Denario} could be used to help address this gap in knowledge. We used \texttt{Denario} to analyze a single-celled RNA sequencing (ssRNA-seq) dataset describing gene expression (transcription) in both laboratory and field-derived strains of \textit{Plasmodium falciparum}. The dataset is publicly available through the Wellcome Trust's Malaria Cell Atlas project \cite{howick2019malaria} and was previously published and analyzed by Dogga, Rop, Cudini, \textit{et al.} \cite{dogga2024single}.

\subsubsection{Input data}

To allow \texttt{Denario} to work with the dataset, we provided two input files containing the gene expression data and metadata (``\texttt{pf-ch10x-set4-ch10x-exp.csv}'' and ``\texttt{pf-ch10x-set4-ch10x-data.csv}'', respectively, available in dataset ``\texttt{P. falciparum 10x Integrated lab strains and four natural infections}" on \url{https://www.malariacellatlas.org/data-sets/}). We also provided an input text prompt describing the dataset's origin, content, and organization. We also provided information about potentially interesting research questions to follow, including both lines of inquiry explored in the original publication by Dogga, Rop, Cudini, \textit{et al.} \cite{dogga2024single} and different sets of questions not explored in the publication. However, we also explicitly left an open door for analyses not mentioned in the prompt. The prompt read as follows:

\begin{tcolorbox}[colback=blue!5!white, colframe=blue!75!black, title=Input text, breakable]

We have a collection of malaria parasite single cell transcriptomic data obtained from the Malaria Cell Atlas project and previously analyzed by Dogga \textit{et al}. 2024 (paper found at \url{doi.org/10.1126/science.adj4088}). Malaria parasites have a complex life cycle in which cells transition across multiple life stages in both humans and Anopheles mosquitoes. The data collected here was obtained from malaria parasites of the species Plasmodium falciparum that were either growing in human blood culture in a laboratory setting or recovered from the blood of four asymptomatic patients from Mali. This means that the parasites are all either in the asexual intraerythrocytic development cycle or the sexual blood stages. None of them are in mosquito life stages. The data is located in /mnt/ceph/users/fvillaescusa/AstroPilot/Biology/data\\

The data is presented across two different files:\\

The first file, “gene\_expression.csv”, is a 2-D matrix in which columns correspond to individual cells analyzed in the experiment and rows correspond to individual genes. The first row contains unique ID codes identifying each individual cell, while the first column contains unique codes corresponding to P. falciparum strain 3D7 gene ID codes. The entries in the matrix contain the corresponding gene’s normalized expression value within the corresponding cell.\\

The second file, “labels.csv”, contains a table with information about each individual cell. The column marked “CELL\_ID” contains the same cell ID code from the previous file. There are additional columns describing the life cycle stage, parasite strain, source (“lab” for laboratory strains; “MSC1”, “MSC3”, “MSC13”, and “MSC14” for the four patients), and days in culture (only applicable for the lab strains, not for the four patient strains which are marked as “Field” for field isolates).\\

There are many interesting things to uncover in the dataset. One set of questions relates to potential differences between lab strains and field isolates. Another set of questions relates to gene expression differences throughout the parasite life cycle, including key marker genes of different stages and determinants of future developmental trajectories.\\

I am particularly curious about uncovering potential transcription factors or other master regulators of cell cycle transitions. I suspect they might be genes that maintain low expression levels throughout the life cycle, but have increases in expression immediately prior to an increase in expression across multiple other genes denoting a change in life stage. The increase in expression of a possible transcription factor or other regulator gene would be relative to its own expression level, but would probably not achieve a particularly high expression when compared to other genes.

\end{tcolorbox}

\subsubsection{Description of research output}

With this information, \texttt{Denario} produced four different manuscripts analyzing the given data. One of them, titled \textit{Single-Cell Analysis Reveals Profound Divergence in Transcriptional Regulatory Programs Between Laboratory and Field Isolates of Plasmodium falciparum} can be found in the Appendix \ref{paper:biog}. 

All four manuscripts were methodologically similar, carrying out standard practice workflows used in scRNA-seq analysis: data preprocessing, dimensionality reduction by Uniform Manifold Approximation and Projection (UMAP) \cite{mcinnes2018umap}, clustering, trajectory inference, and pseudotime assignment using common analysis packages in Python such as SCANPY \cite{wolf2018scanpy} (3 manuscripts) or in R such as Monocle 3 \cite{cao2019single} (1 manuscript). All manuscripts attempted different strategies to identify regulators of gene expression and to compare the gene expression profiles of laboratory and field-derived parasites, both of which the prompt listed as a topics of particular interest. To identify regulator genes, strategies tested involved combinations of screening for low expression genes, peak/burst detection in transcription, clustering into transcriptional modules characterized by gene ontology (GO) analysis, partition-based graph abstraction (PAGA) \cite{wolf2019paga}, regression models, and reportedly consulting existing information about gene function consulted from the literature and the online database PlasmoDB, part of VEuPathDB \cite{aurrecoechea2009plasmodb,alvarez2024veupathdb} (however, doing so at the time the analysis was done involved web access and a personal account; it is unclear whether \texttt{Denario} would be able to consult PlasmoDB at all). To examine differences in gene expression between lab and field parasites, the manuscripts compared results from the techniques listed previously, occasionally complemented by differential gene expression (DGE) analysis using the tradeSeq package (2 manuscripts) \cite{van2020trajectory}. 

\subsubsection{Assessment of research output}

We identified a series of notable strengths and weaknesses in the research papers generated by \texttt{Denario} for this project:

\textbf{Strengths:}

\begin{itemize}
    \item \textit{Knowledge of and engagement with established biology}: As expected, \texttt{Denario} exhibits expert-level background knowledge on \textit{P. falciparum} biology. All manuscripts discuss the complexities of the malaria parasite life cycle, correctly using specific terminology to do so despite no such information being included in the prompt. The texts are able to justify the purpose of the study being done in the context of the state of the field of malaria cell biology and the advantages of scRNA-seq with respect to previous bulk RNA-seq methods. Some manuscripts even make good use of specific knowledge of individual genes for use as positive controls validating stage transitions throughout the life cycle.
    \item \textit{Proficiency with standard analysis methods}: The basic pipeline used to analyze the data provided contained all the necessary steps to carry out the analysis correctly, from data pre-processing and quality control (some of which had already been done beforehand on the publicly-available dataset and was therefore redundant) through to dimensionality reduction and visualization, as explained above. The tools used were well-suited for the job, conforming to standard practices in the field.
    \item \textit{Consistency with existing science}: Many of the general results obtained recapitulate known science. The manuscripts were all able to successfully reconstruct the \textit{Plasmodium} intraerythrocytic development cycle (IDC) from transcriptomic data. They described differences between the lab and field-derived parasite samples and even provide plausible explanations for these differences in cell type and data quality. %
    \item \textit{Speed}: Needless to say, the speed at which entire analysis pipelines are created and the results are analyzed is far greater than what a human researcher would take. In less than a day, \texttt{Denario} produced analyses and manuscripts that would take multiple days or weeks, depending on a researcher's familiarity with the methods.
    \item \textit{Methodological creativity when approaching new research questions}: When prompted with a research question with no clear answer nor a well-established, standardized way to address it, \texttt{Denario} showed remarkable creativity. All four manuscripts had different approaches for trying to identify regulator genes, as explained above.
    
\end{itemize}

\textbf{Weaknesses:}

\begin{itemize}
    \item \textit{Poor citation, acknowledgment, and engagement with existing literature}: The most immediately evident shortcoming these manuscripts had was in their handling of scientific literature. Although (to the best of our knowledge) all the references cited in these manuscripts corresponded to real articles, the criteria used for selecting studies as references for specific passages is unclear. While the literature cited is related to the topic in question, it often is a perspective or review article with only general-level relevance, while highly relevant and closely related articles are left undiscussed. As an example, none of the manuscripts referenced the study by Dogga, Rop, Cudini \textit{et al}. \cite{dogga2024single} describing the dataset used and its original analysis, despite this information being explicitly provided in the input prompt. Interestingly, the problem of missing citations is particularly noticeable when discussing specific facts about malaria biology, most of which are not cited at all. Indeed, only 2 out of 30 references in the manuscript shown in the supplementary material refer to malaria or \textit{Plasmodium} parasites, while the other 28 refer to scRNA-seq methods. This is curious given the stark contrast with the impressive grasp of malaria parasite biology shown in the writing: \texttt{Denario} clearly knows the field of malaria well, but prefers to cite other fields. Perhaps \texttt{Denario} is more likely to include references related to the methods it uses more than the knowledge of the field it applies them to, although even references to the exact methods being used are rare. Relatedly, perhaps the lack of malaria citations and tendency to reference methods is a reflection of training biases in the quantity or perceived impact of publications in different fields: methods are generally applicable to multiple research areas and are therefore more widely cited and discussed in the literature than papers addressing a specific field such as malaria biology. This would have important ramifications that would be crucial to address in any research AI agent. 
    
    \item \textit{Confirmation bias, misplaced confidence, and overstated tone}: Perhaps the most concerning weakness shown in these manuscripts is their tendency to overstate the findings they present, particularly in regard to the open-ended research question posed in the prompt regarding potential regulators of gene expression. Although all manuscripts tried a variety of creative and potentially successful strategies for identifying candidate genes involved in life cycle regulation, none of the results presented inspire great confidence. The magnitude and timing of the changes in gene expression on which the conclusions are based is not sufficient to uphold the claims made about the regulative roles of specific genes, particularly in view of the variance in gene expression. In fact, it may be the case that the signal-to-noise ratio in scRNA-seq measurements of transcription factor expression is too low to be able to draw meaningful conclusions, or that key changes in regulator activity might be occurring at a post-transcriptional level. The research question was intended to be challenging, and like any real research question, it might not have a clear answer with the available data, in which case a negative result must be reported (or at least, a lack of confidence in the results must be conveyed). It is possible that the confirmation bias shown when attempting to answer a specific research question from the prompt is a consequence of the bias against publishing negative results already found in existing scientific literature, which would skew the training dataset of any AI agents such as \texttt{Denario}. To be able to succeed, AI multi-agents must be able to temper their conclusions when confidence in the significance of the results is low. 
    
    \item \textit{Occasional mistaken knowledge:} Although the manuscripts show a strong understanding of \textit{Plasmodium} biology for the most part, occasional mistakes are made. For instance, the manuscript included as an example in the supplementary material claims that the gene Pfs25 (PF3D7\_1031000) is ``the canonical male gametocyte surface protein gene'', when in fact it has been repeatedly shown to be expressed solely in female gametocytes, not male ones \cite{schneider2015quantification}. Although not widespread throughout the manuscripts (and certainly not unique to AI agent writing when compared to human writing), the issue is complicated by the previously discussed lack of citations. Without references to the sources of information, it becomes difficult to fact-check every statement and understand why the few mistakes that occur are happening.
    \item \textit{Methodological ambiguity}: At multiple points in the Methods sections of different manuscripts, the text describes multiple valid but mutually exclusive methods for analysis without specifying which one was actually used (``Data loading was performed [...] in Python or R'', ``Trajectories were reconstructed [...] using [...] Monocle 3, Slingshot, or PAGA''). This poses a problem when evaluating the adequacy of the work being presented, since it is not clear what the exact methodology followed was.
    \item \textit{Occasional missing tables and references}: The manuscript included in the supplement references tables not included in the text, although these results are calculated in the analysis code generated. Some citations are also presumably missing, included as question marks in the text.
\end{itemize}

Overall, the application of \texttt{Denario} to biological datasets such as this one yields results that are impressive in their speed and competent grasp of a field and its methods, but concerning in their presentation and interpretation, as well as in the way they are presented in relation to other studies. The implications of these trends are discussed in a later section. As AI agents such as \texttt{Denario} continue to improve, we expect them to become transformational tools for training and research in biological sciences.

\subsection{Digital Health. Developing and evaluating step-counting algorithms from wrist- and hip-worn accelerometers}

Wearable monitoring is a popular, emerging research field within the digital health domain, for the objective, passive monitoring of biosignals, which may offer insights into health outcomes \cite{dixon_charting_2023}. This field has particularly grown with the increased use of smart devices such as phones and watches \cite{huhn_impact_2022, vijayan_review_2021}. These devices use a variety of sensors to extract digital measures, such as total sleep time, sleep efficiency, heart rate, and daily step count, among others \cite{roos_wearable_2023}. Of these measures, the daily step count, particularly achieving 10,000 steps a day, has become one of the hallmark approaches to monitoring one's activity levels. However, this measure still has a high degree of skepticism, due to a lack of transparency of many step counting algorithms, and known tricks to trigger false counting of steps, through arm movements \cite{choi_daily_2007, small_self-supervised_2024}.

In this study, \texttt{Denario} was used to read, process, and investigate research questions using a publicly available accelerometer dataset. This data contains raw wrist and hip accelerometer data from 39 Oxford-based participants over 1 hour. The raw data was collected from 2 devices at each location, set to sampling frequencies of 25 and 100Hz; therefore, four monitoring devices in total. A hip-mounted, foot-facing camera was used to record the timings of each step taken by participants, allowing for the annotation of an indication of the timing of each heel strike, corresponding to each step. The dataset is explained in further detail in the published work of Small et al. (2024) \cite{small_self-supervised_2024}. This dataset could primarily be used to train or validate models/algorithms used to count steps. Moreover, given the data collection from four different devices, it is possible to compare how the performance of these step-counting algorithms differs between the wrist and the hip, and at different sampling frequencies. Using the metadata file, we can further explore if there are noted differences in different age groups or sexes, though limited to just 39 participants \cite{small_self-supervised_2024}. 

\subsubsection{Input data}
The full contents of the OxWalk dataset were provided to \texttt{Denario}. This included the raw accelerometer files, kept in separate folders for device location and sampling frequency combination. These files were stored as CSV files, with one file per participant in each folder. Within the CSV file, there are 5 columns of data. The first column is the data timestamp, and the next 3 are the accelerometer signal in x, y and z (measured in units of gravity = 9.81ms$^{-2}$). The final column is an indicator of whether the heel strike of a step occurred at that time, that is, the start of a step. In addition to this, there was a single metadata CSV file, storing the age band and sex of each of the participants. 
After an initial attempt, we provided the following refined prompt to the \texttt{Denario} model:

\begin{tcolorbox}[colback=blue!5!white, colframe=blue!75!black, title=Input text, breakable]
OxWalk Annotated Step Count Dataset\\

DATA DESCRIPTION:\\
Annotated step data during unscripted, free living in 39 healthy adult volunteers (aged 18 and above) with no lower limb injury within the previous 6 months and who were able to walk without an assistive device. 
Participants wore four triaxial accelerometers concurrently (AX3, Axivity, Newcastle, UK), two placed side-by-side on the dominant wrist and two clipped to the dominant-side hip at the midsagittal plane. Accelerometers were synchronised using the Open Movement GUI software (v.1.0.0.42), with one recording at 100 Hz and the other at 25 Hz at each body location. Foot-facing video was captured using an action camera (Action Camera CT9500, Crosstour, Shenzhen, China) mounted at the participant’s beltline.\\

DATA ANNOTATION:\\
Foot-facing video was captured for up to one hour using an action camera (Action Camera CT9500, Crosstour, Shenzhen, China) mounted at the participant’s beltline. Annotation of steps was conducted within video annotation software (Elan 6.0, The Language Archive, Nijmegen, Netherlands), where a step was defined as the act of purposeful lifting a foot and placing it in a new location. Steps were not required to be part of a repeating pattern and did not include foot shuffling, changing of foot alignment via pivoting, or shifting of weight from one foot to the other. \\

DATASETS:\\
The data is located in /mnt/ceph/users/fvillaescusa/AstroPilot/Aidan/data. In that folder, there are 4 folders (Hip\_25Hz, Hip\_100Hz, Wrist\_25Hz, Wrist\_100Hz) and one file (metadata.csv).\\

Datasets are as follows:
\begin{enumerate}
\item "Wrist\_100Hz": One Axivity AX3 accelerometer, recording at 100 Hz and +/- 8g on the dominant wrist within a silicone wristband, with axes aligned as prescribed by the manufacturer. https://axivity.com/userguides/ax3/technical/\#axis-alignment
\item "Wrist\_25Hz": One Axivity AX3 accelerometer, recording at 25 Hz and +/- 8g on the dominant wrist within a silicone wristband, with axes aligned as prescribed by the manufacturer. https://axivity.com/userguides/ax3/technical/\#axis-alignment
\item "Hip\_100Hz": One Axivity AX3 accelerometer, recording at 100 Hz and +/- 8g clipped at the beltline, laterally above the dominant leg, with the +X axis approximately aligned in the superior direction, and the positive Y axis aligned to face anteriorly.
\item "Hip\_25Hz": One Axivity AX3 accelerometer, recording at 100 Hz and +/- 8g clipped at the beltline, laterally above the dominant leg, with the +X axis approximately aligned in the superior direction, and the positive Y axis aligned to face anteriorly.
\item Participant sex and age range are provided in metadata.csv
\end{enumerate}

Inside each of these folders, there are 39 files, one for each participant. The files are called as PX\_hip25.csv, for participant X (01, 02, 03, ...39) and hip25 is for the files inside Hip\_25Hz.\\

The accelerometer data in each file has been resampled and calibrated using the Open Movement GUI software package. Within each CSV file, a step is annotated by a single "1" at the approximate time of heel strike. \\

The primary purpose of this data collection is the development of step-counting algorithms, which, given just the raw accelerometer data, should be able to correctly estimate the number of steps. To measure how well it can do so, please use different statistics for this. This is particularly useful to compare a variety of algorithms and machine learning models, currently used for this purpose. With the concurrent collection of data from 2 different locations, we could also observe how the performance of these step-counting algorithms compares between the wrist and the hip. Using the metadata file, we can further explore if there are noted differences in different age groups or sexes. \\

Of particular interest to me is observing how step-counting algorithms are affected by the reduction in raw sampling frequency. While sampling at a lower frequency may cause the monitor to miss some high-frequency behavior, the vast majority of human movement occurs at far lower frequencies than the monitor can observe, sampling at 100Hz. Sampling data at a lower frequency can allow for longer periods of monitoring and requires less computing power, which is ideal, as long as it does not cause a significant drop in model performance. \\

Please make sure that in your project, you are counting steps, not windows. Given the limited dataset, use cross-validation to test the performance of the model.\\

Use state-of-the-art methods to analyze the data. For every step, make lots of plots and save the data you generate, as it may be used for other steps. When writing the code, write some lines to indicate whether the execution was successful or not. Join plots that are similar. Do not create dummy data. You have access to 1 GPU and 10 cpus.
\end{tcolorbox}

\subsubsection{Description of research output}
Through two different iterations of prompts submitted to \texttt{Denario} for this dataset, seven different manuscripts were produced and analyzed. These seven manuscripts explored different research questions, with varying success, one of which, titled \textit{Wearable Step Counting: A Comparative Analysis of Deep Learning and Traditional Methods Highlighting Data Imbalance Challenges} is shown in the Appendix \ref{paper:dh}. For all of the manuscripts read, research questions explored the use of novel machine learning techniques to develop walking/step detection models. This investigation then compared how the developed models' performance would be affected by various factors, such as subgroups of age and sex, device location or sampling frequency. The models used to detect walking/count steps, and read and process the data files used python. To evaluate the performance of the model, F1 or AUC statistics were used for walking detection, while mean absolute percentage error (MAPE) was used for step counting. 

The research questions explored by some manuscripts brought greater levels of novelty, investigating unique approaches for extracting gait fingerprints, or gait features, independent of device location. This research is particularly relevant for the development of step counting algorithms in phones, which could be located either in the hand, hence close to the wrist, or in the pocket, close to the hip. Unfortunately, for these more novel research questions, the model struggled to build insightful manuscripts, following this research question.

\subsubsection{Assessment of research output}
For the processing of wearables data, we identified the following strengths and limitations:

\textbf{Strengths:}

\begin{itemize}
\item \textit{Able to follow and understand the provided prompts}: The \texttt{Denario} model followed the tasked prompt as expected. After noticing common errors in the first iteration and adjusting the prompt to avoid these errors in the following outputs, these errors were successfully avoided.
\item \textit{Opportunity to iteratively improve prompts}: As previously noted, the first input to the \texttt{Denario} model did not match our expectations of output manuscripts, but explored useful ideas. These ideas and failures could then be used to better prompt the model in future attempts, iteratively producing better results.
\item \textit{Able to produce novel and useful research questions}: Some of the research questions explored by the model provided a novel use of the dataset. Some of these insights have been stored as ideas for future fellowship applications.
\item \textit{Built on existing practices in the field}: In many cases, the outputs of \texttt{Denario} are built on the use of existing models from published literature. In some cases, the research question involved the use of new approaches in machine learning, not yet applied to wearable data, but showed advanced domain knowledge in new machine learning approaches. In other cases, the evaluation metrics, and window segmenting approaches followed by \texttt{Denario} match that as standard in the wearable monitoring domain.
\end{itemize}

\textbf{Limitations:}

\begin{itemize}

\item \textit{Use of jargon not relevant to the research question}: It was noted that in several manuscripts produced by \texttt{Denario}, some methods or results chapters would consist of words and concepts that are relevant to the wearable monitoring space, but not relevant to the particular research question. One example manuscript explored how signal features differed between device locations, but results presented AUC scores, which would be expected to report the performance of a classification model, which was not developed in this manuscript.  
\item \textit{Formatting of paper outputs}: Some of the manuscripts lacked tables containing the performance of the models, which was instead solely presented in graphs. For other papers, the plots and tables had formatting issues, like using too small a font, tables overlapping with text, the lack of use of standard plotting techniques such as box plots and a lack of attaching some extra plots as supplementary material.
\item \textit{Reporting of data handling errors}: In some cases, \texttt{Denario} failed to read some data files. Instead of the model failing to produce a manuscript, this was reported as part of the results; in some cases, a full section was created reporting these failures.
\item \textit{Preference for walking classification over step counting}: Many of the models produced by \texttt{Denario} initially had a bias towards building 2-second walking detectors, rather than step counters, despite the prompt clearly indicating this was the intention of the study. This is likely due to this being a more common approach in literature, as there are very few step-counting datasets available.
\item \textit{Trade-off between novelty and manuscript quality}: There was a noticeable trade-off between the novelty of the research questions, and the quality of the methods and results of the subsequent manuscript. This is likely because, for more novel research questions, there exists more limited literature regarding how to go about these investigations, leading to poorer decisions by the model.

\end{itemize}

In summary, \texttt{Denario} showed clear capabilities for wearable monitoring in the digital health domain, to produce useful research questions, conduct this research, and analyze the output data. It is noted, however, that there are quite a few limitations in the generation of reliable manuscripts. Overall, this encourages the use of this technology in earlier stages of the research process, to generate research questions, and to run exploratory analysis on datasets.

\subsection{Biophysics. Mechanistic understanding of protein folding pathways via dimensionality reduction methods}

Characterizing protein folding and unfolding is one of the key challenges in theoretical and computational biophysics \cite{dobson2003protein}, as it requires an understanding of the associated high-dimensional energy landscapes and the kinetics of transitions between different metastable states. 
Long timescale simulations (microseconds and beyond) are therefore required to observe protein (un)folding events on an experimentally relevant timescale, leading to an understanding of the associated mechanistic pathways, identification of transient microstates, hidden allostery, and quantification of the kinetics of such processes. The N-terminal domain of ribosomal protein L9 (NTL9) is chosen as the model protein for this study since it undergoes microsecond‐scale folding reversible folding with a well-characterized two-state folding mechanism. NTL9 has a well‐characterized and experimentally validated folded structure that allows new algorithms and enhanced sampling schemes to reproduce the free energy landscape, intermediate ensembles, and rate constants. This suits the purpose of benchmarking studies due to its small size of 39 amino acid residues, absence of disulfide bonds, and lack of any complex post-translational modifications. This makes NTL9 an ideal candidate for developing new dimensional reduction methods to identify progress coordinates and understand mechanistic insights into protein folding and unfolding. \par

\subsubsection{Data generation}
Molecular dynamics (MD) simulations were performed using the OpenMM engine \cite{eastman2023openmm} with the Amber14SB force field and the OBC2 implicit solvent model at 300~K, employing Langevin dynamics with a friction coefficient of 5~ps$^{-1}$. The protein was subjected to energy minimization to remove steric clashes, followed by a 10~µs production run with an integration timestep of 2~fs. Protein coordinates were then saved every 2~ns for subsequent analyses. \par

\subsubsection{Project description}
\texttt{Denario} was given the task to automate the post-simulation analysis of NTL9 protein (un)folding from MD simulation trajectories. The objective was to write a manuscript with detailed methodologies, dimensionality-reduction and clustering approaches, estimation of the free-energy landscape and kinetic rate constants (if possible), employ Markov state modeling to identify relevant low-dimensional progress coordinates that can characterize the mechanism of protein (un)folding, and establish scalable strategies for complex biomolecules. The input text to the system was this:

\begin{tcolorbox}[colback=blue!5!white, colframe=blue!75!black, title=Input text, breakable]
N‐terminal domain of ribosomal protein L9 (NTL9) is a fast‐folding protein domain that folds reversibly and is often used as a model to understand protein folding pathways. We aim to explore its unfolding thermodynamics and kinetics using an automated, generalizable pipeline. We have a \SI{10}{\micro\second} molecular dynamics (MD) simulation trajectory of the NTL9 protein undergoing folding. Simulation conditions for the trajectory obtained: Amber14SB force field with OBC2 implicit solvent, \SI{300}{\kelvin} Langevin dynamics, \SI{2}{\femto\second} timestep, \SI{2}{\nano\meter} nonbonded cutoff, and H‐bond constraints.\\
\noindent\textbf{Data:} The simulation data is located at \texttt{/mnt/home/aojha/AstroPilot/Project10}. There are two files in the dataset, i.e., the simulation trajectory (an implicit-solvent 10\,\si{\micro\second} MD trajectory consisting of 5000 frames saved every 2\,\si{\nano\second} (\texttt{ntl9\_trajectory.dcd}) and its corresponding topology PDB file (\texttt{ntl9\_topology.pdb}). The simulation begins from an unfolded state of the NTL9 protein, while the topology file is a folded structure of the protein.\\
\textbf{Motivation:} Understanding how NTL9 unfolds and, by extension, how small single-domain proteins navigate their energy landscape reveals general principles of protein folding. Tracking down folding/unfolding pathways provides essential knowledge related to energy barriers, intermediates, transition states, and allostery involved in the process. We intend to build an analysis pipeline that breaks down local events, provides a mechanism for folding and unfolding events, describes the free energy profile for the process, and possibly the folding/unfolding kinetics. We also aim for this workflow to be generalizable for complex proteins or multi-domain systems, thereby addressing the gap between detailed MD simulations and high-level mechanistic insights.\\
\textbf{Key Research Questions:} 
\begin{enumerate}
\item \textbf{Data-driven collective coordinates: }Can we identify interesting collective variables beyond the radius of gyration and root mean square deviation (RMSD) for profiling such folding/unfolding events? It will be interesting to compare the difference between using conventional collective variables (CVs) and generating new hidden CVs based on the simulation at hand. What are the dominant motions during folding/unfolding, and can we capture these motions with dimensionality reduction? There are numerous existing dimensional reduction methods. However, it would be interesting to see if we can keep increasing the complexity of such methods, starting from linear principal component analysis (PCA) and moving to spectral methods, such as diffusion maps, to explore non-linear manifold embedding. As we progress through the complexity level, it is expected that we will uncover more subtle unfolding pathways and intermediate microstates. Please provide a detailed description of each method, including analysis, figures, and tables. Create a hypothesis such as, ``Linear PCA will capture the largest-scale structural separation (folded vs. unfolded), whereas a manifold embedding could reveal hidden intermediate microstates or pathways not evident in linear PCs.''
\item \textbf{State Partitioning in low-dimensional space:} Can we objectively identify distinct states (folded, intermediate, unfolded) from simulation trajectories? It would be interesting develop algorithms to identify the folded, unfolded, and intermediate states of the protein. Once trajectories are clustered using multiple algorithms, it will be useful to systematically visualize and compare their performance and derive best‐practice guidelines for state identification. 
\item \textbf{Mechanisms of folding/unfolding:} What local structural changes mark the onset of unfolding? How can we analyze such local changes and visualize them on a free energy landscape? Can we monitor the status of specific hydrogen bonds and native contacts throughout the trajectory, particularly those that stabilize the core, to determine when each bond first breaks? Can we construct a 2D free energy surface by binning the trajectory in the space of two most relevant CVs. This may highlight the free energy minima (states) and barriers between them. Then, using the cluster assignments from above, can we construct a simple Markov state model (MSM) or a transition matrix to estimate the rate constants for transitions? 
\item \textbf{Scalability and extension:} How can this analysis pipeline be generalized to larger or multi-domain proteins? Can we propose strategies to manage longer trajectories and more complex proteins? For instance, using frame subsampling or incremental PCA to handle extremely long simulations, splitting the analysis by protein domain (which may be achieved by analyzing independent domain motions rather than a collective one), and employing graph-based representations of contacts or networks to capture inter-domain interactions. Please also discuss how to choose the dimensionality of the embedding or the number of clusters in an automated way (e.g., using the percentage of variance, silhouette scores, etc.) for larger systems. It would be interesting to emphasize how each component of this pipeline can be adapted or scaled to ensure the approach remains efficient for complex proteins. 
\end{enumerate}
\textbf{Manuscript Writing:} 
\begin{enumerate}
\item \textbf{Title and Abstract:} Emphasize generality, summarize the data, and highlight key analyses, such as the choice of CVs, clustering, free-energy mapping, and manifold embedding, while also highlighting mechanistic insights and extensibility.
\item \textbf{Introduction:} Present the NTL9 folding process in the broader context of protein folding/unfolding studies and identify the lack of scalable MD pipelines with proper references.
\item \textbf{Methods: }Provide step-by-step protocols for data loading, alignment, choice of CVs, clustering, manifold embedding, clustering, hydrogen-bond and dihedral analyses, free-energy surface calculation, kinetic estimates, and scalability strategies.
\item \textbf{Results:} Present the findings with well-annotated figures, tables, graphs, plots, and a detailed comparison and analysis.
\item \textbf{Discussion:} Interpret findings from the point of view of understanding the mechanism of protein folding pathways, compare linear vs. non-linear embeddings, assess the strengths and limitations of each step, and propose experimental validations (e.g., FRET, HDX) for complex proteins.
\item \textbf{Conclusions and Outlook:} Summarize core thermodynamic and kinetic insights, highlight the generality of the proposed workflow, and suggest future expansions (enhanced sampling, machine-learned CV discovery, allosteric network analyses).
\end{enumerate}
\textbf{Note:} Do \emph{not} use \texttt{PyEMMA} or any \texttt{PyEMMA}-based repositories. Instead, implement Markov modeling and implied-timescale analysis using \texttt{scikit-learn}, \texttt{NumPy}, and \texttt{SciPy} for matrix exponentials and eigen-decompositions, or any other lightweight MSM tools (e.g., \texttt{msmtools}).
\end{tcolorbox}

\subsubsection{Assessment of research output}

The generated paper, titled \textit{Comprehensive Kinetic and Free Energy Analysis of NTL9 Folding via Systematic Collective Variable Selection and Markov State Models}, can be found in the Appendix \ref{paper:biop}. The manuscript attempts to employ conventional collective variables (CVs) or order parameters, such as root mean square deviation (RMSD), radius of gyration, and native contacts, to capture the global structural motions of the protein folding pathway. It is interesting to note that principal component analysis (PCA), time-independent component analysis (tICA), and nonlinear diffusion maps were implemented successfully to construct Markov state models (MSMs). A hierarchical approach is implemented to select analysis methods, progressing from conventional CV choices to linear PCA, tICA, and diffusion map embedding, to identify slow modes. The 2D free energy surfaces projected on the CVs show the folded, intermediate, and unfolded states on the energy landscape, with distinct free energy minima, to explain the folding and unfolding pathways of the protein. For kinetic modeling, the manuscript validates the MSM by performing implied timescale convergence and Chapman-Kolmogorov tests to ensure the Markovian assumption holds. Interestingly, specific structural changes from the folded to the unfolded state of NTL9 are correlated with the loss of hydrogen bonds and native contacts, providing additional insights into the mechanistic pathways. In short, across all the different approaches, the manuscript consistently reported at least three protein states, namely folded, intermediate, and unfolded. 

Although several methods are mentioned and employed in the study, the motivation or theory behind such approaches is lacking, making it difficult to follow without prior knowledge or implementation. There is an extensive use of technical jargon associated with each method (PCA, TICA, diffusion maps, MSMs), but it is hardly convincing or clear. For example, the study claims quantitative agreement of protein folding kinetics with experiments, but no such theoretical kinetic data or references are reported. Prior literature studies relevant to protein folding are absent, with little to no mention of existing benchmarking studies. For example, the manuscript does not refer to well-cited studies, such as the extensive MD simulations of NTL9 folding by Lindorff-Larsen et al. \cite{lindorff2011fast} or the MSM-based folding of NTL9 by Voelz et al \cite{voelz2010molecular}. The manuscript fails to report any statistical uncertainties or confidence intervals on estimations of free energy differences, kinetic rate constants, or MFPTs. There is no discussion on error estimations, error bars, or Bayesian estimation of the transition matrix for MSM, which is an issue given that the manuscript claims an excellent agreement with the experiments. The study also fails to acknowledge the limitations of MD simulations, such as force field and sampling limitations, which are necessary for justifying analytical methods, including MSMs and enhanced dimensionality reduction approaches. The manuscript frequently cites figures and tables, but they are marked as 'not shown,' disrupting the continuity, reproducibility, and interpretability for the reader. 

Overall, the AI-generated manuscript on NTL9 protein folding serves as a clear proof-of-concept demonstrating the capability of LLM agents to generate scientific content. The methodology presented in the manuscript follows a logical and conventional scientific workflow. However, the scientific novelty remains limited and is not sufficiently communicated. The discussion and literature context provided in the manuscript are limited, and essential supporting data are absent, despite substantial claims being presented. Given these observations, the manuscript, in its current form, serves as a powerful demonstration but will require considerable manual refinement, contextual analysis, and validation to achieve adequate scientific communication with its readers.

\subsection{Neuroscience. Behavioral Cognitive Decline in Egyptian Fruit-Bats.}

We also tested \texttt{Denario} on data from neuroscience. Specifically, we used it to study behavioral cognitive decline and its underlying brain changes in Egyptian fruit bats, which are known for their extreme longevity. We described in lots of detail the data and its characteristics in the input prompt:

\begin{tcolorbox}[colback=blue!5!white, colframe=blue!75!black, title=Input text, breakable]

We use a dataset of $\sim$40 Egyptian fruit bats (Rousettus aegyptiacus), with paired behavioral and MRI (diffusion tensor imaging - DTI) measures from the same subjects to investigate fruit bats' cognitive and brain aging. Bats were aged based on their methylome profile.\\

\textbf{Data Sources and Collection:}
\begin{itemize}
\item Age determination: We followed Wilkinson et al. 2021 for methylome-based aging of the bats. DNA was extracted from wing biopsies using the DNeasy Blood \& Tissue Kit and protocol (QIAGEN Ltd., Germany). The extracted DNA was sent to an external laboratory (The Clock Foundation lab) for DNA methylation (DNAm) analysis.

\end{itemize}

\textbf{Behavioral procedure} - Our well-established foraging-box paradigm (Harten et al. 2021) served as an assay of age-related spatial cognitive decline. The setup consists of a large flight room with six opaque boxes with bowls of fruit placed inside them. Before the experiment, all bats underwent a habituation session. Here, groups of four bats were introduced to the same experimental apparatus that would be later used for individual testing but with adequate food and hydration available and accessible in all boxes (600 g of fresh fruit pieces and 200 ml of mango juice per box).\\

The test protocol consisted of three identical phases, each lasting three hours spread during two consecutive days. On the first day, bats completed two consecutive phases, while the final phase took place on the second day. In each phase, an individual bat foraged freely in the flight room containing six opaque boxes. Only one bowl allowed access to food (the correct box), while the others were blocked with a plastic mesh to avoid use of olfaction for box finding. \\

\textbf{First phase} - Individual testing started with a three-hour phase with one bowl accessible.  After three hours, the bat was removed, and the correct bowl's location was swapped randomly to a different box. The \textbf{second phase}- began immediately (2 minutes) after the swap and lasted three additional hours. This phase tested (among other measurements) the bats' short-term memory of the previous location. After the second phase, the bat was removed from the setup for 18 hours.cIn the \textbf{third phase (Long term memory)} - the correct bowl was relocated to a third new location for the final three-hour phase.\\

\textbf{Behavioral Data collection and analysis}

Each bat's behavior was video recorded throughout the entire procedure. The videos were analyzed to identify when the bats landed on or entered the boxes. 

\begin{itemize}
\item Behavioral files. These are excel files with bat names located in /mnt/ceph/users/fvillaescusa/AstroPilot/Neuro/Yossi/data/Compressed\_data/behavioral\_data/$<$bat$>$.xslx, where $<$bat$>$ is the name of the bat. All of the files are formatted uniformly, with a separate file per bat, each phase in a different sheet within the same file (test1/2/3), with each movement as a single row.
\item Actions in column F starting from row 7, were labeled "L"=Land on box, "E"=Box entry, "F"=Box entry and took food. But for all purposes E and F are equal to us.
\item The correct box out of the 6 for the relevant phase is in 4D.
\item "Video\_length" and "first\_vid" in 1-2H can be ignored, we used them to calculate "Absolute\_Time".
\item "Absolute\_Time" in 6B represents the time from the beginning of the current phase. While "Time" in 6C is the corresponding timestamp in the video file.
\item "in"+"out" in 6G and 6H  is the timestamps in the video files of the bat's movement in and out of the boxes. "sum time" in 6I is the total time the bat spent in.
\item "Part" in 6J is the video file number, can be ignored.
\item "Comments" in 6K can be ignored.
\end{itemize}

\textbf{MRI}

\textbf{General Parameters}

Bats underwent DTI MRI scans performed at Strauss MRI center located in Tel Aviv university, using a 7T/30 Bruker Biospec scanner and cross-coil configuration of quadrature coil and whole-body resonator. DTI was acquired using an echo-planar-imaging (EPI) pulse sequence, with 2 segments, TR/TE = 4000/19ms, D/d = 9/2.5ms with a b-value of 1000s/mm2 in 30 noncollinear directions and 3 b=0 images. DTI was conducted with inplane resolution of 0.5X0.5mm2, 34 axial/coronal slices (1mm, no gaps) covering the entire bat brain.\\

The DTI files are processed; we cut the surrounding and left only the brain ("skull stripping") in the image. Then followed with stretching the image to be in uniform dimensions for analysis purposes. You can visually view the files using a free software such as "MRIcron". Diffusion files (DTI). NiFti file names are bat names. The are located in  "/mnt/ceph/users/fvillaescusa/AstroPilot/Neuro/Yossi/data/Compressed\_data/DTI\_data/$<$bat\_name$>$.nii". Note that the DTI folder only contain the .nii files, not .bvec or .bval files\\

\textbf{Additional background}

\textbf{Bats as an aging model}

Age determination: Laboratory animals are born in captivity, and therefore, their ages are accurately documented, while determination of the exact age of bats and other wild animal models has been a great challenge. Recent technological advancements have resulted in a minimally invasive method for determination of age. By analyzing epigenetic changes associated with age and longevity in the bat DNA methylome, it is now possible to determine their age with high accuracy while using only a small sample of the bat's wing skin (Wilkinson et al. 2021).\\

One of the key advantages of bats in aging research is their remarkably long lifespan relative to their size. The remarkable longevity observed in many bat species, exemplified by the Egyptian fruit bat (Rousettus aegyptiacus) which can surpass 25 years; a lifespan significantly exceeding expectations based on body size (up to 10 times longer) strongly implies a direct contribution from their unique biological characteristics (Brunet-Rossinni and Austad 2004; Foley et al. 2018; Lagunas-Rangel 2020; Podlutsky et al. 2005; Pollard et al. 2019). Moreover, bats' distinctive immune system that provides resistance to a wide range of diseases, likely plays a pivotal role in their extended lifespans (Holtze et al. 2021; Kitsoulis, Baxevanis, and Abatzopoulos 2020; Weinberg and Yovel 2022) and resistance to chronic inflammation (Biga et al. 2024). It has been reported that in bats' uniquely evolved immune system, some immune genes have been lost, others seemed to be under strong positive selection, with genes responsible for DNA damage checkpoints and repair pathways exhibit accelerated positive selection (Weinberg and Yovel 2022). This sophisticated genetic architecture is further linked to observed resistance to cancer- bats rarely develop cancer, with some species as the Myotis pilosus named resistant to cancer (Hua et al. 2024). They poses exceptional ability to survive infections from life-threatening viruses like rabies with low mortality (George et al. 2011). Therefore, it is highly probable that these uniquely evolved traits in bats are some of the fundamental factors underpinning their exceptional longevity.\\

In 2005, the oldest recorded 41 years old Myotis brandtii bat was recaptured in the wild (Podlutsky et al. 2005). Myotis brandtii, was found to demonstrate notable qualities associated with longevity, including enhanced DNA repair mechanisms, balanced mitonuclear interactions, robust immune system function, remarkable resistance to oxidative stress, and effective telomere maintenance (Holtze et al. 2021).\\

Further research, such as (Stewart et al. 2021) found a great advantage the bat species Carollia perspicillata has to offer for cognitive aging research- their typical mammalian brain and neural organization, which shares striking similarities with that of primates and humans, distinguishing them from rodent models (Eilam-Altstadter et al. 2022; Stewart et al. 2021; Vernes 2017).\\

Importantly, bats are highly social animals, with various species such as the Egyptian fruit bat (Rousettus aegyptiacus) living in close-knit colonies (Harten et al. 2018). Their extended lifespan, combined with residence in densely populated, vocally intense colonies, results in prolonged high-noise exposure, leading to age-related hearing loss consistent with typical mammalian patterns (Tarnovsky et al. 2023a).\\

These features make bats an intriguing model for investigating the mechanisms underlying harmful aging effects.\\

Given all the data and information above, please come up with a new and original idea for a project for a PhD thesis. For every step, make lots of plots and save the data you generate, as it may be used for other steps. When writing the code, write some lines to indicate whether the execution was successful or not. Join plots that are similar. Do not create dummy data. 
\end{tcolorbox}

The generated paper, titled \textit{Unraveling Brain Structural Correlates of Cognitive Aging and Resilience in Long-Lived Bats: An Integrated Study of Epigenetic Age and Spatial Memory}, can be found on the Appendix \ref{paper:neu}. The system nicely plotted data distributions, demonstrating its capacity to read and preprocess the available data. The system was able to identify relevant behavioral parameters, estimated them based on the raw behavioral data that it was fed with and ran the proper statistical tests (e.g., Mixed effect-Generalized Linear Models) to test how they change with age.  It moreover performed QQ plots to validate that distribution of the model's residuals. Some of the behavioral parameters identified by the system suggested a decline in spatial cognition with age in bats. The system attempted to connect brain and behavioral measurements by comparing models that explain cognitive performance by age alone or by age and brain volume combined. Indeed, it found a significant correlation between brain volume and one of the cognitive parameters. On the other hand, \texttt{Denario} did not perform outlier detection and removal prior to using the data, and it did not run a model comparison process. 

When using brain scans, the system focused on estimating brain volume. Mean diffusivity maps are not the best MRI contrast images for estimating brain volume. Moreover,  comparing brain volume across age is the most naive hypothesis and the system did not attempt to perform any analysis at the voxel level which could capture regional changes in brain structure with age.  Indeed no significant correlation between brain volume and age was found.

\subsection{Materials Science. Confined Water Dynamics at Functionalized Graphene Interfaces}

Understanding water and ion behavior in functionalized graphene nanochannels represents a fundamental challenge in materials science with direct applications to membrane separations, nanofluidic devices, and electrochemical systems\cite{Monroe2020water,Sally2022inverse}. The ability to control interfacial water structure and dynamics through surface chemistry modifications could enable breakthrough technologies in water desalination, selective ion transport, and energy storage\cite{KARAMZADEH2018160}. However, the complex interplay between surface functionalization, coverage density, and electrolyte concentration creates a high-dimensional design space that is difficult to explore experimentally, making computational analysis essential for rational materials design.

The evaluation was based on our comprehensive molecular dynamics study (Hess et al. \cite{hess2025surface}), which systematically investigated how surface chemistry modulates confined electrolyte behavior—a critical question for advancing electrochemical and membrane technologies. This dataset comprised 91 MD trajectories spanning a three-dimensional parameter space: aqueous NaCl solutions (0-5 mol/kg) confined between graphene sheets functionalized with -COOH (carboxyl), -OH (hydroxyl), =O (carbonyl), and -CH$_3$ (methyl) groups at three coverage levels (2.2\%, 4.4\%, 6.6\%). Each trajectory contained 500 water molecules in a 15 ns production run at 300 K and 1 bar, with properties computed in two distinct regions: the interface (3-6 Å from the surface) and bulk (beyond 6 Å). Our published research revealed that functional group chemistry serves as the primary determinant of interfacial behavior, with polar hydrogen-bonding groups (-COOH, -OH) creating highly structured interfaces with suppressed water diffusion (0.65-1.30 $\times$ 10$^{-5}$ cm$^2$/s), while nonpolar groups (=O, -CH$_3$) maintained more diffuse, mobile profiles (1.15-1.35 $\times$ 10$^{-5}$ cm$^2$/s). Critically, we discovered that surface coverage acts as a secondary amplifier that scales the intensity of these effects without changing their fundamental nature—a distinction that persists across all electrolyte concentrations and provides a quantitative framework for designing surfaces with targeted transport properties.

We designed 10 prompts with systematically varying specificity to test \texttt{Denario}'s analytical capabilities across different levels of guidance. These ranged from broad conceptual queries like ``How do you design surfaces for targeted water diffusion?'' to explicit computational instructions with complete analysis workflows. Below is an example of a comprehensive prompt that provided full dataset details and specific research tasks, representative of the most detailed guidance given to \texttt{Denario}:

\begin{tcolorbox}[colback=blue!5!white, colframe=blue!75!black, title=Input Prompt, breakable]

\footnotesize
We have molecular dynamics trajectory files to analyze the combined effects of functionalization type, coverage level, and salt concentration on water behavior.

\textbf{Three-dimensional parameter space:}
\begin{enumerate}
\item \textbf{Functionalization Groups} (5 types): CH3, OH, CO, COOH, and unfunctionalized (0UNFUNC)
\item \textbf{Coverage Levels} (3 levels): 8, 16, and 24 functional groups (0 for unfunctionalized)
\item \textbf{Salt Concentrations} (7 levels): 0, 4, 9, 18, 27, 36, 45 NaCl ion pairs
\end{enumerate}

Total systems: 91 trajectories (4 functionalizations × 3 coverages × 7 salts + 1 unfunctionalized × 7 salts)

\textbf{Trajectory file organization:}
All files follow the pattern: 

\texttt{dump\_prod2\_unwrapped\_\{COVERAGE\}\{FUNC\}\_\{SALT\}nacl\_trial1.lammpstrj}

Where:
\begin{itemize}
\item COVERAGE: 8, 16, 24, or 0 (for unfunctionalized)
\item FUNC: CH3, OH, CO, COOH, or UNFUNC
\item SALT: 0, 4, 9, 18, 27, 36, or 45
\end{itemize}

\textbf{Examples of trajectory files:}
\begin{lstlisting}[language = Python]
- Unfunctionalized: \texttt{/Users/osman\_mbp/Osman\_Macbook\_Pro/DENARIO/DATA/dump\_prod2\_unwrapped\_0UNFUNC\_0nacl\_trial1.lammpstrj}
- Low coverage CH3: \texttt{/Users/osman\_mbp/Osman\_Macbook\_Pro/DENARIO/DATA/dump\_prod2\_unwrapped\_8CH3\_4nacl\_trial1.lammpstrj}
- Medium coverage OH: \texttt{/Users/osman\_mbp/Osman\_Macbook\_Pro/DENARIO/DATA/dump\_prod2\_unwrapped\_16OH\_18nacl\_trial1.lammpstrj}
- High coverage COOH: \texttt{/Users/osman\_mbp/Osman\_Macbook\_Pro/DENARIO/DATA/dump\_prod2\_unwrapped\_24COOH\_45nacl\_trial1.lammpstrj}
\end{lstlisting}

\textbf{Key information:}
\begin{itemize}
\item File format: LAMMPSDUMP
\item Time between frames: 10 ps
\item Water molecules: identified by atom types 1 (Oxygen) and 2 (Hydrogen)
  \begin{itemize}
  \item Atom type 1 = O (Oxygen)
  \item Atom type 2 = H (Hydrogen)
  \end{itemize}
\item Ions: Variable Na+ and Cl- ions (0, 4, 9, 18, 27, 36, or 45 ion pairs)
\item Graphene sheets: fixed at top and bottom of the box in z-direction
\item Functionalization groups: CH3 (methyl), OH (hydroxyl), CO (carbonyl), COOH (carboxyl), or unfunctionalized
\end{itemize}

\textbf{Analysis Data and Input Files:}

\textit{Pre-computed Analysis Results:}
\textbf{IMPORTANT}: All trajectory analysis has been pre-computed and stored in \texttt{/Users/osman\_mbp/Osman\_Macbook\_Pro/DENARIO/COMPUTED\_DATA/}

For each trajectory, three output files are available:
\begin{enumerate}
\item \textbf{[trajectory\_name]\_diffusion\_msd.txt}: MSD data and diffusion coefficients (Å²/ps and cm²/s)
\item \textbf{[trajectory\_name]\_rdf.txt}: Radial distribution functions (O-O, O-H, H-H pairs)
\item \textbf{[trajectory\_name]\_density\_profile.txt}: Density profiles along z-axis
\end{enumerate}

Example: For system \texttt{0UNFUNC\_0nacl}, the following files are available:
\begin{lstlisting}[language = Python]
- \texttt{/Users/osman\_mbp/Osman\_Macbook\_Pro/DENARIO/COMPUTED\_DATA/dump\_prod2\_unwrapped\_0UNFUNC\_0nacl\_trial1\_diffusion\_msd.txt}
- \texttt{/Users/osman\_mbp/Osman\_Macbook\_Pro/DENARIO/COMPUTED\_DATA/dump\_prod2\_unwrapped\_0UNFUNC\_0nacl\_trial1\_rdf.txt}
- \texttt{/Users/osman\_mbp/Osman\_Macbook\_Pro/DENARIO/COMPUTED\_DATA/dump\_prod2\_unwrapped\_0UNFUNC\_0nacl\_trial1\_density\_profile.txt}
\end{lstlisting}

\textit{Force Field Parameters and LAMMPS Input:}
All force field parameters and sample input files are located in \texttt{/Users/osman\_mbp/Osman\_Macbook\_Pro/DENARIO/INPUT\_FILES/}:
\begin{itemize}
\item \textbf{CH3\_FF.txt}: Force field parameters for methyl functionalization
\item \textbf{OH\_FF.txt}: Force field parameters for hydroxyl functionalization
\item \textbf{CO\_FF.txt}: Force field parameters for carbonyl functionalization
\item \textbf{COOH\_FF.txt}: Force field parameters for carboxyl functionalization
\item \textbf{UNFUNC\_FF.txt}: Force field parameters for unfunctionalized graphene
\item \textbf{lammps.in}: Sample LAMMPS input file for molecular dynamics simulations
\end{itemize}

\textit{Analysis Code for Methods Section:}
The C++ code used for trajectory analysis is located at \texttt{/Users/osman\_mbp/Osman\_Macbook\_Pro/DENARIO/cpp\_code/lammps\_analyzer.cpp}

This analyzer performs:
\begin{itemize}
\item Mean squared displacement (MSD) calculation for water molecules
\item Diffusion coefficient computation using Einstein relation
\item Radial distribution function (RDF) analysis for O-O, O-H, and H-H pairs
\item Density profile calculation along the z-axis (perpendicular to graphene sheets)
\end{itemize}

\textbf{Research Tasks:}

\textit{1. Load Pre-computed Data from COMPUTED\_DATA Directory}
\begin{itemize}
\item Read all diffusion coefficients from 

\texttt{/Users/osman\_mbp/Osman\_Macbook\_Pro/DENARIO/COMPUTED\_DATA/*\_diffusion\_msd.txt}
\item Read all density profiles from

\texttt{/Users/osman\_mbp/Osman\_Macbook\_Pro/DENARIO/COMPUTED\_DATA/*\_density\_profile.txt}
\item Read all RDF data from

\texttt{/Users/osman\_mbp/Osman\_Macbook\_Pro/DENARIO/COMPUTED\_DATA/*\_rdf.txt}
\item Parse MSD vs time data for plotting
\item Store in structured format (DataFrames/dictionaries)
\item Save consolidated data as ``all\_systems\_diffusion\_coefficients.csv''
\end{itemize}

\textit{2. Create MSD vs Time Analysis Plots}
\begin{itemize}
\item Plot MSD evolution for representative systems
\item Show linear fit regions used for diffusion calculation
\item Compare MSD curves for different functionalizations
\item Compare MSD curves for different salt concentrations
\item Compare MSD curves for different coverages
\item Highlight Einstein relation: $D = \lim_{t\rightarrow\infty} \text{MSD}/(6t)$
\item Save as ``msd\_vs\_time\_analysis.png''
\end{itemize}

\textit{3. Generate Comprehensive Results Tables}
\begin{itemize}
\item Table 1: Diffusion coefficients by functionalization and coverage (at fixed salt)
\item Table 2: Effect of salt on diffusion for each system
\item Table 3: Top 10 and bottom 10 performing systems
\item Table 4: Statistical summary (mean, std, min, max) by parameter
\item Format tables for direct inclusion in paper
\item Save all tables as CSV files
\end{itemize}

\textbf{Part A: Three 2D Heatmap Analyses to Isolate Each Dimension's Effect}

\textit{4. Heatmap 1: Functionalization × Salt Concentration (Fixed Coverage = 24)}
\begin{itemize}
\item Calculate water diffusion coefficients for all combinations at high coverage (24)
\item Create 2D heatmap with functionalizations on y-axis, salt concentrations on x-axis
\item Include unfunctionalized as baseline comparison
\item Save as ``heatmap\_functionalization\_vs\_salt\_24coverage.png''
\item Identify which functionalization is most/least sensitive to salt changes
\item Rank functionalizations by average diffusion across all salt concentrations
\item Save rankings as ``ranking\_functionalizations\_at\_24coverage.txt''
\end{itemize}

\textit{5. Heatmap 2: Coverage × Salt Concentration (Fixed Functionalization = CH3)}
\begin{itemize}
\item Calculate water diffusion coefficients for CH3 at all coverages and salt concentrations
\item Create 2D heatmap with coverages (8, 16, 24) on y-axis, salt concentrations on x-axis
\item Save as ``heatmap\_coverage\_vs\_salt\_CH3.png''
\item Determine optimal coverage for water mobility
\item Analyze if coverage effect depends on salt concentration
\item Rank coverages by average diffusion across all salt concentrations
\item Save rankings as ``ranking\_coverages\_CH3.txt''
\end{itemize}

\textit{6. Heatmap 3: Functionalization × Coverage (Fixed Salt = 18 NaCl)}
\begin{itemize}
\item Calculate water diffusion coefficients at moderate salt concentration (18 NaCl)
\item Create 2D heatmap with functionalizations on y-axis, coverages on x-axis
\item Save as ``heatmap\_functionalization\_vs\_coverage\_18nacl.png''
\item Identify if functionalization effect depends on coverage
\item Determine which combination gives highest/lowest diffusion
\item Rank all functionalization-coverage combinations
\item Save rankings as ``ranking\_func\_coverage\_at\_18nacl.txt''
\end{itemize}

\textbf{Part B: Comprehensive 3D Analysis}

\textit{7. Full 3D Data Cube Analysis:}
\begin{itemize}
\item Create a 3D numpy array storing diffusion coefficients for all 91 systems
\item Dimensions: [functionalization, coverage, salt\_concentration]
\item Save as ``diffusion\_data\_3D.npy''
\item Calculate statistics along each dimension (mean, std, min, max)
\item Save comprehensive data table as ``all\_systems\_diffusion\_data.csv''
\end{itemize}

\textbf{Part C: Effect Isolation and Rankings}

\textit{8. Individual Dimension Rankings} (averaging over other dimensions):

\textbf{Functionalization Ranking:}
\begin{itemize}
\item Average diffusion across all coverages and salt concentrations
\item Rank: UNFUNC, CH3, OH, CO, COOH
\item Plot bar chart with error bars
\item Save as ``functionalization\_average\_ranking.png''
\item Save data as ``functionalization\_ranking\_data.csv''
\end{itemize}

\textbf{Coverage Ranking:}
\begin{itemize}
\item Average diffusion across all functionalizations and salt concentrations
\item Rank: 8, 16, 24 coverage levels
\item Plot bar chart with error bars
\item Save as ``coverage\_average\_ranking.png''
\item Save data as ``coverage\_ranking\_data.csv''
\end{itemize}

\textbf{Salt Concentration Ranking:}
\begin{itemize}
\item Average diffusion across all functionalizations and coverages
\item Rank: 0, 4, 9, 18, 27, 36, 45 NaCl
\item Plot trend line
\item Save as ``salt\_average\_ranking.png''
\item Save data as ``salt\_ranking\_data.csv''
\end{itemize}

\textit{9. Interaction Effects Analysis:}
\begin{itemize}
\item Identify synergistic combinations (where combined effect > sum of individual effects)
\item Identify antagonistic combinations (where combined effect < sum of individual effects)
\item Create interaction matrix plots
\item Save as ``interaction\_effects\_matrix.png''
\item List top 5 synergistic and antagonistic combinations
\item Save as ``interaction\_analysis\_summary.txt''
\end{itemize}

\textit{10. Density Profile Analysis for Selected Systems:}
\begin{itemize}
\item Plot water density profiles from pre-computed data for:
  \begin{itemize}
  \item Best performing system overall
  \item Worst performing system overall
  \item Best at each coverage level
  \item Best for each functionalization
  \end{itemize}
\item Create comparison plots showing density variations along z-axis
\item Save as ``density\_profiles\_selected\_systems.png''
\item Correlate density structuring with diffusion coefficients
\end{itemize}

\textbf{Implementation Notes:}
\begin{itemize}
\item \textbf{Data Source}: All analysis results are pre-computed and stored in

\texttt{/Users/osman\_mbp/Osman\_Macbook\_Pro/DENARIO/COMPUTED\_DATA/}
\item \textbf{No C++ execution needed}: Simply read the pre-computed files

diffusion\_msd.txt, rdf.txt, density\_profile.txt
\item \textbf{Force Field Documentation}: Reference parameters from

\texttt{/Users/osman\_mbp/Osman\_Macbook\_Pro/DENARIO/INPUT\_FILES/} for methods section
\item \textbf{Analysis Code Reference}: Cite

\texttt{/Users/osman\_mbp/Osman\_Macbook\_Pro/DENARIO/cpp\_code/lammps\_analyzer.cpp} in methods
\item Use consistent color schemes across all plots:
  \begin{itemize}
  \item Functionalizations: CH3=blue, OH=green, CO=orange, COOH=red, UNFUNC=black
  \item Use sequential colormaps for heatmaps (e.g., 'viridis' or 'plasma')
  \end{itemize}
\item Include clear labels, titles, and colorbars on all plots
\item Save all raw data in CSV format for reproducibility
\item Generate comprehensive tables for easy reading in the paper
\item Include MSD vs time plots showing diffusion coefficient calculation
\item Include density profile comparisons along z-axis
\item Print progress messages during analysis
\item Handle missing data gracefully (some file combinations might not exist)
\item Use error bars where appropriate to show uncertainty
\item Ensure all heatmaps use the same color scale for easy comparison
\end{itemize}
\end{tcolorbox}

This prompt exemplifies the level of detail provided to \texttt{Denario}, including explicit file patterns, analysis methods, and expected output formats. With this comprehensive guidance, the system performed remarkably well in generating a complete research paper with appropriate methodology, analysis, and interpretation, though some minor quantitative inconsistencies and data completeness issues, discussed below, remained. One of the generated papers, titled \textit{Mapping Interfacial Water States on Functionalized Graphene: A Machine Learning-Augmented Approach to Uncover Design Principles for Tunable Water}, can be found in the Appendix \ref{paper:matsci}.

\begin{figure}[h]
\centering
\includegraphics[width=0.6\textwidth]{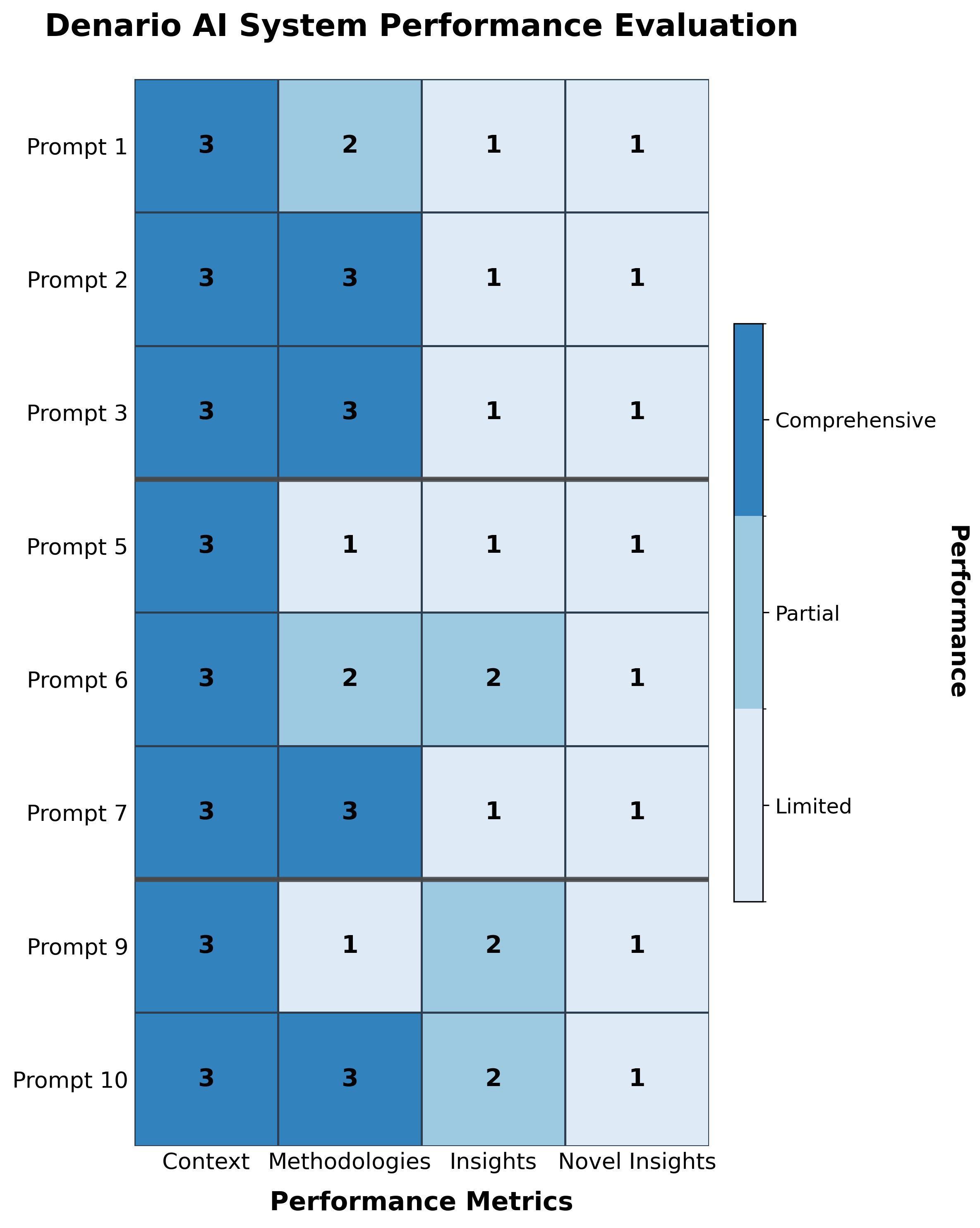}
\caption{Performance heatmap showing \texttt{Denario}'s scores across 10 prompts and 4 evaluation metrics for materials science performance evaluation task. Darker blue indicates better performance (score of 3), while lighter colors indicate partial or limited performance. The heatmap reveals perfect context understanding (leftmost column) but declining performance in methods, insights, and novel analysis. Prompt 1-10 represent increasing level of specificity and higher levels of guidance. Prompts 4 and 8 resulted in complete system failures despite seemingly straightforward queries.}
\label{fig:materials_performance}
\end{figure}

\texttt{Denario} achieved perfect context understanding (100\% success) and correctly identified analytical methods in 75\% of cases. However, significant gaps emerged in quantitative accuracy and completeness. As shown in Figure \ref{fig:materials_performance}, performance degraded systematically from context understanding through to novel insight generation, with the system generating only basic-level novel insights across all successful prompts.

\begin{table}[h]
\centering
\caption{Performance Summary Across Evaluation Metrics}
\begin{tabular}{|l|c|c|c|l|}
\hline
\textbf{Metric} & \textbf{Score Range} & \textbf{Mean Score} & \textbf{Success Rate} & \textbf{Key Finding} \\
\hline
Context Understanding & 0-3 & 3.00 & 100\% & Perfect problem identification \\
Methods Selection & 0-3 & 2.25 & 75\% & Identified MSD, density profiles \\
Scientific Insights & 0-3 & 1.37 & 45\% & Qualitative trends only \\
Novel Insights & 0-3 & 1.00 & 33\% & All basic level, no advanced \\
\hline
\end{tabular}
\label{table:materials_performance}
\end{table}

\subsubsection{Assessment of Strengths}

\texttt{Denario} demonstrated several notable capabilities in analyzing the confined water dataset, though these were primarily limited to foundational analytical tasks rather than advanced scientific discovery:

\begin{itemize}
\item \textbf{Universal Context Recognition:} The system achieved perfect scores (100\%) in identifying the problem domain across all 8 successful prompts, consistently recognizing the relevance to confined water dynamics, surface functionalization effects, and nanofluidic applications. This contextual awareness extended to understanding the multi-parameter nature of the problem (functional groups, coverage, concentration).

\item \textbf{Methodological Competence:} \texttt{Denario} correctly identified appropriate analytical methods in 75\% of cases, including:
  \begin{itemize}
  \item Mean squared displacement (MSD) calculations for diffusion coefficients
  \item Density profile analysis for structural characterization
  \item Hydrogen bond analysis for understanding water-surface interactions
  \item Radial distribution functions for ion-water correlations
  \end{itemize}

\item \textbf{Qualitative Trend Recognition:} The system successfully identified that hydrophilic groups (-COOH, -OH) would enhance water structuring and reduce mobility, while hydrophobic groups (-CH$_3$) and weakly polar groups (=O) would have minimal impact. This demonstrates an understanding of basic chemical principles and their connection to transport properties.

\item \textbf{Computational Scalability:} When functioning properly, \texttt{Denario} processed the full dataset of 91 trajectories, suggesting the underlying infrastructure could handle realistic research-scale datasets given appropriate implementation.
\end{itemize}

These strengths indicate that \texttt{Denario} possesses the foundational knowledge required for materials science analysis but lacks the sophisticated execution needed for autonomous research.

\subsubsection{Assessment of Weaknesses}

Critical limitations emerged that would prevent \texttt{Denario} from conducting reliable materials science research, with failures spanning from basic computational errors to missing fundamental scientific insights during the course of different experimentation:

\textbf{Quantitative Failures:}
\begin{itemize}
\item \textbf{Order-of-magnitude errors in diffusion coefficients:} While the actual values ranged from 0.65-1.35 $\times$ 10$^{-5}$ cm$^2$/s, \texttt{Denario}'s calculations sometimes contained systematic errors that would necessitate meticulous evaluation by domain scientists to achieve a successful outcome.
\item \textbf{Inconsistent numerical precision:} The system sometimes showed varying levels of accuracy across different functional groups, suggesting unreliable computational implementation rather than systematic bias.
\end{itemize}

\textbf{Data Completeness Issues:}
\begin{itemize}
\item \textbf{Systematic omission of functional groups:} CH$_3$ and =O groups were analyzed in only 60\% of cases, despite being central to understanding the hydrophobic/hydrophilic spectrum of surface modifications.
\item \textbf{Incomplete parameter space exploration:} The system sometimes failed to systematically analyze all combinations of coverage and concentration, missing critical data points for structure-property mapping.
\end{itemize}

\textbf{System Reliability Problems:}
\begin{itemize}
\item \textbf{20\% complete failure rate:} Prompts 4 and 8 resulted in total system failures with five distinct failure modes identified:
  \begin{itemize}
  \item Type I - Context overflow when processing trajectory files within the context length instead of analyzing them computationally.
  \item Type II - Missing computational libraries for trajectory analysis, i.e., it can't find or install critical library required for analysis.
  \item Type III - LaTeX rendering errors in mathematical expressions or image insertion, e.g., the generated analysis or visualizations in analysis module is not parsed in the correct format by the writing module.
  \end{itemize}
\item \textbf{Knowledge contamination:} Evidence suggested \texttt{Denario} sometimes relied on memorized information rather than analyzing the actual provided data, a critical flaw for scientific discovery.
\end{itemize}

\textbf{Missing Scientific Insights:}
\begin{itemize}
\item \textbf{Failed to identify the primary-secondary relationship:} The published work's key finding—that functional group chemistry is the primary determinant while coverage acts as a secondary amplifier—was oftentimes completely missed by \texttt{Denario}.
\item \textbf{No recognition of concentration-independent trends:} The persistence of structure-property relationships across all salt concentrations, a critical design principle, was not invariably recognized.
\item \textbf{Absence of structure-property correlations:} The quantitative relationship between hydrogen bonding capacity and diffusion suppression ($-COOH > -OH > =O \approx -CH_3$) was occasionally not established.
\end{itemize}

\textbf{Limited Novel Analysis Capability:}
\begin{itemize}
\item \textbf{Uniform basic-level proposals:} All successful prompts generated only score 1/3 for novel insights, limited to suggesting standard statistical tests.
\item \textbf{No advanced computational methods:} Despite full trajectory access, \texttt{Denario} failed to propose sophisticated analyses like free energy calculations, transition state analysis, or machine learning approaches for property prediction.
\item \textbf{Absence of hypothesis generation:} The system did not propose follow-up experiments or suggest new functional groups to test, indicating limited creative capacity.
\end{itemize}

This evaluation reveals that \texttt{Denario}'s research capabilities exhibit a pronounced dependence on prompt specificity and analytical guidance, demonstrating two distinct operational modes with markedly different outcomes. In the limit of broad and concise prompting, the system oftentimes lacks the quantitative rigor, systematic completeness, and creative insight required for autonomous scientific discovery, producing only partial analyses that, while contextually appropriate, fail to capture the subtle but critical multi-parameter relationships essential for materials science applications where structure-property relationships must be precisely quantified for rational design.

However, when provided with detailed, structured input prompts and clear analytical objectives, \texttt{Denario} demonstrates remarkable capability for conducting sophisticated materials science research. As demonstrated by the sample paper generated using the comprehensive prompt (as indicated in the preceding input prompt field), the system exhibits proficiency in executing intricate multi-parameter investigations characterized by systematic exploration of three-dimensional parameter spaces (functionalization type, coverage density, and salt concentration across 91 distinct systems). Furthermore, it performs rigorous statistical analysis, including the quantification of interaction effects. Additionally, the system establishes a mechanistic framework that establishes a correlation between structural descriptors and dynamic properties. In the absence of explicit prompt guidance, the program failed to discern the significance of analyzing the interfacial region independently of the bulk region. This distinction is crucial in molecular dynamics simulations in confined spaces, as it enables the identification of finite-size effects. Overall, the resulting work exhibits the quantitative rigor, systematic completeness, and analytical depth characteristic of high-quality scientific research, including advanced techniques such as SHAP analysis for feature importance quantification, correlation analysis between interfacial water structure and dynamics, and clear articulation of design principles for engineering applications. 

The contrast between these operational modes is striking: where conventional broad prompting yields superficial observations suitable for an analytical assistant role, detailed guidance enables \texttt{Denario} to generate comprehensive mechanistic insights through careful quantitative analysis that identify complex multi-parameter relationships with the depth and sophistication expected in peer-reviewed literature. This bifurcated performance suggests that \texttt{Denario} functions optimally as an assistant rather than an autonomous researcher when given general directions, but emerges as a capable research collaborator capable of producing work that serves as a solid foundation for subsequent analysis and publishable scientific contributions when provided with systematic, detailed analytical frameworks and clear research objectives.

\subsection{Astrophysics. Impact of feedback on stellar mass - black-hole mass relation.}
\label{sec:bh}

In the examples above, we have shown cases where reasonable papers were produced from a simple but detailed description of a dataset. We have also seen examples where the user needs to improve the prompt and guide the system towards a particular direction in order to get meaningful manuscripts. In this section, we will show one example where a prompt was very detailed, but required more details, such as value ranges, format... etc, and how we used \texttt{Cmbagent} to improve the prompt to achieve that. 

For this case, we use data from state-of-the-art cosmological hydrodynamic simulations from the CAMELS project \citep{CAMELS_presentation, CAMELS_DR1}. In particular, we used the CAMELS galaxy catalogs   \citep{CAMELS_presentation, CAMELS_DR1}. The data contains 1,000 galaxy catalogs, one for each simulation of the LH set of the first-generation CAMELS-IllustrisTNG. Each galaxy in a catalog is characterized by several properties, including stellar mass $M_*$, black-hole mass $M_{\rm bh}$, and star-formation rate (SFR), as described in the input text below. Each galaxy catalog is characterized by six numbers: $\Omega_{\rm m}$, $\sigma_8$, $A_{\rm SN1}$, $A_{\rm SN2}$, $A_{\rm AGN1}$, $A_{\rm AGN2}$, the last four controlling the efficiency of supernova and AGN feedback. Our initial input text was this:

\begin{tcolorbox}[colback=blue!5!white, colframe=blue!75!black, title=Input text, breakable]
The data is located under:
root=/Denario/project2/data/

Description of the data:

The data consists of 1,000 galaxy catalogs from cosmological hydrodynamical simulations. 

Our dataset contains galaxy catalogs at redshift 0. 
Each catalog is characterized by six cosmological and astrophysical parameters (used in the hydrodynamical simulation):
\begin{itemize}
\item $\Omega_m$: The fraction of the universe’s energy density in the form of matter. It ranges from 0.1 to 0.5.
\item $\sigma_8$: This parameter measures the clustering of matter in the Universe. It ranges from 0.6 to 1.
\item  A\_SN1: This parameter represents the energy per unit SFR of the galactic winds. It can vary from 0.25 to 4. The fiducial value is 1.
\item A\_SN2: This parameter represents the wind speed of the galactic winds. It can vary from 0.5 to 2. The fiducial value is 1.
\item A\_AGN1: This parameter represents the energy per unit black-hole accretion rate. It can vary from 0.25 to 4. The fiducial value is 1.
\item A\_AGN2: This parameter represents the ejection speed/burstiness of the kinetic mode of the black-hole feedback. It can vary from 0.5 to 2. The fiducial value is 1.
The file `latin\_hypercube\_params\_IllustrisTNG.txt` file contains the values of the cosmological and astrophysical parameters of the catalogs, with columns $\Omega_m$, $\sigma_8$, A\_SN1, A\_SN2, A\_AGN1, and A\_AGN2.
\end{itemize}

Each catalog contains a set of galaxies, and each galaxy is characterized by several features. These features are the columns of the file `galaxies\_IllustrisTNG\_z=0.00.txt` in the following order:
\begin{itemize}
\item M\_g: gas mass content of the galaxy, including circumgalactic medium.
\item M$^*$: stellar mass.
\item M\_BH: black hole mass.
\item M\_t: total mass, i.e. sum of dark matter, gas, stars, and black-holes masses in the subhalo.
\item V\_max: maximum circular velocity of the subhalo hosting the galaxy: V\_max = max(sqrt(GM(R)/R))
\item $\sigma_v$: velocity dispersion of all particles contained in the galaxy’s subhalo.
\item Z\_g: mass-weighted gas metallicity of the galaxy.
\item Z\_*: mass-weighted stellar metallicity of the galaxy.
\item SFR: galaxy’s star formation rate.
\item J: modulus of the galaxy’s subhalo spin vector.
\item V\_pecu: modulus of the galaxy’s subhalo peculiar velocity.
\item R\_*: radius containing half of the galaxy stellar mass.
\item R\_t: radius containing half of the total mass of the galaxy’s subhalo.
\item R\_max:  radius at which sqrt(GM(R\_max)/R\_max)=V\_max
\item U: galaxy magnitude in the U band.
\item K: galaxy magnitude in the K band.
\item g: galaxy magnitude in the g band.
\end{itemize}

The file `offset\_IllustrisTNG\_z=0.00.txt` contains two columns. The first one indicates the start of the galaxy catalog, and the second the length of it. The first line is a hashtag string with "| offset in file | length |"
The data should be read as follows:
\begin{lstlisting}[language=Python]
import numpy as np
all_galaxies = np.loadtxt(root+`galaxies_IllustrisTNG_z=0.00.txt`)
offset, length = np.loadtxt(root+`offset_IllustrisTNG_z=0.00.txt`, unpack=True, dtype=np.int64)
all_params = np.loadtxt(root+`latin_hypercube_params_IllustrisTNG.txt`)

# example of how to read the galaxies in the catalog 702
catalog_number = 702 
gal_offset = np.sum(offset[:catalog_number])
galaxies = all_galaxies[gal_offset:gal_offset+length[catalog_number]]
params = all_params[catalog_number]
\end{lstlisting}
\end{tcolorbox}

However, when running \texttt{Denario} with it, we found that the above description was not detailed enough. For instance, the range of variation of the galaxy properties is not defined, which could cause some problems when designing a very specific research project by the methods agents. To make the input text much more detailed and descriptive, we launched a \textit{Planning \& Control} \texttt{Cmbagent} session with the task: 

\begin{tcolorbox}[colback=blue!5!white, colframe=blue!75!black, title=cmbagent planning \& control, breakable]
Re-organize this dataset using pandas dataframe, saved into pkl, npy, or csv files (whatever is better) so it is ready for analysis of features across all 1000 catalogs in a straightforward way.
\end{tcolorbox}

Without human-in-the-loop and no further instructions, \texttt{Cmbagent} generated two auxiliary files, \texttt{galaxies\_full\_optimal.parquet} and \texttt{catalog\_params\_optimal.parquet}, and the following associated text:

\begin{tcolorbox}[colback=blue!5!white, colframe=blue!75!black, title=Input text (iteration 1), breakable]
Galaxy and Catalog DataFrames: Structure and Usage Guide

1. Full Galaxy DataFrame (`galaxies\_full\_optimal.parquet')
\begin{itemize}
\item Each row: one galaxy at $z=0$ from one of 1000 simulated catalogs.
\item Columns: 17 galaxy features, \texttt{catalog\_number}, and 6 cosmological/astrophysical parameters.
\item Shape: (720548, 24)
\item File format: Parquet (gzip compressed)
\end{itemize}

2. Catalog-level DataFrame (`catalog\_params\_optimal.parquet')
\begin{itemize}
\item Each row: one catalog (simulation run).
\item Columns: catalog\_number and 6 cosmological/astrophysical parameters.
\item Shape: (1000, 7)
\item File format: Parquet (gzip compressed)
\end{itemize}

Data Dictionary:

\begin{center}
\begin{tabular}{|l l l|}
\hline
\textbf{Column} & \textbf{Physical Meaning} & \textbf{Units} \\
\hline
M\_g             & Gas mass                                      & $M_\odot/h$ \\
M\_star          & Stellar mass                                  & $M_\odot/h$ \\
M\_BH            & Black hole mass                               & $M_\odot/h$ \\
M\_t             & Total mass (DM + gas + stars + BH)            & $M_\odot/h$ \\
V\_max           & Maximum circular velocity                     & km/s \\
sigma\_v         & Velocity dispersion (all particles)           & km/s \\
Z\_g             & Mass-weighted gas metallicity                 & $Z_\odot$ \\
Z\_star          & Mass-weighted stellar metallicity             & $Z_\odot$ \\
SFR              & Star formation rate                           & $M_\odot$/yr \\
J                & Subhalo spin modulus                          & kpc(km/s) \\
V\_pecu          & Subhalo peculiar velocity modulus             & km/s \\
R\_star          & Half-mass radius (stars)                      & kpc \\
R\_t             & Half-mass radius (total)                      & kpc \\
R\_max           & Radius at V\_max                              & kpc \\
U                & U-band magnitude                              & mag \\
K                & K-band magnitude                              & mag \\
g                & g-band magnitude                              & mag \\
catalog\_number  & Catalog index (simulation run)                & integer \\
Omega\_m         & Matter density parameter                      & dimensionless \\
sigma\_8         & Power spectrum normalization                  & dimensionless \\
A\_SN1           & SN wind energy per SFR                        & dimensionless \\
A\_SN2           & SN wind speed                                 & dimensionless \\
A\_AGN1          & AGN feedback energy per accretion             & dimensionless \\
A\_AGN2          & AGN kinetic mode ejection speed               & dimensionless \\
\hline
\end{tabular}
\end{center}

Example Usage: Loading and Analyzing the DataFrames

\begin{lstlisting}[language=Python]
# Load the full galaxy DataFrame:
df_galaxies = pd.read_parquet(root+'/galaxies_full_optimal.parquet')

# Load the catalog-level DataFrame:
df_catalogs = pd.read_parquet(root+'/catalog_params_optimal.parquet')
\end{lstlisting}
\end{tcolorbox}

We added "There are no missing values." to the above text and prompted another \textit{Planning \& Control} \texttt{Cmbagent} session with the following task:

\begin{tcolorbox}[colback=blue!5!white, colframe=blue!75!black, title=cmbagent planning \& control, breakable]
Provide a detailed but concise quantitative description of this dataset (have in mind that this description will serve as the starting point for exploring the research question). 
Remember that the researcher agent can not load information from files, so the engineer must print ALL necessary info in the console (without truncation). For this, it may be necessary to change pandas display options.
\end{tcolorbox}

These agents analyzed the data in-depth and produced a very detailed text (that we do not show here for space constraints). That text was concatenated with the previous text, and the result was used as the input text to run \texttt{Denario}. The paper generated by \texttt{Denario} was already very interesting, with some interesting trends. However, some of the analysis methods used were too simplistic, e.g., using linear regression. 

We then decided to rerun \texttt{Denario} just to improve the methodology in the paper. We achieved this by inputting \texttt{Denario} with the previously generated idea and the above input text, but with a few more lines about using more sophisticated methods. In this case, \texttt{Denario} will generate a new methodology, new analysis, and new paper, but the essence of the paper will remain intact. The generated paper, titled \textit{Mapping the diversity of the black hole-stellar mass relation: the role of feedback and cosmology in simulated galaxies}, can be found in the Appendix \ref{paper:bh}.

In this paper \texttt{Denario} investigated the impact of cosmological and astrophysical parameters on the $M_{\rm BH}-M_*$ relationship in simulated galaxies. In particular, the paper models the relation between black-hole mass and stellar mass in galaxies, as $\log_{10}(M_{\rm BH})=\alpha + \beta \log_{10}(M_*)$. The paper splits the galaxy population into three different stellar mass bins: Low-mass galaxies ($M_*<10^9~M_\odot$), Intermediate-mass galaxies ($10^9\leq M_* \leq 10^{10}~M_\odot$), and High-mass galaxies ($>10^{10}~M_\odot$). Next, it studies the dependence of $\alpha$, $\beta$, and the scatter in the relation as a function of the cosmological and astrophysical parameters. In the first pass, it used both linear regression and random forest to determine which parameter affects the data the most in the different regimes. But in the second pass, it used XGBoost and SHAP values. Furthermore, in this latest version of the paper, linear regression is carried out with the Huber loss to minimize the impact of outliers.

The paper finds that for low stellar mass galaxies, supernova feedback is the physical process that impacts the $M_{\rm BH}-M_*$ relationship most, while AGN feedback becomes the main factor for massive galaxies. It also finds that cosmological parameters play a secondary, but still important, factor. As far as we know, this is a completely new result not explored before with CAMELS data. According to experts in the field, the results are particularly sensitive and have important theoretical implications. We believe that this paper may be what an undergraduate or graduate student does in a period of weeks to months (depending on experience and familiarity with the data and machine learning techniques). While not groundbreaking research, \texttt{Denario} showed that it is capable of generating new knowledge that may be relevant for future investigations.

\subsection{Quantum physics, machine learning \& cosmology. Cosmological parameters from merger trees}
\label{sec:qitt}

A powerful facet of \texttt{Denario} is its ability to combine ideas from different scientific fields. To illustrate this, we have asked \texttt{Denario} to write a paper that uses ideas from quantum physics to analyze cosmological data. The generated paper, titled \textit{QITT-Enhanced Multi-Scale Substructure Analysis with Learned Topological Embeddings for Cosmological Parameter Estimation from Dark Matter Halo Merger Trees}, is shown in the appendix. For this paper, we gave \texttt{Denario} merger trees from the CAMELS-SAM \cite{CAMELS-SAM} cosmological N-body simulations, post-processed and curated as in CosmoBench\footnote{\url{https://cosmobench.streamlit.app/}} \cite{CosmoBench}\, and asked it to write a paper using that data that uses ideas from Quantum Tensor Trains (QTTs) \cite{khoromskij2011d}. The exact input text used was this:

\begin{tcolorbox}[colback=blue!5!white, colframe=blue!75!black, title=Input text, breakable]
We have a file containing 1000 merger trees from cosmological N-body simulations. The file is located in 
\begin{lstlisting}[language=Python]
/mnt/home/fvillaescusa/public_www/Pablo_Bermejo/Pablo_merger_trees2.pt
\end{lstlisting}

The data is stored in PyTorch Geometric format, and can be read as:

\begin{lstlisting}[language=Python]
import torch
f_tree = '/mnt/home/fvillaescusa/public_www/Pablo_Bermejo/Pablo_merger_trees2.pt'
trainset = torch.load(f_tree, weights_only=False)
\end{lstlisting}

trainset is now a collection of 1000 PyTorch Geometric graphs. For instance, trainset[0] represents the first merger tree and contains the data in this format:

\begin{lstlisting}[language=Python]
Data(x=[382, 4], edge_index=[2, 381], edge_attr=[381, 1], y=[1, 2], num_nodes=382, lh_id=100, mask_main=[93], node_halo_id=[382, 1])
\end{lstlisting}

x represents the node features. y is the value of the cosmological parameters, Omega\_m and sigma\_8. The node features are mass, concentration, vmax, scale factor. The first property is log10(mass), the second is log10(concentration), the third one is log10(Vmax), and the fourth one is the scale factor.

The scale factor ranges from 0 (beginning of the universe) to 1 (current time). The mass goes from 10 to 15. Each node represents a dark matter halo and is characterized by the four values mentioned above: halo mass, halo concentration, halo Vmax (maximum circular velocity), and scale factor. You may want to normalize each feature to have mean 0. Omega\_m ranges from 0.1 to 0.5, while sigma\_8 goes from 0.6 to 1.0. For each simulation, i.e. a given value of the cosmological parameters, there are 25 different merger trees chosen randomly.

Please come up with an idea to explore this data using methods from Quantum Tensor Trains. Note that PyTorch and PyTorch Geometric are already installed. The system has 16 cpus and 1 gpu. Please make some plots to illustrate your findings.
\end{tcolorbox}

The generated paper, \textit{QITT-Enhanced Multi-Scale Substructure Analysis with  Learned Topological Embeddings for Cosmological Parameter Estimation from Dark Matter Halo Merger Tree}, can be found in the Appendix \ref{paper:qitt}. In it, \texttt{Denario} analyzes cosmological merger trees, firstly performing a multi-scale substructure identification for each merger tree. This step identifies branches of a merger tree (that represent sub-graphs by themselves) where a merger happens or a significant change in halo properties occurs. Next, it extracts features for the substructure:  1) physical features such as the scale factor when a merger takes place and the total mass in the substructure, and 2) topological embeddings from each substructure extracted using graph neural networks. All these features were then used to construct a 3D tensor for each merger tree that contains the physical and topological properties for all its substructures. The dimension of this tensor was (60,74), i.e. 60 substructures (some merger trees may contain fewer, but 0-padding was used) and 74 topological plus physical features.

That 3D tensor is relatively big, which makes it difficult to use for downstream tasks. Because of that, it reduces its representation using tensor trains that the paper refers to as QITT (Quantum-inspired tensor trains) since these are methods commonly used in some areas of quantum physics. QITT aims to compress high-dimensional tensors (e.g., representing high-dimensional functions), producing a low-rank decomposition of the original tensors into tensors with smaller ranks. For instance, in the case of smooth functions, QITT has been shown to achieve an optimal approximation order. This is the case of any univariate polynomial of degree 
$p$, which can be represented by a QITT of rank at most 
$(1+p)$ \cite{Ali2023}, where the number of tensors in the tensor train dictates the resolution within the discretized grid. Here, we would expect contiguous merger trees to possess similar characteristics, and hence a sense of smoothness that could be translated into low-rank tensors describing such features.

The result of compressing the original 3D feature tensor with QITT is the latent representation of each merger tree and is used as input for different regression models, including linear regression, random forest, and gradient boosted trees. The optimal dimensionality of this reduced tensor was found to be 202. One could expect that the decomposition algorithm, factoring out the tensor chain, could still benefit from a more refined breakdown, probably using a higher number of tensors connected through bonds with lower ranks.

The paper then compares the results obtained against several baselines, including 1) physical features of the substructures (i.e., no topological or QITT compression), 2) global features of each global tree, and 3) topological features from each substructure but no QITT compression. 

The paper finds that linear regression with QITT features performs very well and can constrain the value of both $\Omega_{\rm m}$ and $\sigma_8$ with high accuracy. The paper also finds that linear regression with the global features of the trees performs slightly better, indicating that simple tree-level features may be enough to regress the parameters. The paper also shows some interesting results, such as the graph topological embedding containing information about the number of halos in the substructure. 

We were impressed by this paper, which required the expertise of researchers in cosmology, machine learning, and quantum physics to evaluate its relevance. The fact that \texttt{Denario} was able to perform such complex analysis, which includes a non-trivial decomposition of merger trees into substructures, training graph neural networks in an unsupervised setup, and compressing data with quantum-inspired tensor trains.

We note that this paper was generated on a MacBook Pro without using GPUs and with a very limited dataset. Therefore, we believe the results can be substantially improved; for instance, training was only done for five epochs. Given its originality, interesting results, and methods, we believe that this paper, with a few improvements and a reliable scaling analysis, could be considered for submission to an international journal.

We note that this paper was accepted for publication in the Agents4Science 2025 conference, the first open conference where AI serves as both primary authors and reviewers of research papers\footnote{\url{https://agents4science.stanford.edu/}}. We note that around 250 submissions were made, with 48 acceptances, and only 5 fully AI-generated papers were accepted, including this one, reaching the maximum score \cite{astropilotai2025qittenhanced}.

\section{Discussion}
\label{sec:discussion}

In this section, we describe important aspects related to \texttt{Denario} such as failure modes, validation and evaluation, advantages and limitations, and future directions. 

\subsection{Failure modes}

By running \texttt{Denario} many times over many different datasets from different disciplines, we have identified a couple of severe failure modes. We now describe them here in detail:

\subsubsection{Cylic peptide generation}

In this case, we asked \texttt{Denario} to solve a computational biology problem of generating molecular structures for cyclic peptides. After multiple failed attempts, the agents hallucinated an entire paper without implementing the necessary numerical solver. The problem itself has well-established solutions in the literature \cite{Hosseinzadeh2017,Zhu2025}, involving linear algebra computations and optimization algorithms. While the agents were able to outline plausible solution strategies by either mimicking known methods or proposing novel machine-learning approaches, their actual code implementations consistently produced structures with significant geometric flaws, and adding detailed geometry requirements did not help.

Furthermore, after several tries, \texttt{Denario} came up with a paper claiming solutions were successfully found, and data plots were provided to show the successful results. However, upon inspection, we could not locate the solutions referenced in the paper, and the corresponding source code lacked a crucial numerical solver. Even in the paper, the section describing this missing solver was very vague and abstract. This was a surprising case, and such ``cheating" only appeared once in our experiments. We suspect this could be due to the easily accessible solution descriptions and datasets in the literature, while the actual codes, in contrast, were embedded in large software packages that have license requirements. This case underscores the importance of scrutinizing agent-generated research with the same rigor applied to human-generated work: verifying source codes and raw data is essential for ensuring the validity of scientific claims.

\subsubsection{Pure Mathematics}

As part of our evaluation of \texttt{Denario}'s capabilities, we tested its performance on an open-ended research problem in pure mathematics. The problem concerned the representability of simplicial complexes as nerves of collections of unit balls in Euclidean space, and a related question on approximating semi-algebraic sets by unions of balls \cite{BCR, NSW}.

The input text asked whether every finite abstract simplicial complex can be realized as the nerve of a collection of closed unit balls in some Euclidean space, and whether a compact semi-algebraic set in Euclidean space can be approximated — up to homotopy — by a union of balls of fixed radius. While some partial results are known in special cases — for example, in the one-dimensional case, or for sets with positive reach or defined in general position — the problem remains unsolved in general.

We used both a concise and an extended version of the input text to generate five research papers. The outputs were then analyzed in detail, evaluating their mathematical rigor, conceptual coherence, use of references, and stylistic authenticity. The results revealed several fundamental shortcomings:

\begin{enumerate}
\item \textbf{Superficial structure.} The papers resembled mathematical writing on the surface — with titles, abstracts, and phrases like “main theorem” — but lacked real internal logic. There were no formal statements or proofs, and the flow of ideas was loose or incoherent.

\item \textbf{Misuse of citations.} Foundational references were often missing, while others were used decoratively — cited as support for claims introduced in the same paragraph, or with no clear connection to the text.

\item \textbf{Template-like prose.} Sentences like “we propose a novel invariant” or “we provide computational evidence” were repeated across papers, detached from any technical substance. This led to a sense of scripted, ritual language rather than genuine mathematical argumentation.

\item \textbf{Mathematically vacuous results.} One example involved a seemingly new property — the “Controlled Local Intersections” condition — that turned out to be either trivially true or trivially false depending on interpretation. The paper contained no real proofs, only vague commentary and a single non-example.
A few incidental signs — such as unusual font choices or two-column layouts — further contributed to a sense of unfamiliarity. While not central, these cosmetic features reinforced the impression that the texts were not fully integrated into the stylistic norms of mathematical writing.
Some local ideas had superficial plausibility, but none were developed in a mathematically meaningful way. The outputs resembled a stream of initial intuitions that a human mathematician might have before formalizing anything — but without the subsequent refinement or insight.
\end{enumerate}
This case study highlights a central limitation: the system can imitate the form of mathematical writing, but not yet its function. Mathematics is not merely a collection of symbols or formulas, but a social and rhetorical practice governed by implicit conventions. Validity in mathematics is not conferred by structure alone, but by participating convincingly in a shared discourse — one that includes expectations of clarity, responsibility, and logical development.

While the experiment did not uncover any convincing or usable results, it revealed how close the system is to replicating outward patterns of mathematical behavior. The fact that it can generate seemingly plausible definitions or mimic standard stylistic elements is nontrivial. Still, the outputs lacked the clarity, progression, and internal necessity that characterize mathematics, at least as conventionally recognized within the mathematical community.

\subsection{Validation and Evaluation}

Formally, validation is the act of providing objective evidence that a system, model, or method is fit for its intended purpose. In the case of AI-generated papers, validation should be understood as the process of checking whether the claims made in the paper are true or not. In other words, validation tasks include verifying that the codes written are correct and that the results obtained contain enough evidence to support the claims in the paper. For standard papers, this process is called peer review.

We note that LLM hallucination is a well-known problem that can cause agents write incorrect code, reach conclusions or make claims that do not have enough evidence. In other cases, they can simply make up things. This problem can occur at any stage, from the generation of ideas and coding to paper writing. While \texttt{Denario} contains multiple agents whose role is to check the work of others, ultimately, a human expert should be responsible for checking and verifying all steps involved in the process. 

We have performed an exhaustive validation of some of the papers generated by \texttt{Denario}. We now discuss two particular examples: 
\begin{itemize}
\item For the cyclic peptide generation mentioned above, we found that even if all the agents involved in coding failed, the paper-writing agent wrote an entire paper claiming things without any evidence at all, likely being dominated by its prior knowledge and not the evidence from its calculations. 

\item To evaluate the scientific integrity of one of the generated papers in astrophysics, in person, our team read the paper in detail and wrote code to reproduce the figures and claims in the paper. We note that we did not provide access to the codes written by \texttt{Denario} for that paper. The domain expert was able to reproduce very similar metrics for the correlation and validate that an independent human reviewer can arrive at the same conclusion and scientific discovery.
\end{itemize}

An obvious conclusion is that a human expert should always check the codes and claims made by \texttt{Denario} written papers. However, we also note that this should also be done for standard human-created papers. We note that there have been multiple cases of standard papers that have passed peer-review validation and have lately been found to be wrong due to authors modifying or inventing data in order to provide evidence for a preconceived idea. 

Another important aspect in the revision of scientific papers is evaluation. In this context, we understand evaluation as the process of determining the quality of the paper, its value in comparison with previous works, and its contribution to the field. We note that evaluating the quality of scientific papers is a highly non-trivial and likely subjective task. In his current form, the evaluation of AI-generated papers by \texttt{Denario} is done by human domain experts\footnote{Although we note that \texttt{Denario} contains the reviewer module that is designed to provide a referee-report-like as a proxy for performing evaluation.} 

\begin{figure*}[ht!]
\centering
    \includegraphics[width=0.9\textwidth]{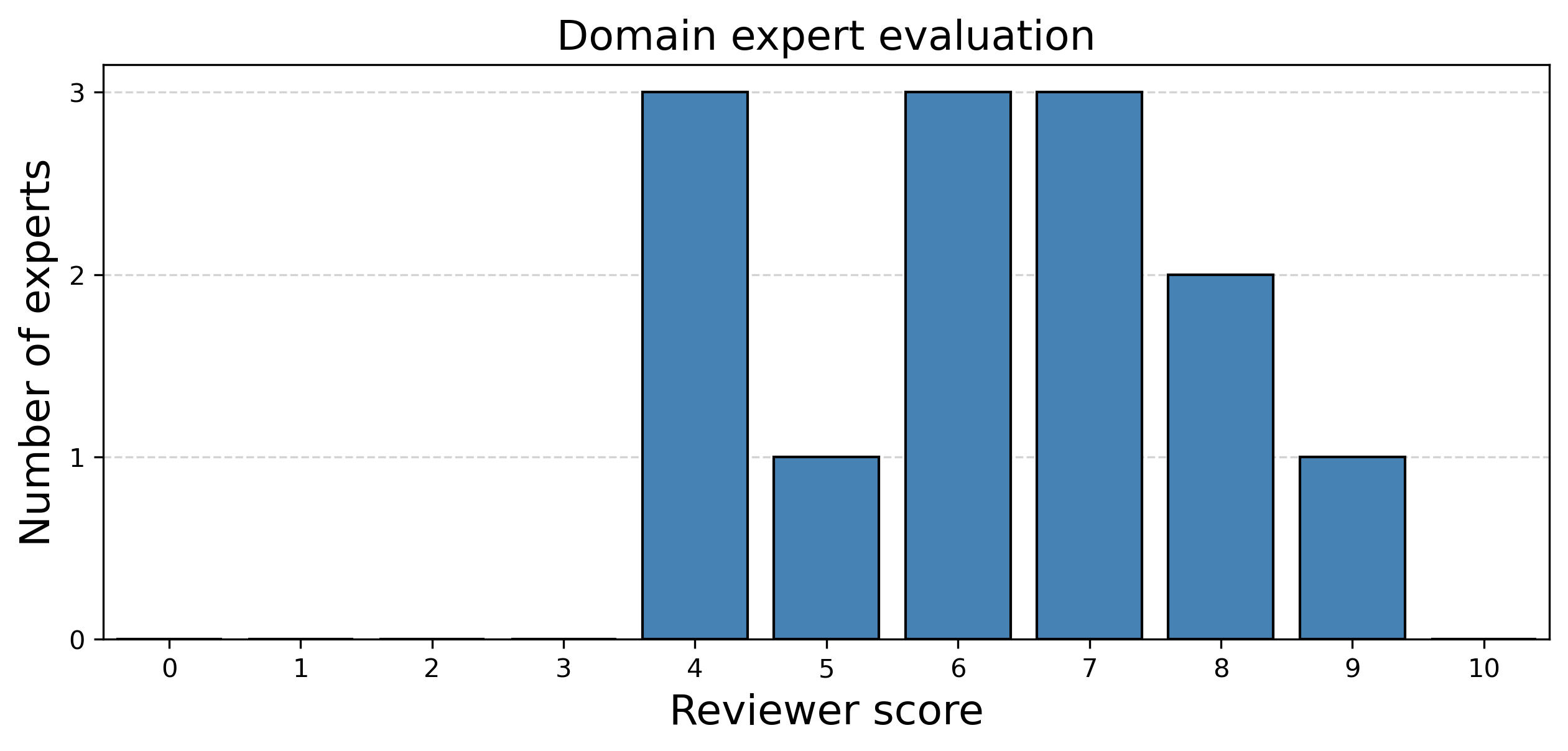}
    \caption{\label{fig:evaluation} We have assigned each of the papers in the Appendix to a domain expert for evaluation. Each expert, knowing that these are AI-generated papers, was tasked with providing a numerical score to quantify the quality of the paper, from 0 (really bad paper) to 10 (really good paper). This graph shows the distribution of the scores. As can be seen, while some papers are ranked below average, most papers are above average, and some of them were highly ranked.}
\end{figure*}

For the papers discussed in this paper (the ones located in the Appendix) we have asked the domain experts in our team to evaluate them and to provide a numerical score to them. The range was from 0 (for really bad papers) to 10 (really good papers). We show the distribution of the numerical scores provided by the experts in Fig. \ref{fig:evaluation}. We note that a numerical score of 5 represents a "normal" paper. As can be seen, 3 papers were ranked below the average, while the majority of them were above. In some cases, the human experts provided scores as large as 8 and 9. From this experiment, we can conclude that \texttt{Denario} has the capacity of producing "reasonable" and "interesting" papers, given all the caveats discussed in Sec. \ref{sec:examples}. We note that this score is highly subjective and does not pretend to be a rigorous criterion, but to provide some insight into the generative capabilities of Denario.

\subsection{Advantages and limitations}

According to the analysis of the example papers discussed in Sec. \ref{sec:examples}, we believe agentic systems like \texttt{Denario} will bring some important positive aspects to science in general:
\begin{itemize}

\item The ideas generated by \texttt{Denario} can enhance the researcher's capabilities by unveiling concepts (from mathematics, machine learning, or other disciplines) unknown to the researcher. We believe this may help researcher be more creative and discover techniques that may be relevant to their particular problem.

\item System like \texttt{Denario} excel at performing shallow research very quickly. In other words, while current LLMs may not have the capabilities to abstract the solution to a problem to something fundamental (e.g., a single mathematical equation governing the physics of many different systems), they are very good at performing relatively simple analyses and reaching very useful conclusions. This capability may enable researchers to identify promising research avenues and focus on those faster than with traditional human approaches. 

\item All text, plots, and code are output by \texttt{Denario} and therefore provide a high level of transparency and enable reproducibility. This is an important step for many disciplines, where calculations are complex and results cannot easily be reproduced.

\item The potential connection with other fields at the level of ideas of methods may enable more collaboration and interdisciplinary work.

\item Systems like \texttt{Denario} are excellent at exploring risky ideas. For instance, if a study is highly uncertain, and the outcome may not be very rewarding (e.g., exploring an idea that even if it works, may not lead to a publication), many researchers will feel discouraged from pursuing it. However, that study may be very useful for other analyses, perhaps completely unrelated. We believe researchers may consider exploring such ideas if the effort required is significantly reduced, as tools like \texttt{Denario} can do.

\item We believe that systems like \texttt{Denario} can help researchers in many different tasks, from idea generation to method implementation. Ultimately, if we lead AI to perform some of these tasks, under our supervision, scientists will have more time to think deeply and perform higher-level research. This could make science deeper, broader, and more robust.
 
\end{itemize}

It is also important to discuss the current limitations associated to \texttt{Denario}
\begin{itemize}

\item Citations. The citations in both papers typically refer to relevant papers for the considered sentence, but perhaps not the most relevant ones. An expert reader can easily spot an \texttt{Denario} AI-generated paper from a human one just from the citations. 

\item Depth. From the generated papers, we can see that the explanations of the results tend to be relatively shallow and not comparable to what an experienced researcher would write. As we show in the discussion of many papers in Sec. \ref{sec:examples}, current system behaves more like a good undergraduate or early graduate student rather than a full professor in terms of big picture, connecting results... etc.

\item LaTeX. Sometimes, figures are generated with non-LaTeX y/x labels or legends. For instance, some figures have omega\_m in the y-axis instead of $\Omega_{\rm m}$. Another problem is that the paper writing agent, more rarely, makes small mistakes in LaTeX, e.g., dealing with underscores in equations, text, and references. These errors usually prevent the paper from being compiled properly, although they are easy to fix.

\item Coding. In its current form, \texttt{Denario} sometimes may struggle with writing and executing code. This may happen for multiple reasons, such as changes in the API for recent packages and failures when running code in parallel. This behavior may cause the system to crash during the analysis module.

\end{itemize}

We believe all the above points can be addressed, or at least significantly improved, by improving the agents in charge of some tasks (e.g. citations), by advances in LLMs (e.g. reasoning), or by including more agents to fix the small mistakes made by the other agents (LaTeX).

\subsection{Future directions}

There are numerous ways to improve \texttt{Denario}, which we plan to implement in the near future:
\begin{itemize}
\item While in its current version \texttt{Denario} made some asynchronous LLM calls, it does not fully exploit its potential. Reformatting some agents to make more asynchronous calls can significantly speed up the time spent on the different tasks and lower the cost.  
\item In its current form, \texttt{Denario} may fail or significantly deviate from its goals at several stages of the process. Adding new agents that can provide feedback on the outcome of some agents will improve the stability of the framework and make it more robust.

\item Parallelization and optimization. Our current workflows are sequential: we deal with tasks step-by-step, doing one thing at a time. Efficiency will be greatly improved using parallelization strategies, e.g., \cite{niu2025flowmodularizedagenticworkflow} or \texttt{massgen}\footnote{\url{https://docs.massgen.ai/en/latest/}}.

\item Adaptive \textit{Planning \& Control}. Currently, once a plan is fixed, it is executed until the end. This is not how research is done. Next steps are always conditioned, or even determined, by past ones. A more refined  \textit{Planning \& Control} strategy will have adaptive planning so that after each task, the new steps left to be done can be re-assessed and potentially changed, removed, or some new steps can be added. 

\item The paper writing module can be significantly improved by adding more context information about the code written and the figures created. Furthermore, specialized agents to check claims and put them in touch with literature agents can significantly help write things with a broader perspective.

\item Currently, Denario only supports paid LLM providers through their API, but there are plenty of open-source models that can be installed and run locally. We plan to provide support to run our agents with locally installed LLMs, which will give the user more freedom to choose among the available LLMs, use custom models, and get more control over their resources.

\item We keep working on easing access to Denario. Besides improving the Python API and the GUI, we plan to include a Command Line Interface (CLI) to run the different modules of the multi-agent system without the need to write code explicitly.

\item Develop evaluation frameworks that will allow us to improve the quality of the research assistance tools in a systematic way. This will rely on extended benchmark datasets, following the efforts of TPBench \cite{Chung:2025nsd} or ReplicationBench \cite{2025arXiv251024591Y}.

\end{itemize}

\section{Ethical implications and scientific challenges} 
\label{sec:ethics}

Systems like \texttt{Denario} are designed to facilitate the work of scientists. However, the powerful technology behind them raises many ethical considerations. In this section, we describe some of them and also discuss how AI agents fit in the way we do science from a historical perspective.

\subsection{Ethics}

We now describe a few points where the usage of systems like \texttt{Denario} will lead to ethical considerations.

\begin{itemize}

\item \textbf{Quality of research}: Generative AI tools such as Denario allow users to easily and quickly generate large numbers of potentially low-quality papers. Some of the incentive structures of academic research may lead to individuals using generative AI tools to artificially increase their research output and/or inflate citation numbers for their own articles. 

\item \textbf{Implications for scientific training and education}: Tools such as \texttt{Denario} have the potential to make it easy to write an entire paper with just a few lines of description about a project idea, its methods, results, and the plots made. This poses questions and challenges for scientific training and education. In what ways are scientific writing and creativity in data analysis valuable as skills inherent to the researcher? To what degree or in what fashion should AI tools play a role in helping researchers carry out this type of work to begin with? How should we teach students the skills they will need to succeed as producers and interpreters of research? 

\item \textbf{Assessment of quality and legitimacy}: If these tools start being used widely, there may be a risk of trusting results from agents without any validation. For instance, an agent may perform some calculation and decide that it is not interesting. Without validation, that result may block a potentially interesting research direction. Indeed, as AI agents become both increasingly common producers and consumers of research, we are faced with important questions about the nature of quality and legitimacy in research. How is quality and correctness defined, and by whom? If AI agents become the standard by which to assess the quality of a paper or whether it is ``correct" or ``wrong", how does one assess which AI agent to trust? We can therefore fall into a situation of crisis with respect to responsible scientific working guidelines. If anyone can use AI to generate paper, and if no one can control this process and oversee standards of originality, who can be held responsible?

\item \textbf{Homogenization:} A pattern seen across some of the application cases presented is \texttt{Denario}'s difficulty proposing new research questions or identifying serendipitous, potentially novel findings. This is representative of a greater concern that AI agent-driven research might be biased towards producing research that confirms and aligns with the existing scientific literature of its training dataset, potentially leading to homogenized research and stifling innovation. Further experimentation with model architectures, such as the multi-agent frameworks implemented here, could help address this issue.

\item \textbf{Access:} AI tools such as \texttt{Denario} require significant investment in computational infrastructure and energy for their training and operation. These constraints shape which actors have access to their use and operation, affecting the fields of science being chosen for AI agent-driven research, the types of problems being tackled, and the methods used to do so. We must take deliberate steps to democratize both access to the benefits of AI agent-driven research and participation in its creation and training, in order to make sure science addresses the full breadth of interests and problems across societies.

\item \textbf{Authorship:} Generative AI has already sparked debates over the nature of scientific authorship and \texttt{Denario} is no different. A key requirement for understanding the debate is a careful definition of the roles of authorship in the scientific literature. One clear role of authorship is its use to allocate professional credit and recognition for doing work, which is important for career development. This aspect of authorship raises important questions that we must build answers to as a scientific community: if a human author uses AI agents to aid their research, in what ways is that different from previous research tools, and should that change the way we attribute professional credit to the human author? Should AI systems be assigned that kind of professional credit, and if so, for what purpose? How should the corpus of work used to train the AI agent be recognized, and what kind of professional credit should be used? Moreover, a second role of authorship implies responsibility for the quality and some implications of the results presented, an aspect that also raises critical questions for both the scientific community and wider society. Who should bear the ultimate responsibility for the work presented in a paper? What kind of guarantees and accountability should be upheld by the individuals and organizations training the model? A valuable but certainly insufficient step to addressing these issues would be to clearly state when a research article has been generated by agentic systems, which would have the added benefit of flagging the article as such when considering the training datasets of future AI agents, improving their performance \citep{schmidgall2025agentrxiv}.

\item \textbf{Misalignment of incentives in scientific publishing:} Given the speed and volume with which research articles can be generated using agentic tools like \texttt{Denario} and the career incentives that are associated with inflated publishing statistics in our current academic system, the proliferation of AI agent-driven research may lead to a deluge of low-quality papers overwhelming scientific publishing and readership capabilities. This misalignment of career incentives with the purposes of productively advancing scientific knowledge can also lead to situations in which users maliciously generate research articles that purposefully cite their own previous research. These challenges pose questions not only about the incentive structure in academic publishing and career building, but also about what the role of a ``research article'' is in the age of generative AI and what form or medium can most effectively communicate and evaluate science in this new regime.

\item \textbf{Environmental concerns:} Training and running LLMs come at a significant cost in the consumption of energy, water, coolant, and rare earth metals, among others \cite{dauner2025energy, 9869565}. As with other massive applications of machine learning, the scale of this resource consumption is far from sustainable at the time of writing. Researchers must take active roles in advocating for and developing solutions to address this major issue.

\item \textbf{Malicious use:} AI agents could be used to quickly flood the scientific literature with claims driven by a particular political agenda or specific commercial or economic interests. The volume of papers created could then influence the corpus used to train non-malicious AI agents, reinforcing and effectively ``laundering" facts. Developing ways to detect this at the technical level and discuss it at the societal level is critical.

\end{itemize}

Overall, tools like \texttt{Denario} can enhance researchers' capabilities by helping them in tasks that require a significant amount of time, such as writing papers, debugging code, or exploring ideas. We believe the best usage of \texttt{Denario} would be as an AI-assistant for different tasks. For instance, it can be used to improve an idea (by exploiting methods not known to the researcher), or help with the writing of the paper once the results are in place. \texttt{Denario} is able to perform complex tasks orders of magnitude faster than humans. We think the quality of the research papers will crucially depend on how the system is used:

\begin{itemize}

\item \textit{Responsible usage}. We believe that under responsible usage, the quality of the research papers will improve. In order to explain this apparently contradictory logic (fast generation vs quality), we believe that researchers are usually time-limited; e.g., an undergraduate needs to graduate or a graduate student needs to apply for postdoc/faculty jobs. Given a fixed amount of time, the fact that a very small fraction of that time goes into generating a draft with all associated codes and plots, will enable the researcher to dedicate more time to thinking deeper, improving the quality of the paper, improving the methods, than if the user needs a very large fraction of the time to just produce a first draft.

\item \textit{Irresponsible usage}. If the usage takes place in an irresponsible manner, i.e. just generating papers without or little validation, then given the current capabilities of systems like \texttt{Denario}, the quality of the papers will degrade.

\end{itemize}
While we expect both usage modes, we think most researchers will make use of these tools responsibly. Finally, we would like to stress that systems like \texttt{Denario} will further emphasize the need to give more weight to quality rather than quantity when evaluating scientists' careers. However, given the fact that quality is very hard to quantify and likely subjective, the current system may need to reshape itself for the benefit of science and society.

\subsection{Epistemic changes in a science driven by AI agents}

The integration of LLMs and AI agents into scientific practice marks a profound inflection point in the philosophy of science. These systems are not mere computational tools; they increasingly participate in the core activities of science—formulating hypotheses, synthesizing literature, generating theoretical connections, falsification procedures, theory choice, epistemic virtue-discussions, and even simulating reasoning. This shift forces us to re-examine long-standing philosophical frameworks concerning the nature of scientific rationality, understanding, and progress. In particular, the proliferation of AI agents in scientific research poses questions regarding the nature of scientific knowledge, its legitimacy, its quality, and its ultimate purpose. 

\subsubsection{The nature of scientific knowledge in AI-driven science}

To examine this, we can first turn to the notion of “research programmes” as defined by Imre Lakatos (1978) \cite{lakatos1978methodology}: long-term, theoretical frameworks with a stable “hard core” and evolving “protective belts” of auxiliary hypotheses. Lakatos argued that scientific rationality lies in comparing the relative progress of these programmes: a progressive programme predicts novel facts and accommodates anomalies, while a degenerative one merely accommodates data retrospectively. Within this framework, LLMs raise fundamental questions. Are they simply auxiliary tools bolstering existing programmes—helping refine theories, conduct literature reviews, and test hypotheses more efficiently? Or do they represent the birth of a new kind of research programme altogether—one not driven by causal explanation, but by linguistic modeling, probabilistic reasoning, and massive-scale pattern recognition?

If the latter is true, then LLMs bring with them a new type of “hard core”: the assumption that knowledge is fundamentally representable and usable in statistical-linguistic form. This new type of hardcore could be ultimately defined as the model embeddings of a training dataset and interpreted in human terms as the weighted aggregate of a body of scientific literature. Weighted how? Nobody can fully understand, because understanding mechanistically is no longer the goal. This contrasts sharply with traditional programmes based on mechanistic or formal-logical representations. Moreover, LLMs produce novel predictions, connections, and reformulations—yet not always in a way that is interpretable or falsifiable in Lakatosian terms. As such, they may challenge the very criteria by which progress in science is traditionally measured. Lakatos maintained a vision of rational science driven by theoretically motivated explanations; LLMs, in contrast, are increasingly effective despite their lack of explicit theory. Their success poses a challenge: can a research programme be progressive even if it lacks a human-understandable theoretical core?

W.V.O. Quine’s epistemology—particularly as articulated in "Two Dogmas of Empiricism" (1951) \cite{van1976two} and his broader naturalized epistemology—offers a subtle but profound framework for understanding the epistemic disruptions caused by LLMs. Quine rejected the analytic-synthetic distinction and argued for a holistic view of knowledge, where statements face the tribunal of experience not in isolation, but only as part of an interconnected web of belief. In this view, revision is always possible, even for logical or mathematical statements, and empirical data can force changes anywhere in the system.

From this perspective, LLMs do not simply introduce new tools into the scientist’s arsenal; they reconfigure the web itself. LLMs are not theory-driven or tethered to empirical input in the traditional sense. They model language, not the world—but through language, they influence belief revision, hypothesis generation, and theoretical framing. As such, they intervene in science at the level of conceptual structure, shifting what is considered coherent, plausible, or meaningful within the web of scientific belief.

\subsubsection{The nature of scientific legitimacy, quality, and purpose in AI-driven science}

The notion that science is a historically contingent and socially negotiated enterprise, insisted upon by writers such as Paul Feyerabend (1975) \cite{feyerabend2020against}, finds new relevance in the age of AI. As LLMs are integrated into research workflows, they begin to shape what is seen as legitimate knowledge, not through philosophical argument but through infrastructural embedding—by determining what is written, cited, or ignored. This has deep implications for the sociology of knowledge: the epistemic authority of LLMs may evolve not because they “know” in any human sense, but because they shape the conditions under which knowledge is produced and circulated.

Additionally, Quine’s naturalized epistemology—his call to study knowledge using the methods of science itself—opens the door to seeing LLMs as epistemic agents within a naturalized framework. They are not mere passive instruments but active components in the scientific ecosystem, akin to laboratories, statistical methods, or even communities of inquirers. Yet they differ in one crucial way: they operate in a non-transparent manner, lacking not only introspection but causal understanding. This opacity generates an ontological challenge: What is the status of knowledge produced by a system that lacks grounding in experience or semantics, and whose justification lies in performance rather than comprehension? How is the quality of this mode of scientific knowledge judged, and should it be judged differently at all?

From a Quinean standpoint, LLMs are both natural extensions of scientific inquiry and radical disruptors of its structure. They unsettle the boundary between tool and theorist, instrument and agent. They force us to reconsider what counts as observation, evidence, and explanation, and they deepen the holistic fluidity of the scientific web—potentially accelerating progress, but also undermining the interpretive clarity on which much of science has historically depended. Ultimately, AI-driven research opens (or reopens) a question on the fundamental purpose and value of scientific knowledge and enterprise, highlighting a tension between the goal of a mechanistic, human understanding of the world and the goal of achieving statistical practical utility first and foremost.

An interesting situation in which this tension surrounding the ultimate purpose of agentic AI research manifests is the Turing Trap \cite{brynjolfsson2023turing}. The term “Turing Trap” refers to the tendency to develop AI systems that mimic or replace human cognition, instead of enhancing or complementing it. Named in reference to Alan Turing’s famous test for machine intelligence, the trap emerges when passing as human becomes the benchmark for success, thus narrowing the scope of what AI can and should do. In science, this translates into a shift from tools that assist human reasoning toward agents that simulate it. LLMs such as GPT-5 are increasingly being used not merely for clerical tasks, but for co-authoring papers, generating reviews, or even proposing research directions. This then underscores the question: if AI agents have the potential to exceed human capabilities in at least certain regards, should we reevaluate the basic tasks and units of how science is practiced, communicated, and evaluated in non-human terms?

\subsection{The impact of AI agent-driven science on the structure of scientific work}

Besides the questions raised regarding epistemic changes in the nature of science and scientific knowledge, agentic AI-driven research also appears poised to reconfigure the structure of scientific practice. To understand the ways in which this may occur, it is helpful to use the framework developed by Thomas Kuhn (1975) \cite{kuhn1970structure} according to which science develops through periods of “normal science” in which knowledge is incorporated into existing theoretical frameworks, alternated by “paradigm shifts” during which those frameworks are overhauled. 

\subsubsection{AI agent-driven science as an enabler of innovation}

From a Kuhnian perspective, LLMs could be seen as catalysts for a paradigm shift. Kuhn emphasized that science does not evolve through continuous accumulation of knowledge, but rather through disruptive revolutions in which one paradigm is replaced by another, reshaping the methods, standards, and even the very questions that science considers meaningful. LLMs introduce precisely this kind of rupture: they not only change how science is done, but what it means to do science. Their ability to generate plausible hypotheses, summarize vast literatures, and recombine knowledge across disciplinary boundaries at scales and speeds beyond the capabilities of a single human researcher challenges the prevailing paradigm in which singular human cognition, creativity, and interpretive judgment are central to scientific discovery.

Kuhn’s notion of “normal science”, routine puzzle-solving within an accepted paradigm, may become destabilized as LLMs increasingly participate in tasks once thought uniquely human. Moreover, their epistemic style is deeply different: LLMs operate not on causal mechanisms but on statistical and linguistic associations. Insofar as LLMs begin to influence theory choice, experimental design, or even what is considered an interesting or valid research question, they may be ushering in a new epistemic regime, one where prediction trumps explanation, pattern overrules theory, and where coherence across texts replaces ontological depth. Kuhn’s framework helps us see that the rise of LLMs may not be a simple technological enhancement. It could even bring us to a deep transformation of the values and structures that undergird scientific practice. This presents a challenge for the very practice of science.

Paul Feyerabend's radical critique of scientific rationality offers another perspective. In Against Method (1975) \cite{feyerabend2020against}, Feyerabend argued that science does not—and should not—operate according to fixed methodological rules. Progress, he claimed, often occurs when conventions are violated and when heterodox or even “irrational” approaches are permitted to flourish. From this viewpoint, LLMs may be the ultimate Feyerabendian agents. They defy the norms of methodological consistency, rely on opaque black-box architectures, and generate insights not through logic or controlled experimentation but through stochastic generation and pattern inference across language corpora. Precisely because of this transgressive character, LLMs may serve as engines of epistemic pluralism and innovation. They break down disciplinary barriers, resurface forgotten ideas, and draw novel analogies that human researchers may overlook. Feyerabend’s critique of scientific monism—the idea that there is one best method or one correct epistemic framework—is echoed in the way LLMs expose the contingencies and limits of human-centric scientific norms. In his terms, they have the potential to be tools that reassert the chaotic, cultural, and creative dimensions of science, reminding us that genuine progress often arises not from strict adherence to method but from bold and unpredictable departures from it.

\subsubsection{AI agent-driven science as a stifling, homogenizing force}

Despite this potential, an argument can also be made that AI agent-driven science risks minimizing innovation rather than enhancing it. In Kuhnian terms, AI agent-driven scientist could lead to a reduction in paradigm shifts and a slump into "perpetual normal science". When AI agents carry out entire cycles of knowledge production, we risk creating a system that perpetually produces research extending the status quo, the risk of homogeneity. Given that these agents are by design trained on broadly similar bodies of knowledge, one wonders to what degree they are intrinsically biased towards what Kuhn calls "normal science production", tending to generate output that fits well to knowledge available in the training set. In Feyerabendian terms, the nightmare of total scientific monism could be a possible outcome of AI-driven homogenization if AI agent research converges on the same training data sets, architectures, and desired outcomes. 

In addition to the biases inherent in their own architecture, AI agent-driven research can also reinforce biases in the structures of scientific institutions. The ease with which agentic AI can be used to produce scientific output that easily conforms to existing dogma—useful in some environments to inflate statistics valuable for career advancement—could result in less paradigm-shifting work. Furthermore, the potential for a rapid feedback loop in which future agentic AI is training on AI-generated scientific literature with the same bias towards "extending" existing theory (rather than disrupting it) could accelerate this trend.

However, this doesn't have to be the case: we can build in statistical and agent-level architectures that minimize this bias towards normal science and emphasize the role of human scientists within the loop of AI-assisted knowledge production. Achieving this implies an active effort from all parts of the scientific enterprise, from the computer science behind the design of AI agents to the career incentives for individuals and funding mechanisms used to encourage scientific production. It is crucial to democratize both access to the tools themselves as well as the skills and infrastructure for their development, and to be deliberate and conscientious when compiling training data.

\subsection{Conclusions from philosophy of science}

LLMs do not merely support current scientific practice; they transform it. Through the lens of Kuhn \cite{kuhn1970structure}, they potentially inaugurate a new paradigm of machine-augmented science—or plunge us into perpetual normal science. Through Lakatos \cite{lakatos1978methodology}, they challenge existing research programmes and perhaps herald a new one grounded in probabilistic modeling rather than explanatory theory, with the potential to bring about unprecedented pragmatic relevance—and perhaps a vision of science divorced from human understanding. And through Feyerabend \cite{feyerabend2020against}, they emerge as agents of epistemic disorder and creativity, unsettling dogma and enabling new modes of inquiry—or trap us in sterile, homogenized scientific monism. As the transformation of science and its practice by AI tools continues, it is worth remembering that an LLM is ultimately an ``universal interpolator" that retrieves information—directly or interpolated—from a body of knowledge, with all the advantages and flaws that this entails.

Ultimately, AI agent-driven science raises crucial but uncomfortable questions at the heart of science: are we outsourcing not just labor, but judgment and meaning? When the outputs of LLMs are treated as autonomous knowledge contributions—rather than the mediated results of probabilistic associations—they can displace the interpretive and evaluative labor that defines good science. Scientific rationality becomes performative: coherence, plausibility, and citation density stand in for theoretical depth, causal understanding, or experimental validation. To engage seriously with the implications of LLMs, we must therefore go beyond technical or instrumental perspectives and grapple with the philosophical foundations of science itself. What counts as knowledge, explanation, or progress may no longer be what it once was—and it is only by revisiting and reevaluating previous frameworks of science that we can begin to navigate this new epistemic terrain.

\section{Summary and Conclusions}
\label{sec:conclusions}

In this work, we have introduced \texttt{Denario}, an AI multi-agent system designed as a research assistant. \texttt{Denario} is built in a modular way, and can perform many different tasks, such as 
\begin{itemize}
\item Generate novel or refine existing research ideas
\item Check the literature to determine whether an idea is new and its potential interest
\item Develop detailed research plans
\item Write and execute code, make plots, and summarize results
\item Draft a scientific paper
\item Review a scientific paper
\end{itemize}
\texttt{Denario} can also perform end-to-end scientific research by running all the above modules sequentially. Different modules require different inputs and produce different outputs (see Table \ref{tab:denario_io}). When run in end-to-end mode, the system only requires an input text containing general instructions to the system (e.g., describing the data to be used). We emphasize that the goal of \texttt{Denario} is not to automate science, but rather to serve as a powerful tool to help researchers across disciplines accelerate scientific discovery.

In Section \ref{sec:methods}, we described the different components needed for creating multi-agent systems, from Large Language Models (LLMs) to agentic orchestration. In Section \ref{sec:architecture}, we presented in detail the different modules in \texttt{Denario} and the tasks they carry out. We have illustrated the capabilities of \texttt{Denario} in Section \ref{sec:examples} by evaluating papers generated by it in many different scientific disciplines, from astrophysics to biology. From the expert evaluation of these manuscripts, we conclude that the generated papers are at the level of what a good undergraduate or early graduate student could achieve after weeks or a few months of work. In comparison, it took \texttt{Denario} around 30 minutes per paper and a cost of around \$4. We have discussed the failure modes, the importance of validation and evaluation, and the advantages and limitations of \texttt{Denario} in Sec. \ref{sec:discussion}. Finally, in Sec. \ref{sec:ethics} we discussed the ethical considerations that agentic systems like \texttt{Denario} lead to and discussed how these technological advances could be perceived from a philosophy of science perspective. 

We note that we have not discussed important aspects related to AI agents such as benchmarking \cite{astabench} and risks \citep{Agents_risks}. However, these are active areas of research. 

We conclude by stating that \texttt{Denario} represents one more example that the latest advances in deep learning, in particular in large-language models and AI agents, may help scientists with their work \citep{AI_scientist, AI_cosmologist}. As we mentioned above, the papers that \texttt{Denario} can generate right now do not have the breadth and reach of those written by the more experienced researchers in the field. Whether future LLMs or more sophisticated systems with more specialized agents can achieve that remains an open question. 

We believe that tools like \texttt{Denario} can open up new ways of doing science. Traditionally, researchers focus deeply on a small set of questions, since tasks such as coding, reviewing the literature, and learning new concepts or tools require significant time and effort. By contrast, \texttt{Denario} excels at rapidly exploring a broad range of ideas, albeit at a more superficial level. These approaches are highly complementary: AI can scan widely and highlight promising directions, while human experts can select the most relevant ideas and pursue them in depth to achieve a deeper understanding.

\section*{Acknowledgements}

The work of FVN, AEB, RJ, and PTA was supported by the Simons Foundation. The work of BB was partially funded by an unrestricted gift from Google, the Cambridge Centre for Data-Driven Discovery Accelerate Programme and the Infosys-Cambridge AI Centre. Funding for the work of RJ and PTA was partially provided by
project PID2022-141125NB-I00, and grant CEX2024-001451-M funded by MICIU/AEI/10.13039/501100011033. PTA is supported by the project “Dark Energy and the Origin of the Universe” (PRE2022-102220), funded by MCIN/AEI/10.13039/501100011033. WHZ acknowledges the funding support from Japan Society for the Promotion of Science. PB acknowledges constant support from DIPC. \'{I}Z acknowledges support from the STFC (grant numbers ST/W000977/1 and ST/X006387/1.  MV is supported by Fondazione ICSC, Spoke 3 ``Astrophysics and Cosmos Observations''  Project ID CN00000013 ``Italian Research Center on High-Performance Computing, Big Data and Quantum Computing''  funded by MUR Next Generation EU (NGEU); the INAF Theory Grant "Cosmological Investigation of the Cosmic Web' and by the INFN INDARK grant.

We thank Viviana Acquaviva, Tal Adi, Yacine Ali-Haimoud, David Alonso, Daniel Angles-Alcazar, Juan Navarro-Arenas, Paramita Barai, Jane Beaufore, Stefano Borgani, Juan Camilo C\'ardenas, Matteo Cantiello, Isabella Carucci, Gabriella Contardo, Belen Costanza, William Coulton, Stefani Cristiani, Carolina Cuesta-Lazaro, Steve Cunnington, Julianne Dalcaton, Carmen Embil-Villagra, Serafina di Gioia, James Fergusson, Zucheng Gao, Shy Genel, Antoine George, William Handley, Mike Hobson, Teresa Huang, Kartheik Iyer, Yannis Kevrekidis,  Mica Komarnyckyj, Ely Kovetz, Sven Krippendorf, Andrew Laverick, Ilem Leisher, Antony Lewis, Sarah Libanore, Stephane Mallat, Sukanya Mallik, Dinko Milakovic, Andy Millis,  Andrei Mesinger, Azadeh Moradinezhad, Bruno Motta,  Moritz Münchmeyer, Yueying Ni,  Lorenzo Pacchiardi, Gonzalo Palma, Isabel Pederneiras, Vincent Prevosto, Amanda Prorok, William Ramirez,  Rogerio Rosenfeld, Walter Riquelme, Gary Shiu, Marta Spinelli, Riccardo Sturani, Mario Santos, Emanuel Schaan, Thomas Sumner, Irene Testini, Paul Torrey, Roberto Trotta, Amita Ummadisingu, Soledad Villar, Sebastian Wagner-Carena, Risa Wechsler, Benjamin Wandelt, Nils Wentzell, Jeannette Wing, Sandy Yuan  for useful discussions. We thank Licia Verde for insightful conversations regarding publication of AI generated papers and her participation in the launch webinar. We also thank Laura Edwards for her collaboration in the preparation of the video tutorials and the launch webinar. We thank Mathew E. Diamond for his invaluable guidance on the spike sorting and curation pipeline. We also acknowledge Davide Giana and Francesca Pulecchi for their expert surgical work, which enabled the acquisition of the Neuropixels data used in this study.

\bibliographystyle{unsrt}
\bibliography{bibliography}{}

\appendix
\section{Example papers}
\label{sec:papers}

\subsection{Mathematical physics}
\label{paper:matphys}
\includepdf[pages=-,landscape=false,fitpaper=true]{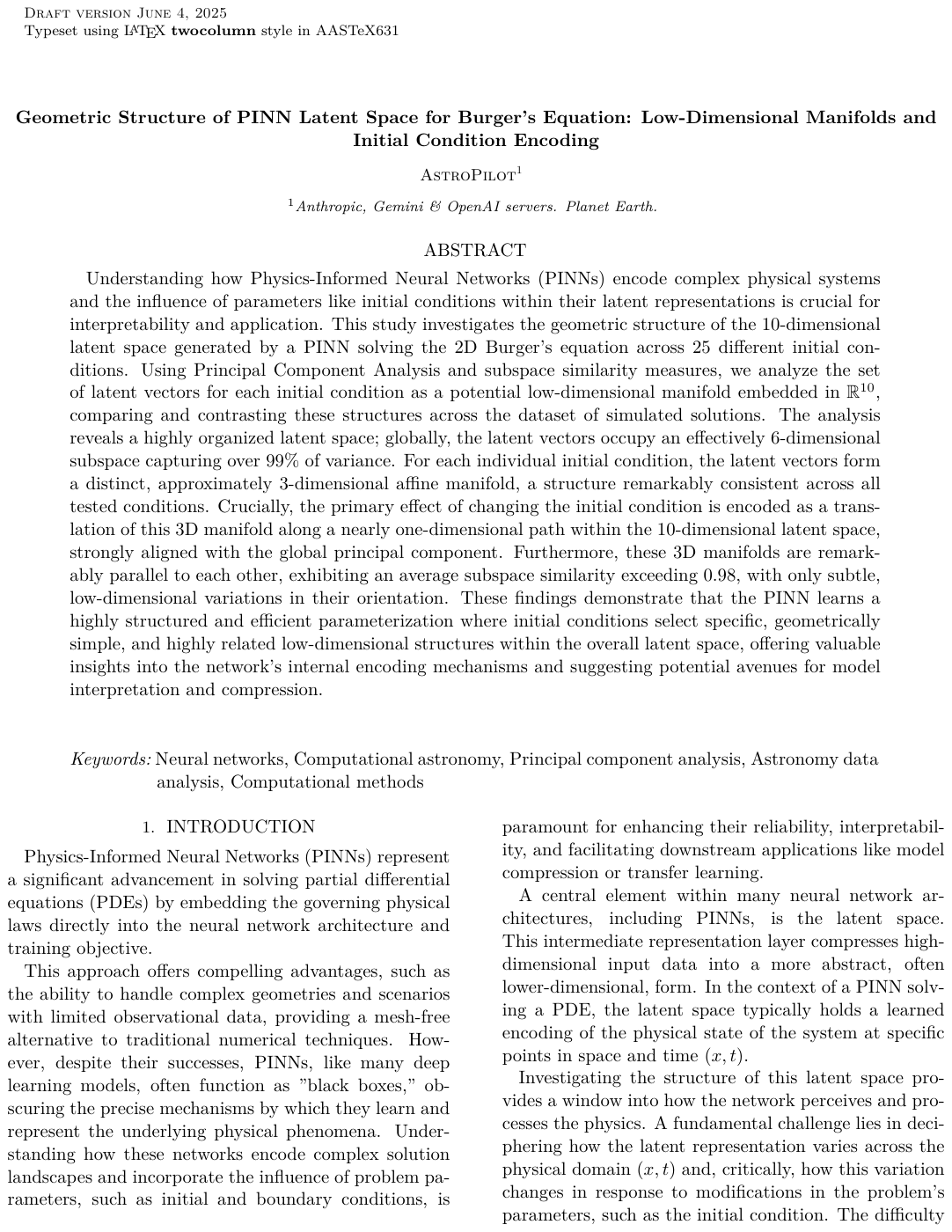}
\subsection{Planetary science}
\label{paper:planet}
\includepdf[pages=-,landscape=false,fitpaper=true]{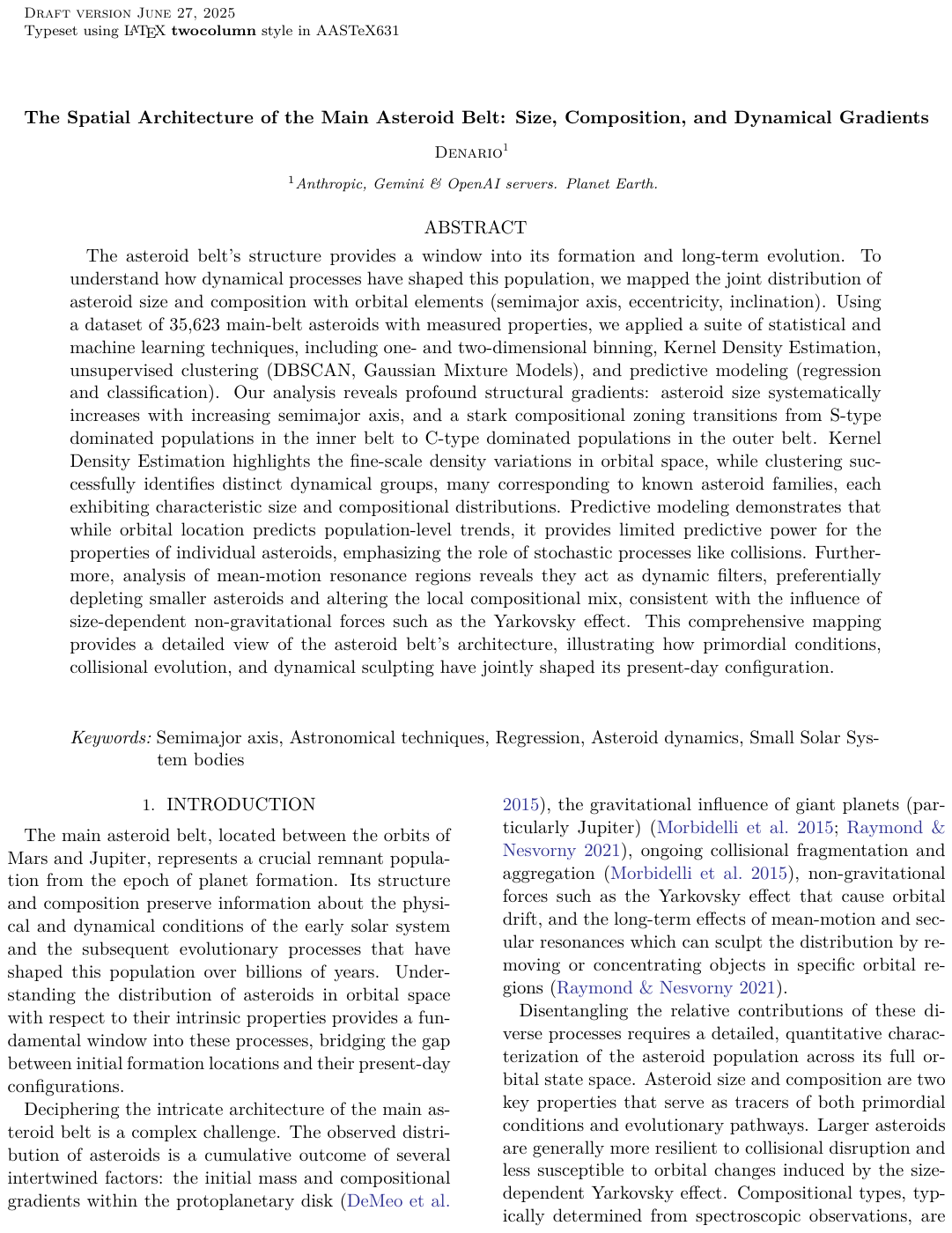}
\subsection{Medicine and Health Services Research}
\label{paper:med}
\includepdf[pages=-,landscape=false,fitpaper=true]{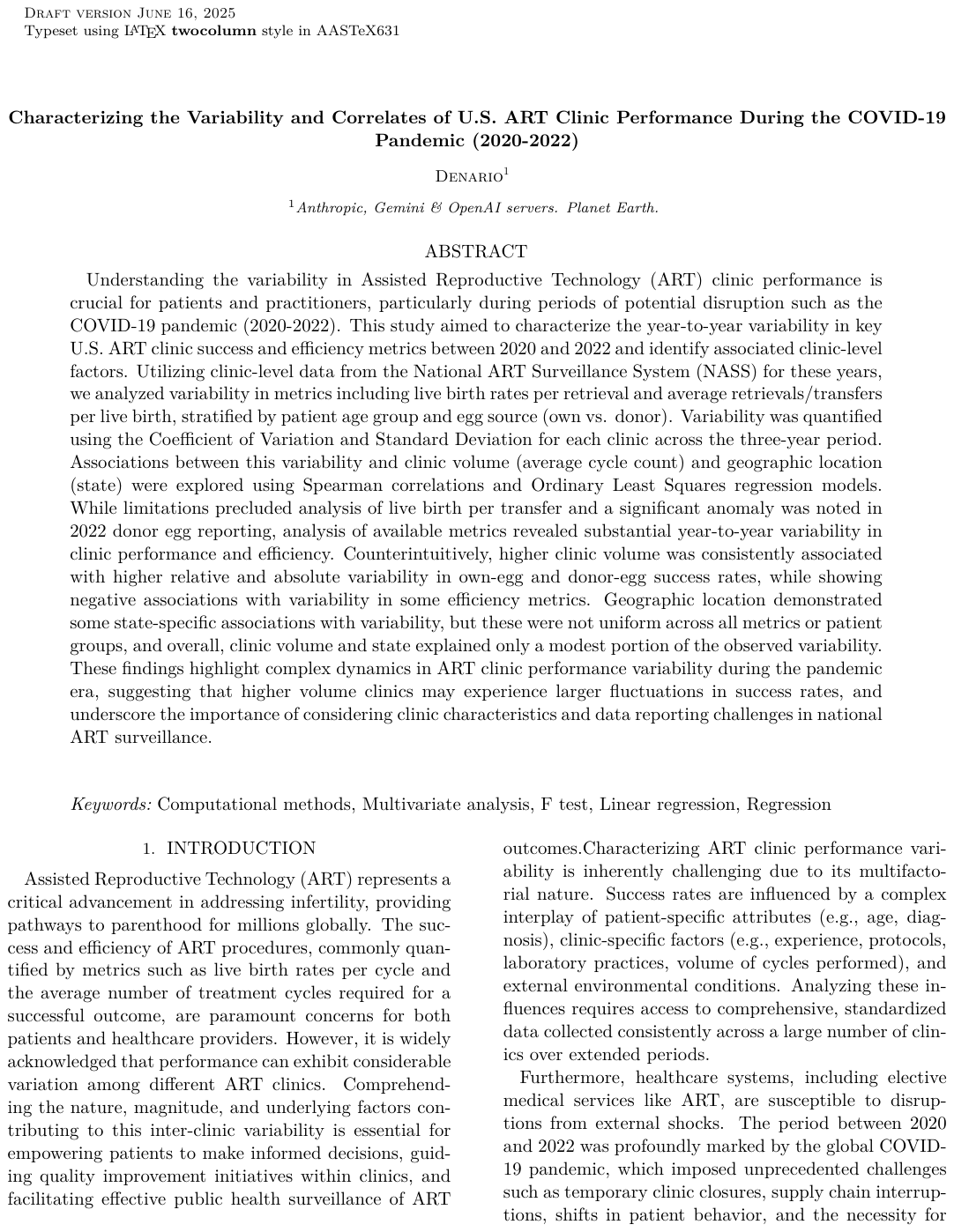}
\subsection{Chemistry}
\label{paper:chem}
\includepdf[pages=-,landscape=false,fitpaper=true]{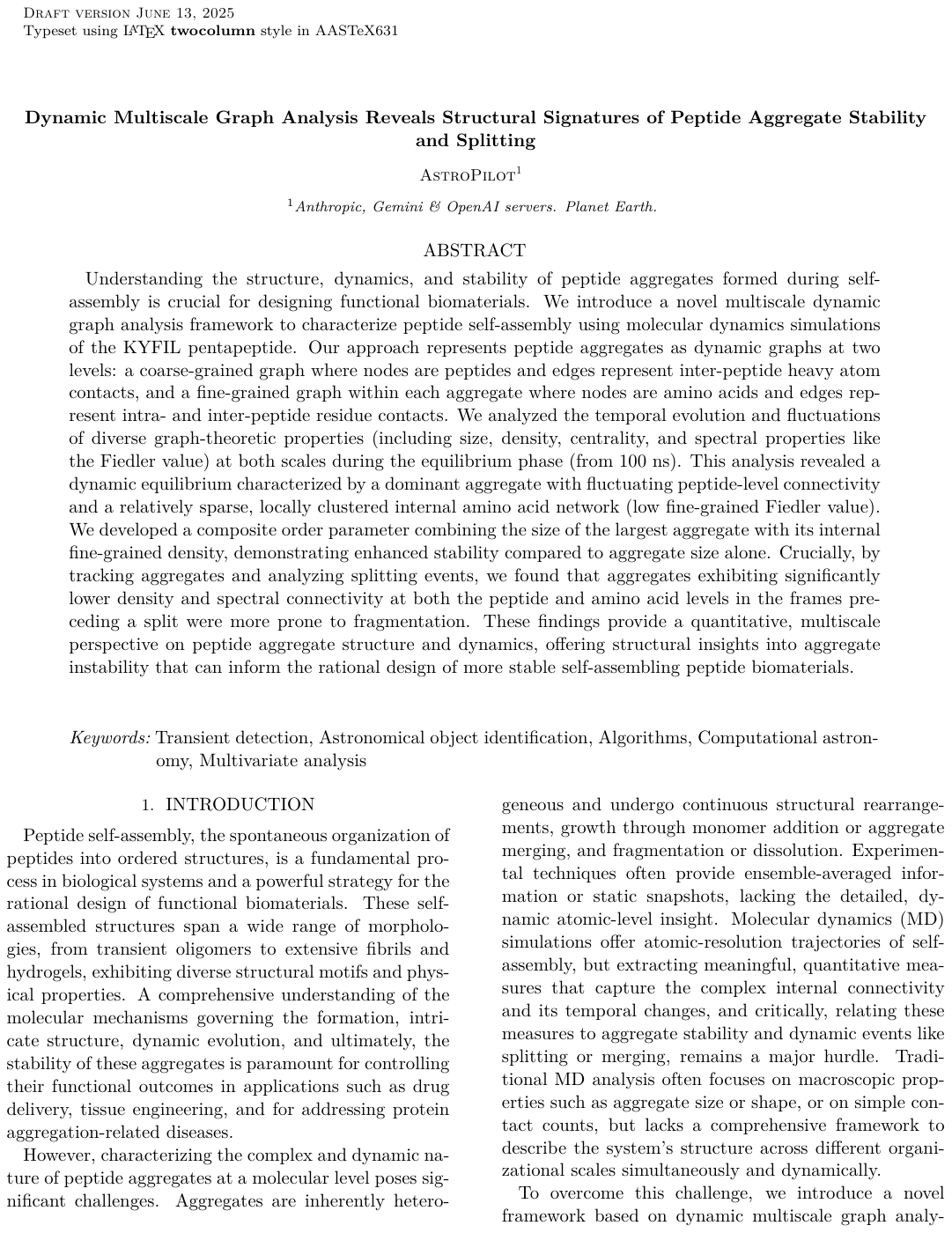}
\subsection{Astrophysics: gravitational waves}
\label{paper:gw}
\includepdf[pages=-,landscape=false,fitpaper=true]{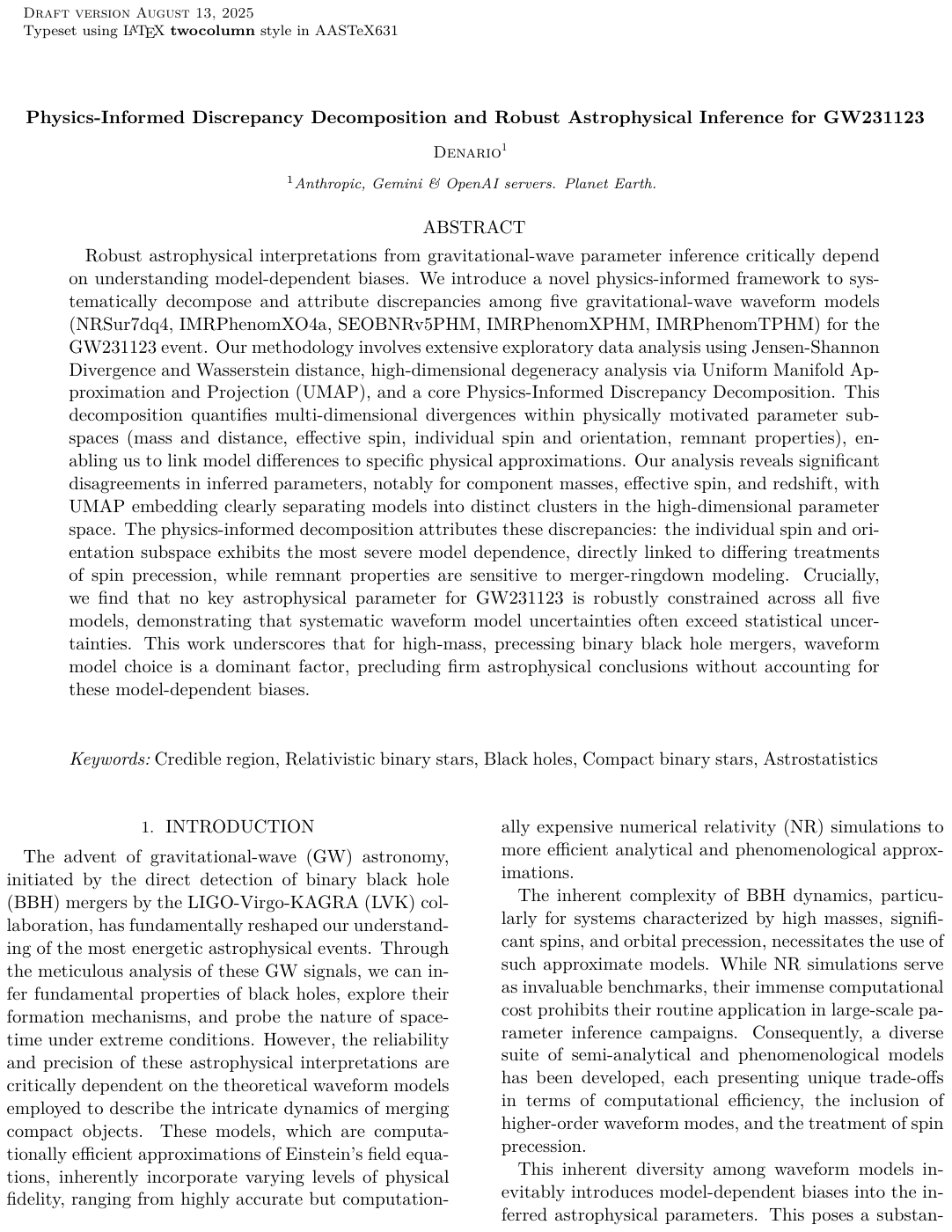}
\subsection{Astrophysics: stellar physics}
\label{paper:stel}
\includepdf[pages=-,landscape=false,fitpaper=true]{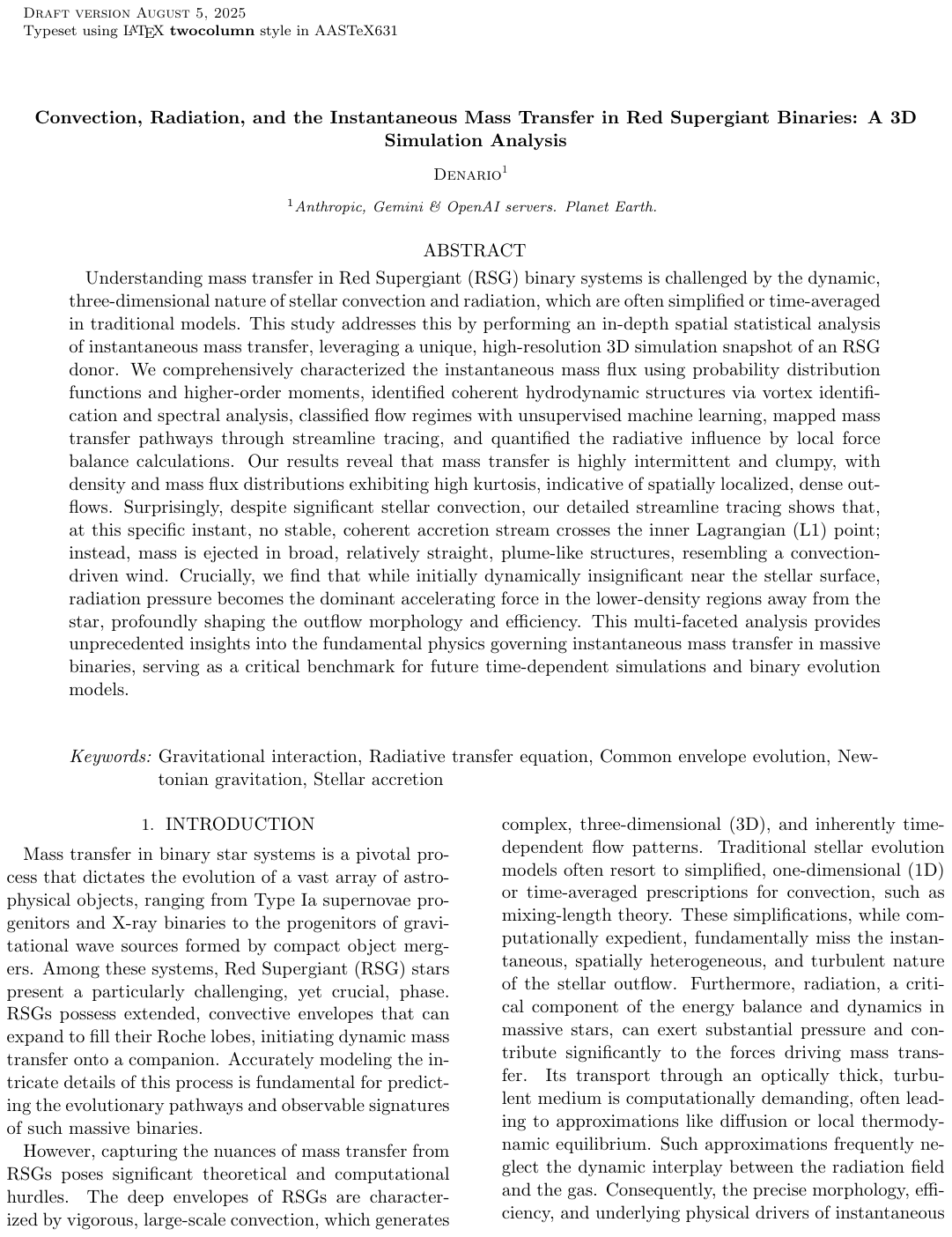}
\subsection{Biology}
\label{paper:biog}
\includepdf[pages=-,landscape=false,fitpaper=true]{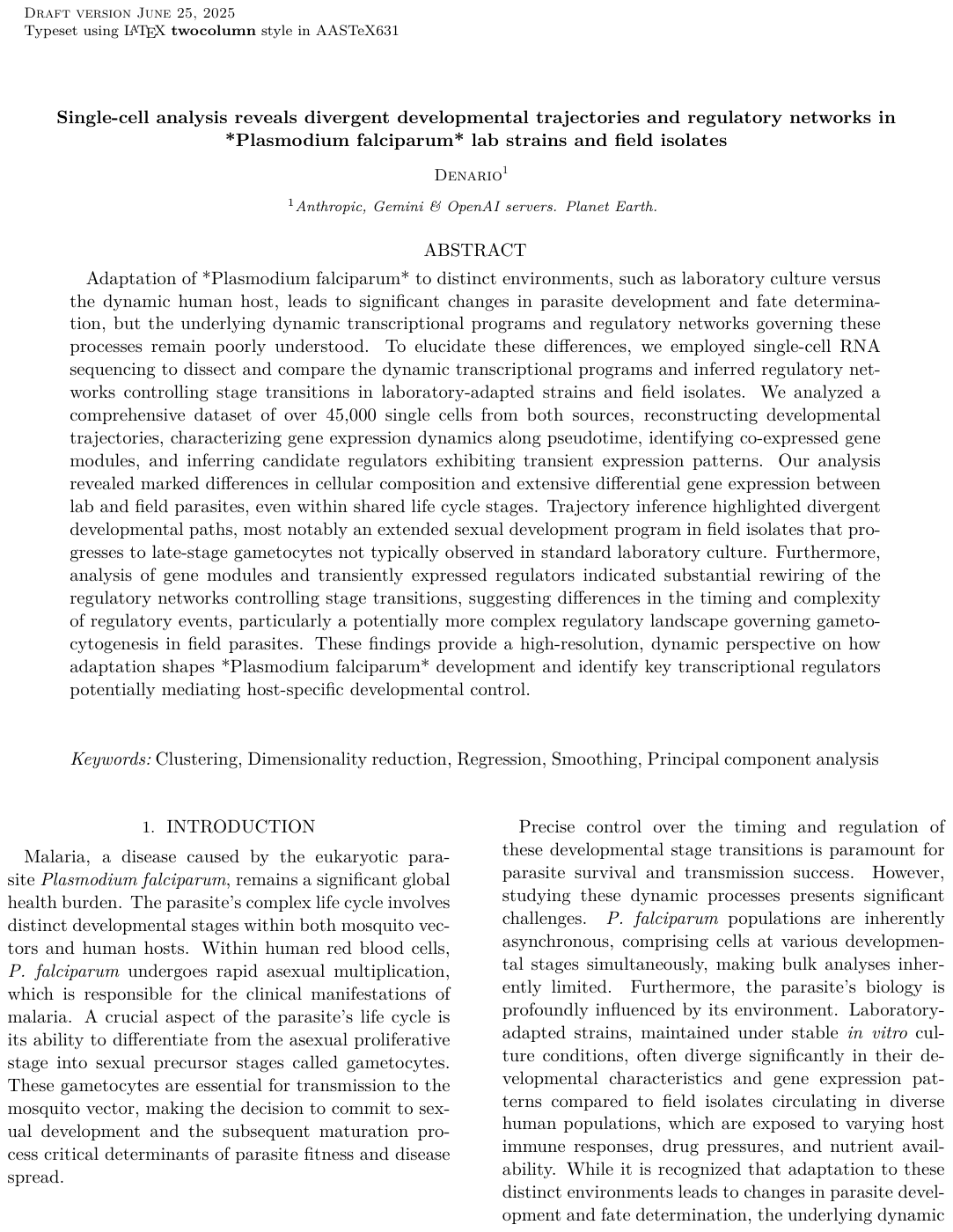}
\subsection{Digital health}
\label{paper:dh}
\includepdf[pages=-,landscape=false,fitpaper=true]{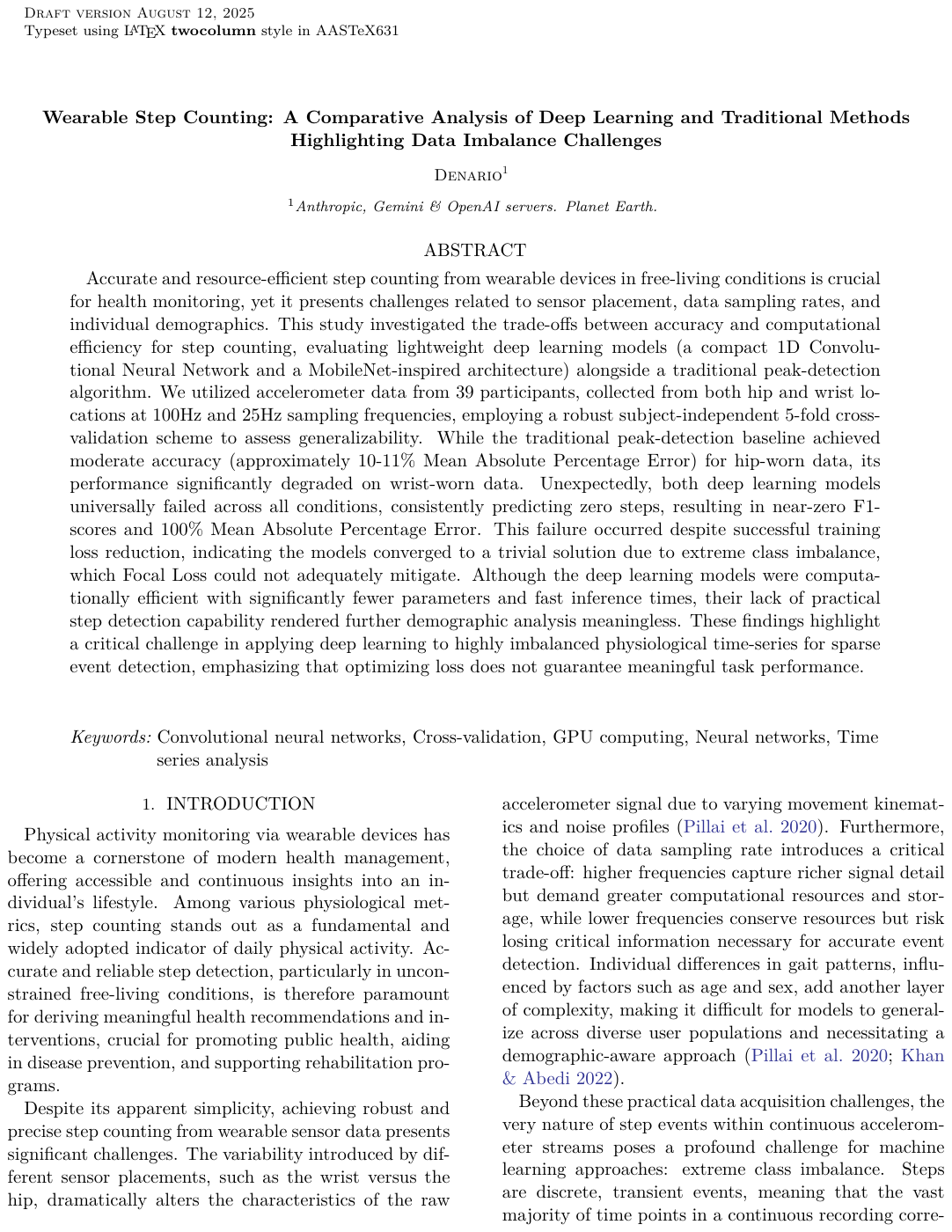}
\subsection{Biophysics}
\label{paper:biop}
\includepdf[pages=-,landscape=false,fitpaper=true]{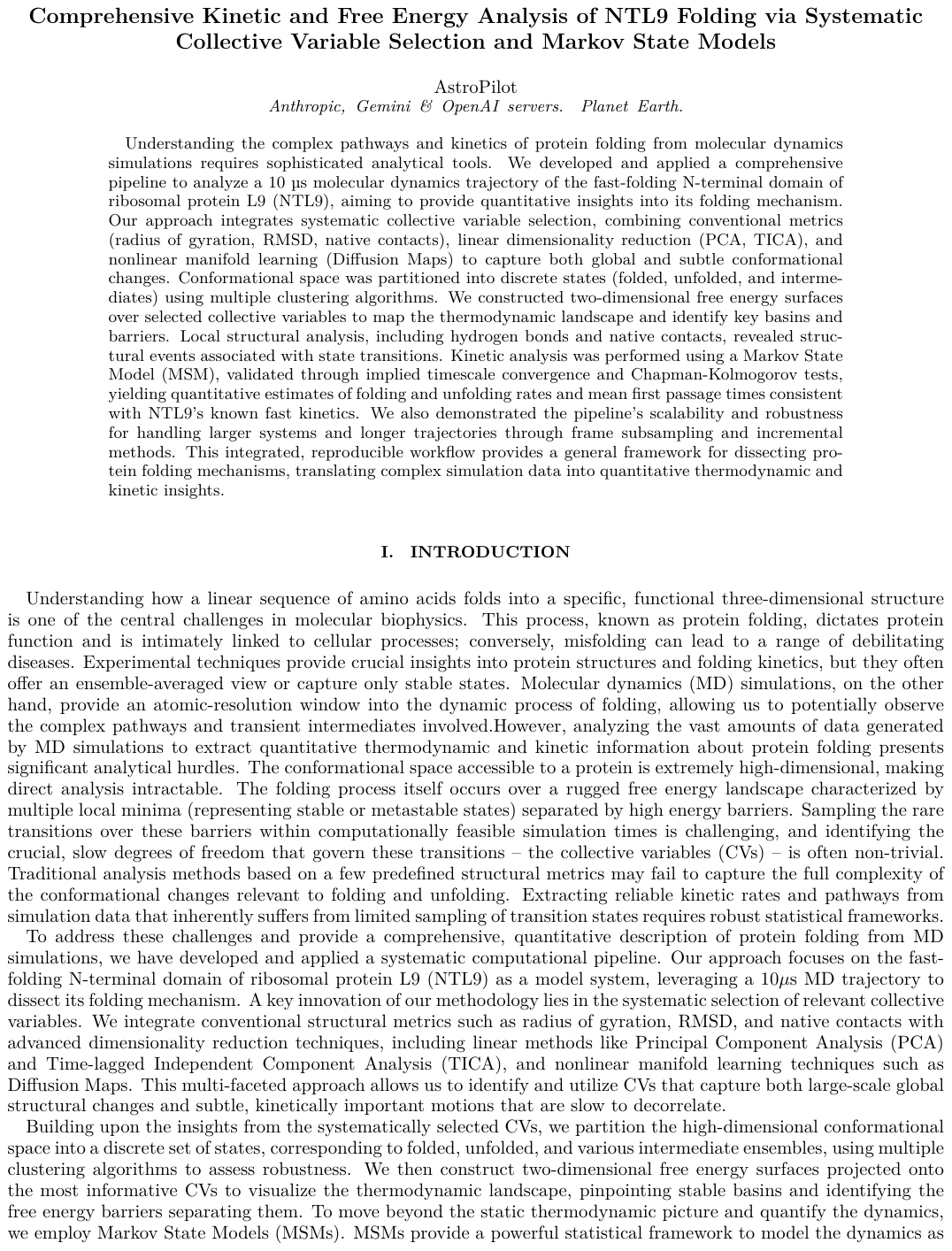}
\subsection{Neuroscience}
\label{paper:neu}
\includepdf[pages=-,landscape=false,fitpaper=true]{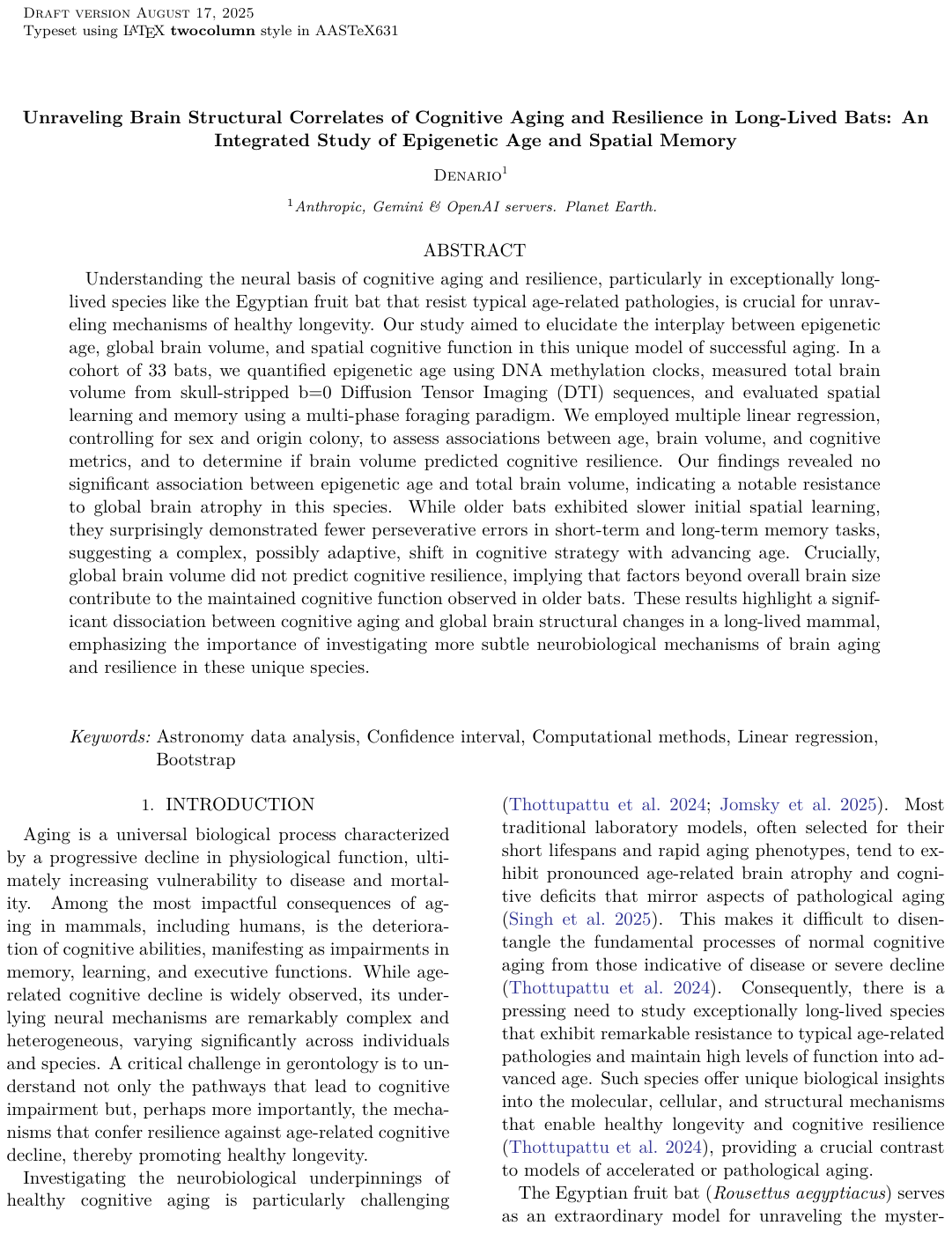}
\subsection{Material science}
\label{paper:matsci}
\includepdf[pages=-,landscape=false,fitpaper=true]{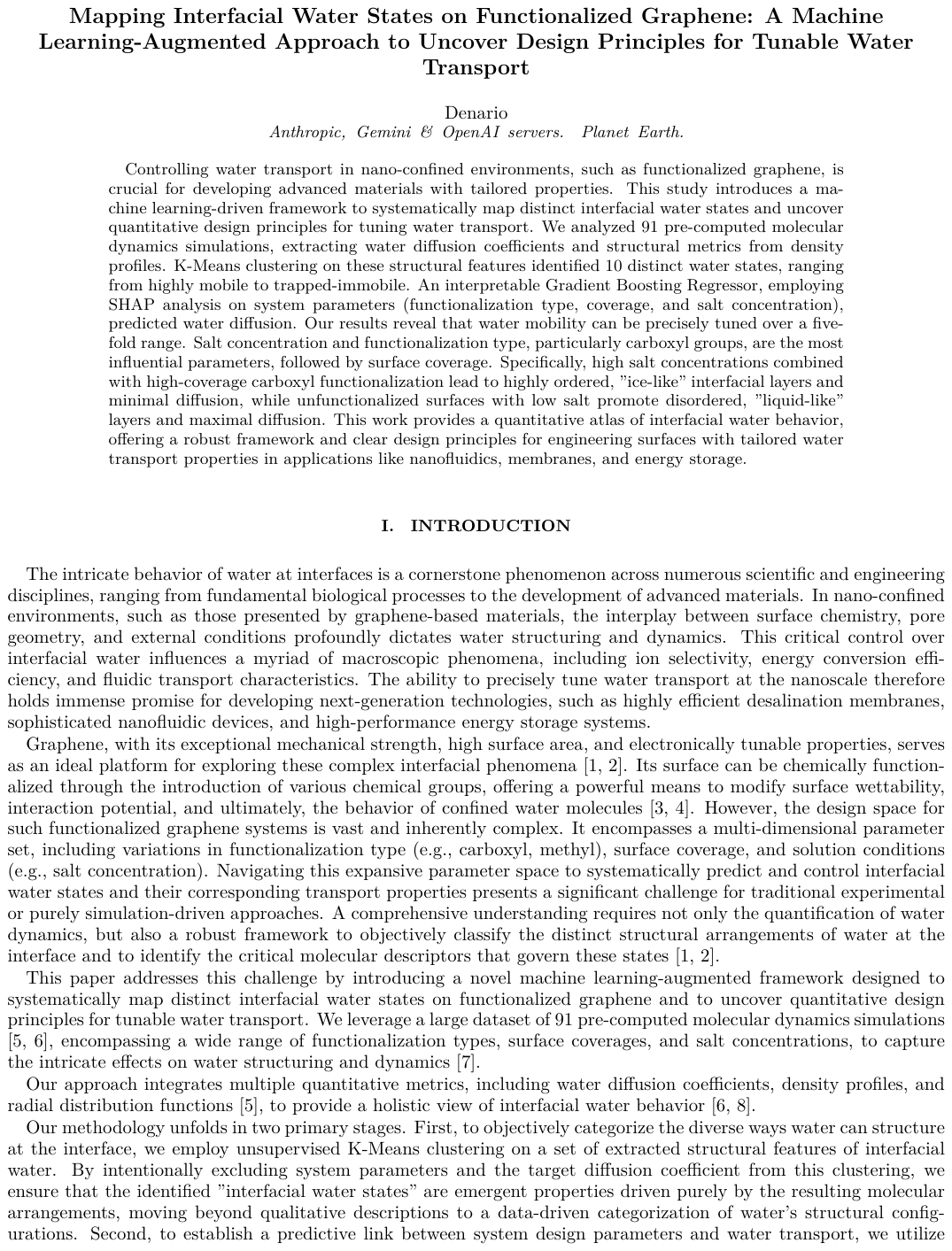}
\subsection{Astrophysics: black hole-stellar mass relation}
\label{paper:bh}
\includepdf[pages=-,landscape=false,fitpaper=true]{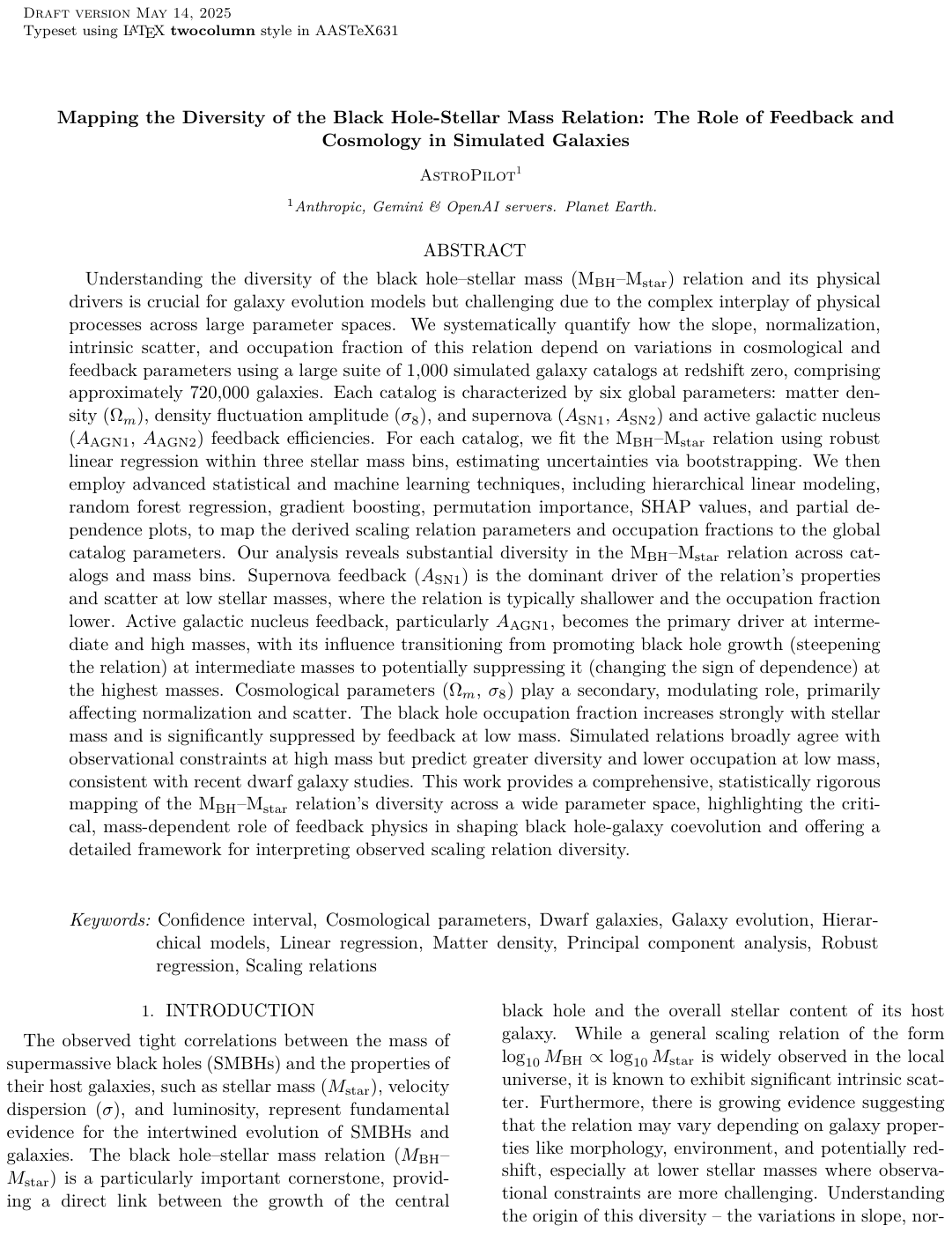}
\subsection{Quantum physics, deep learning and cosmology}
\label{paper:qitt}
\includepdf[pages=-,landscape=false,fitpaper=true]{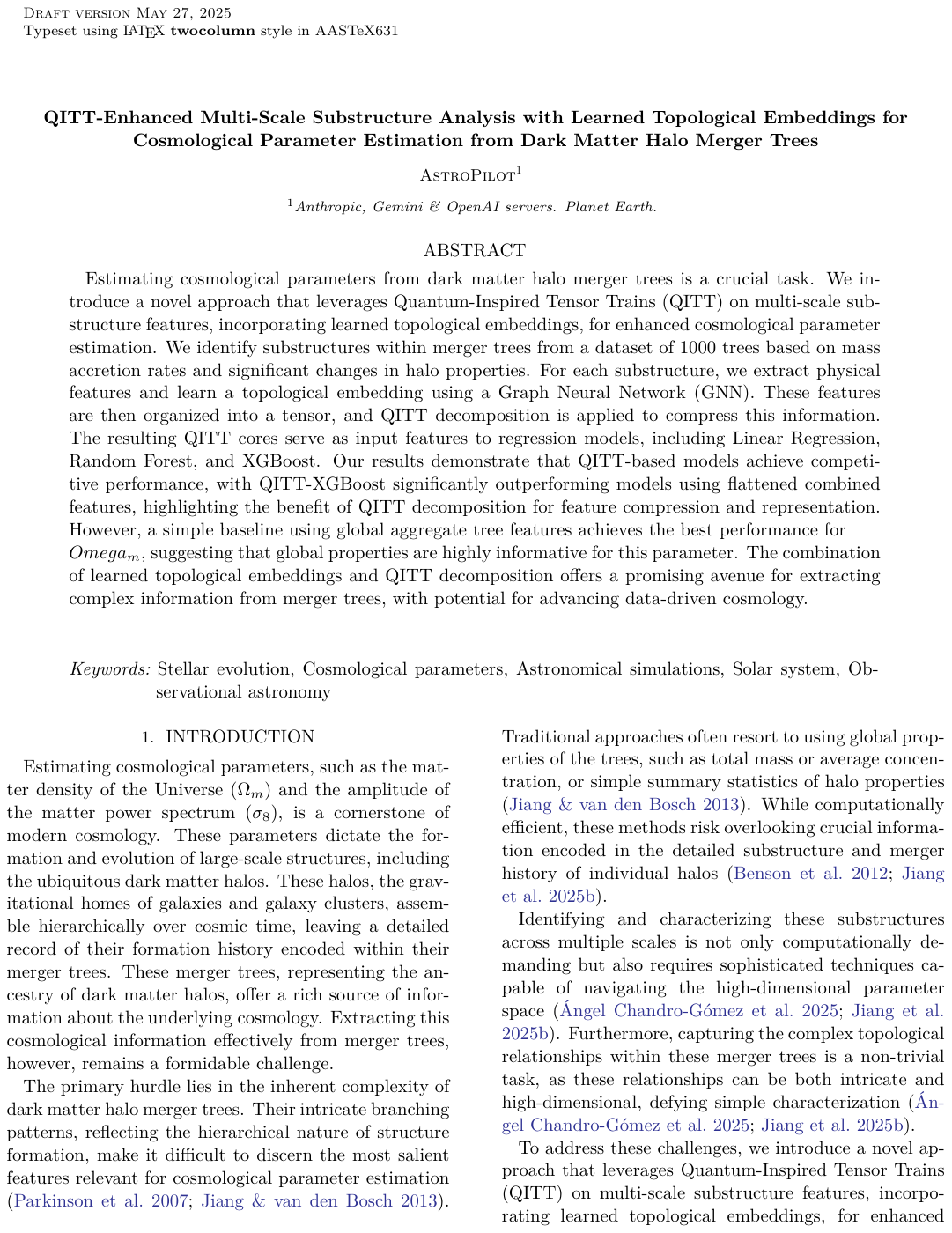}

\end{document}